\pdfoutput=1

\documentclass[11pt]{article}

\usepackage[]{ACL2023}

\usepackage{amsmath,amsfonts,bm,amsthm}

\def\Figref#1{Fig.~\ref{#1}}

\def\Tabref#1{Tab.~\ref{#1}}

\def\Quadfigref#1#2#3#4{Figures \ref{#1}, \ref{#2}, \ref{#3} and \ref{#4}}

\def\Secref#1{Sec.~\ref{#1}}
\def\Appref#1{Appx.~\ref{#1}}

\def\eqref#1{eq.~\ref{#1}}
\def\Eqref#1{Eq.~\ref{#1}}

\def\1{\bm{1}}
\newcommand{\train}{\mathcal{D}}

\def\vh{{\bm{h}}}

\def\vu{{\bm{u}}}

\def\vx{{\bm{x}}}
\def\vy{{\bm{y}}}

\def\mA{{\bm{A}}}
\def\mB{{\bm{B}}}

\def\mP{{\bm{P}}}

\def\mV{{\bm{V}}}
\def\mW{{\bm{W}}}

\DeclareMathAlphabet{\mathsfit}{\encodingdefault}{\sfdefault}{m}{sl}
\SetMathAlphabet{\mathsfit}{bold}{\encodingdefault}{\sfdefault}{bx}{n}

\def\sU{{\mathbb{U}}}
\def\sV{{\mathbb{V}}}

\newcommand{\E}{\mathbb{E}}

\newcommand{\R}{\mathbb{R}}

\newcommand{\KL}{D_{\mathrm{KL}}}

\DeclareMathOperator*{\argmax}{arg\,max}

\newcommand{\terminalcost}{\mathcal{L}_{T}}
\newcommand{\runningcost}{\mathcal{L}_{R}}
\newcommand{\megatronbert}{{\text{BERT}_\text{large}}}
\newcommand{\deberta}{{\text{Deberta}_\text{xlarge}}}
\theoremstyle{definition}
\newtheorem{definition}{Definition}[section]

\newtheorem{proposition}{Proposition}[section]

\usepackage{times}
\usepackage{latexsym}

\usepackage[T1]{fontenc}

\usepackage[utf8]{inputenc}

\usepackage{microtype}

\usepackage{inconsolata}

\usepackage{hyperref}
\usepackage{url}
\usepackage{booktabs}
\usepackage{graphicx}
\usepackage{caption}
\usepackage{subcaption}
\usepackage{wrapfig}
\usepackage{tabularx}

\title{Stochastic Bridges as Effective Regularizers for Parameter-Efficient Tuning}

\author{Weize Chen$^1$, Xu Han$^1$\thanks{\ \ Corresponding author.}, Yankai Lin$^2$, Zhiyuan Liu$^1$\footnotemark[1], Maosong Sun$^1$\footnotemark[1] and Jie Zhou$^3$\\
$^1$NLP Group, DCST, IAI, BNRIST, Tsinghua University, Beijing \\
$^2$Gaoling School of Artificial Intelligence, Renmin University of China, Beijing \\
$^3$Pattern Recognition Center, WeChat AI, Tencent Inc. \\
\texttt{chenwz21@mails.tsinghua.edu.cn}\\
\texttt{hanxu2022@tsinghua.edu.cn}
}

\begin{document}
\maketitle
\begin{abstract}
Parameter-efficient tuning methods (PETs) have achieved promising results in tuning large pre-trained language models (PLMs). By formalizing frozen PLMs and additional tunable parameters as systems and controls respectively, PETs can be theoretically grounded to optimal control and further viewed as optimizing the terminal cost and running cost in the optimal control literature. Despite the elegance of this theoretical grounding, in practice, existing PETs often ignore the running cost and only optimize the terminal cost, i.e., focus on optimizing the loss function of the output state, regardless of the running cost that depends on the intermediate states. Since it is non-trivial to directly model the intermediate states and design a running cost function, we propose to use latent stochastic bridges to regularize the intermediate states and use the regularization as the running cost of PETs. As the first work to propose regularized PETs that use stochastic bridges as the regularizers (running costs) for the intermediate states, we show the effectiveness and generality of this regularization across different tasks, PLMs and PETs. In view of the great potential and capacity, we believe more sophisticated regularizers can be designed for PETs and better performance can be achieved in the future. The code is released at \url{https://github.com/thunlp/stochastic-bridge-pet/tree/main}.
\end{abstract}

\section{Introduction}
\label{sec:introduction}

Recent years have witnessed the dramatic growth of pre-trained language models (PLMs) in various fields~\citep{DBLP:conf/naacl/DevlinCLT19,DBLP:conf/iclr/DosovitskiyB0WZ21}. As the size of PLMs continues to increase, the number of parameters has now even reached hundreds of billions~\citep{DBLP:conf/nips/BrownMRSKDNSSAA20,DBLP:journals/corr/abs-2201-11990}, making fine-tuning the whole PLM both computationally impractical and environmentally unfriendly. In view of this, a variety of Parameter-Efficient Tuning methods (PETs) are proposed~\citep{DBLP:conf/icml/HoulsbyGJMLGAG19,DBLP:conf/iclr/HuSWALWWC22,DBLP:conf/acl/ZakenGR22,DBLP:conf/emnlp/LesterAC21}. By only tuning a small number of additional parameters, PETs can be comparable to full-parameter fine-tuning.

Despite the success of PETs, their underlying mechanism remains an open problem. Recently, several works have proposed to interpret PETs with optimal control theory. \citet{DBLP:conf/iclr/YangL22a} first show that the optimization in Prefix Tuning~\citep{DBLP:conf/acl/LiL20} (a typical method of PETs) can be considered as the search for optimal control variables in the context of optimal control, i.e., the trainable prefixes can be seen as the control variables that drive the PLM (the system) to the desired output. \citet{DBLP:journals/corr/abs-2203-06904} further show that the optimal control perspective can be applied to almost all PETs.
The optimization of PETs' parameters can be seen as minimizing the two cost functions in the optimal control literature: (1) \textit{terminal cost $\terminalcost$,} which measures the quality of the terminal state, and (2) \textit{running cost $\runningcost$,} which measures the feasibility of the controlled intermediate states and the control variables. Although $\terminalcost$ can well correspond to the loss function of the model output, $\runningcost$ 
is only vaguely described as the regularizers on the parameters of PETs (control variables) in~\citet{DBLP:conf/iclr/YangL22a} and~\citet{DBLP:journals/corr/abs-2203-06904}, \emph{ignoring the dependency of $\runningcost$ on the intermediate states}.

In this work, 
we show that designing a running cost to regularize intermediate states not only makes the optimal control perspective of PETs more theoretically sound but also empirically leads to better PETs. 
We begin by assuming that in PLMs, the intermediate hidden states for generating different tokens in a sentence have different dynamics (or trajectories), and the dynamics can be approximated with stochastic processes in a latent space. Specifically, we first freeze the PLM and learn a mapping from the original hidden state space of the PLM to a latent space. In the latent space, the dynamics of the intermediate hidden states for generating different target tokens can be approximated with different target-specific \textit{diffusion bridges}. The obtained mapping can then be plugged to the model to regularize the intermediate hidden states when training PETs. Besides, since a diffusion bridge is (1) a Markov process and (2) a solution to a stochastic differential equation (SDE), we correspondingly propose two methods to learn the mapping: (1) fitting the Markov transition probability density function (PDF) and (2) fitting the SDE directly. The two methods act as a trade-off between efficiency and effectiveness: the first method incurs only negligible computational cost and has satisfactory results, while the second one is slower but yields better regularizers.

We conduct experiments on different PLMs of different sizes, and the experimental results on GLUE~\citep{DBLP:conf/iclr/WangSMHLB19} under both full-set and few-shot settings demonstrate the effectiveness of our proposal across four different PETs. Further analyses show that the learned regularizer helps pull apart the hidden states of different label words. We also observe that when we project the intermediate hidden states of PETs without our regularizer into our latent space, the better the PETs perform, the closer the latent states are to our latent bridges. This spontaneous approaching behavior may indicate that stochastic-bridge-like latent dynamics naturally exist in well-trained PETs.

In summary, our work has the following contributions:
(1) Guided by the perspective of optimal control for PETs, we design latent stochastic bridge regularizers on the intermediate states during the training of PETs.  
(2) We propose two methods to construct the latent space according to the two representations of stochastic bridges, offering a trade-off between efficiency and effectiveness.
(3) Our regularizers are shown to be effective and general across different PLMs, different PETs, and different tasks.
(4) We show that well-trained PETs without any regularization spontaneously exhibit stochastic-bridge-like latent dynamics.

\section{Background}
\label{sec:background}

\begin{figure*}[t!]
    \centering
    \includegraphics[width=0.85\linewidth]{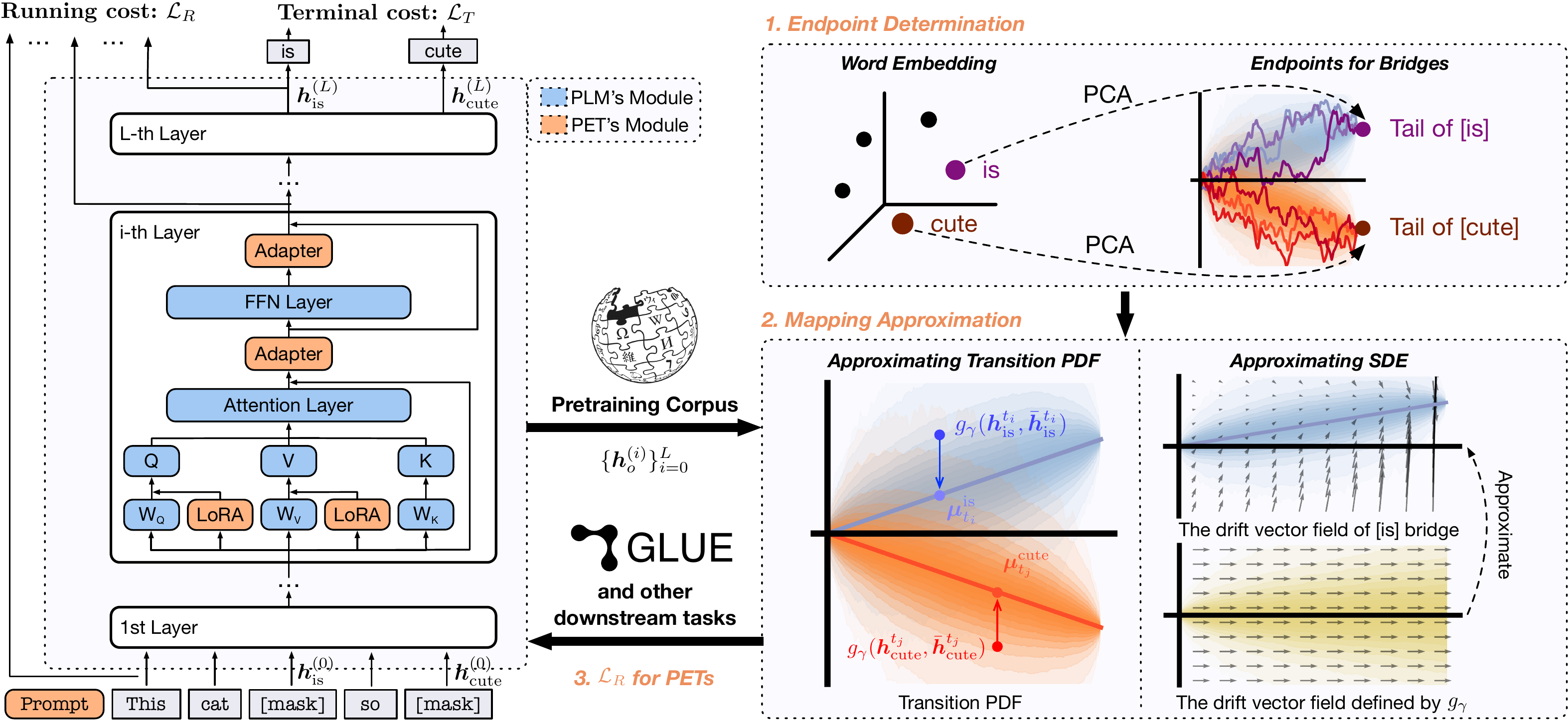}
    \caption{An overview of our proposed latent stochastic bridge regularizer.}
    \label{fig:overview}
\end{figure*}

\subsection{Definition and Mathematical Notations} 
Consider using a $L$-layer PLM with the vocabulary $\sV$ to handle a text-to-text task $\train$. For each sample $(\vx, y) \in \train$, $y\in\sV$ is the output token and $\vx\in\sV^N$ is the input token sequence~\footnote{Here we assume $y\in\sV$ since a sample where $\vy\in\sV^M$ can be decomposed to $M$ samples, The $i$-th sample is ([$\vx;\vy_{<i}$], $y_i$) for auto-regressive language modeling or ([$\vx;\vy_{-i}$], $y_i$) for auto-encoding language modeling.}, where $N$ is the length of $\vx$. With $\vx$ as the input, each layer of the PLM will output a sequence of hidden states, and we denote the hidden states of the $i$-th PLM layer as $\vh^{(i)} = \{ \vh^{(i)}_j\}_{j=1}^N\in\R^{d\times N}$.
We denote the position where the model outputs the target $y$ as $o$, i.e., the model should predict $y$ with the hidden states $\vh_o^{(L)}$.

\subsection{Optimal Control Perspective of PETs}
\label{sec:background-optimal-control-perspective-of-pets}

Conventionally, adapting the PLM to $\train$ requires full-parameter fine-tuning, which is given as:
\begin{equation}
\small
\label{eq:full-set}
\begin{aligned}
    &\min_{\Delta \theta}\E_{\vx,y \sim \train}\Big[
    \mathcal{L}\big(\vh_o^{(L)}, y\big)+
    \mathcal{R}\big(\Delta \theta\big) \Big], \\
    \vh^{(i)}=&\left\{
    	\begin{array}{lr}	
			\vh^{(i-1)}+\mathcal{G}_{\theta+\Delta \theta}^{(i)}\big(\vh^{(i-1)}\big), & i=1,\dots,L,\\
			\texttt{Embed}(\vx), i=0,
		\end{array}\right.
\end{aligned}
\end{equation}
where $\theta$ is the parameters, $\Delta \theta$ is the full-parameter update, $\mathcal{L}$ is the loss function, $\mathcal{R}$ is the regularization function, $\mathcal{G}_{\theta+\Delta \theta}^{(i)}$ is the $i$-th layer forward propagation with updated parameters, $\texttt{Embed}$ transforms the input tokens into embeddings.

As $\lvert \theta \rvert $ continues to increase, full-parameter fine-tuning becomes impractical, and various PETs are proposed to mitigate this problem. Let $\phi=\{\phi^{(i)}\}_{i=0}^{L}$ be PETs' parameters. 
\citet{DBLP:journals/corr/abs-2203-06904} give a unified view of PETs from the perspective of optimal control, and~\Eqref{eq:full-set} can be re-written as 
\begin{equation}
\small
\label{eq:pet-optimal-control}
\begin{aligned}
    &\min_{\phi}\E_{\vx,y \sim \train}\Big[\terminalcost\big(\vh_o^{(L)}, y\big)+\sum_{i=0}^{L}\runningcost \big(\phi^{(i)}\big) \Big], \\
        \vh^{(i)}&=\left\{
    	\begin{array}{lr}
			\vh^{(i-1)}+\mathcal{\tilde{G}}_\theta^{(i)}\big(\vh^{(i-1)}, \phi^{(i)}\big), & i=1,\dots,L,\\
			\big[\phi^{(0)};\texttt{Embed}(\vx)\big], & i=0,    
		\end{array}\right.
\end{aligned}
\end{equation}
where $\mathcal{\tilde{G}}_{\theta}^{(i)}$ represents the $i$-th layer forward propagation intervened by PETs, $[\cdot;\cdot]$ is the concatenation operation, $\terminalcost$ is the terminal cost and $\runningcost$ is the running cost. Since $\lvert\phi \rvert \ll \lvert\theta \rvert$, PETs can greatly reduce the tuning cost (more details in~\Appref{sec:appendix-pet}). 
Typically, $\terminalcost$ corresponds to the prediction loss, and $\runningcost$ can be seen as the regularizer on PETs' parameters $\phi$. However, in the optimal control literature, $\runningcost$ depends on not only the control variables $\phi$, but also the controlled intermediate states $\{\vh_o^{(i)}\}_{i=1}^{L}$. 
In this paper, we show that additionally introducing dependence on $\{\vh_o^{(i)}\}_{i=1}^{L}$ for $\runningcost$ 
makes the optimal control perspective of PETs more theoretically sound, and empirically leads to better PETs.

\subsection{Diffusion Bridges}
\label{sec:background-stochastic-bridges}

A diffusion process $X=(X_t)_{t\in[0,T]}$ is a continuous-time Markov process. For any $t_a < t_b$, the diffusion process is equipped with a transition Probability Density Function (PDF) $p(t_b, b | t_a, a)$, which gives the probability density of reaching $b$ at time $t_b$ given the history of reaching $a$ at time $t_a$. A diffusion process is also the solution to an Itô SDE 
$d\tilde{X}_t=\mu(t, \tilde{X}_t)dt+\sigma(t, \tilde{X}_t)dB_t$, 
where $B_t$ is a standard Brownian motion, $\mu(\cdot, \cdot)$ is called drift function and $\sigma(\cdot, \cdot)$ is called diffusion function. 

A diffusion bridge $X^{T;\alpha,\beta}$ is a diffusion process conditioning on the path observations of the two endpoints $(0, \alpha)$ and $(T, \beta)$, i.e., $X^{T;\alpha,\beta}_0=\alpha$ and $X^{T;\alpha,\beta}_T=\beta$. For simplicity, we assume $\alpha$=0 in this work, and omit the superscript $\alpha$. 
We consider two typical diffusion bridges, the Brownian bridge and the Ornstein-Uhlenbeck bridge (OU bridge). We present here the properties of the Brownian bridge and leave the properties of OU bridge to~\Appref{sec:appendix-ou-property}.
\begin{proposition}[Properties of Brownian Bridge]
A Brownian bridge $X^{T;\beta}$ with $X^{T;\beta}_0=0$ and $X^{T;\beta}_T=\beta$ is the solution to the following SDE:
\begin{equation}
\label{eq:brown-sde}
\small
\begin{aligned}
    d\tilde{X}_t=(\beta-\tilde{X}_t)/(T-t)\ dt+dB_t,\quad \tilde{X}_0=0.
\end{aligned}
\end{equation}
The transition PDF from $X^{T;\beta}_0=0$ to $X^{T;\beta}_{t_b}=b$ is given as
{\small
\begin{equation}
\label{eq:brown-transition}
\textstyle
p^{T;\beta}(t_b, b| 0, 0)=\frac{1}{\sqrt{2\pi t_b(T-t_b)}}\exp\Big[-\frac{(b-(t_b/T)\beta)^2}{2t_b(T-t_b)}\Big].
\end{equation}}
\end{proposition}
Diffusion bridges and SDEs are battle-tested tools to model the stochastic dynamics of complex systems in engineering~\citep{sobczyk2013stochastic}, finance~\citep{DBLP:journals/ior/WangS11}, etc. Considering the dynamics of PLMs' hidden states are necessarily complex, diffusion bridges and SDEs serve as ideal tools for us to model the dynamics.
\section{Latent Stochastic Bridges Regularizer}
\label{sec:method}

\subsection{The Overall Framework}
\label{sec:method-formulation}

\textbf{Building latent dynamics in the latent space.} Since directly regularizing the intermediate states and constructing the running cost are non-trivial, we introduce a projection from the intermediate state space to a latent space, and leverage diffusion bridges as regularizers to construct the running cost. Specifically, 
we define a $r$-dimensional latent space $\sU \subseteq \R^r (r<d)$ and a learnable mapping $g_\gamma:\R^d \times \R^d \rightarrow \sU$, where $\gamma$ denotes the parameters. $g_\gamma$ projects the hidden state $\vh_o^{(i)}$ and its context state $\bar{\vh}^{(i)}$ into the latent space $\sU$ at each layer of the PLM. Since $\vh_o^{(i)}$ is contextualized while latent bridges are not, introducing the dependency on $\bar{\vh}^{(i)}$ can inform $g_\gamma$ about the context at the $i$-th layer and allow $g_\gamma$ to decontextualize the hidden states. We simply take the averaged states at the $i$-th layer $\bar{\vh}^{(i)}=\frac{1}{N}\sum_{j=1}^N\vh^{(i)}_j$ as the context. We define the latent states with discrete time as 
\begin{equation}
\label{eq:latent-states-discrete}
\small
\begin{aligned}
\vu_D(g_\gamma, \{\vh_o^{(i)}\}_{i=0}^L)&=\{t_{i+1}, g_\gamma(\vh_o^{(i)},
\bar{\vh}^{(i)})\}_{i=0}^L,\\
t_{i+1}&=(i+1)/(L+2),
\end{aligned}
\end{equation}
where $t_{i+1}$ is the normalized layer index. We include the $0$-th layer (input layer) because some PETs (e.g., prompt tuning) act on the $0$-th layer. We use $t_0=0, t_{L+2}=1$ represent the two endpoints. 
By using natural cubic spline knotted at $\{\vh_o^{(i)}\}_{i=0}^L$ to interpolate over $\mathcal{L}=[-1, L+1]$, we further give a continuous representation of the states in the latent space $\sU$ as 
\begin{equation}
\label{eq:latent-states-continuous}
\small
\begin{aligned}
\vu_C(g_\gamma, \{\vh_o^{(x)}\}_{x\in\mathcal{L}})&=\{t_{x+1}, g_\gamma(\vh_o^{(x)}, \bar{\vh}^{(x)})\}_{x\in\mathcal{L}},\\
\quad t_{x+1}&=(x+1)/(L+2)\in[0, 1].    
\end{aligned}
\end{equation}

\noindent
\textbf{Learning the mapping from hidden state space to latent space.}
Since adapting PLMs to downstream tasks can be seen as transferring the knowledge obtained from pre-training tasks to downstream tasks, we argue that the latent dynamics of intermediate hidden states for generating the same token $y$ should be similar in both the pre-training and downstream tasks. 
Therefore, we train the mapping $g_\gamma$ on the corpus that is used to pre-train the backbone PLM\footnote{A small portion (0.1\%) of the pre-training corpus is sufficient, see~\Appref{sec:appendix-tiny-corpus}}, and then apply the learned mapping to downstream tasks to encourage the latent dynamics to be similar to that in pre-training. 

Specifically, we assume that the states to generate the token $y$ in the latent space $\sU$ form a trajectory that is a path sampled from $X^{1;\beta_y}$ with high probability, where $X^{1;\beta_y}$ is the diffusion bridge describing the latent dynamics to generate ${y}$, and $\beta_y$ is the tail endpoint of the diffusion bridge. 
More details of $X^{1;\beta_y}$ will be discussed in~\Secref{sec:method-endpoint}.

On the corpus where the PLM is pre-trained, we fix the PLM and use its hidden states $\{\vh^{(i)}_o\}_{i=1}^L$ to learn $g_\gamma$ by maximizing the goodness of approximation for latent states $\vu$ under the bridge $X^{1;\beta_y}$:
\begin{equation}
    \small
    \gamma \leftarrow \argmax_{\gamma'} \big[\text{goodness}\big(\vu(g_{\gamma'}, \{\vh_o^{(\cdot)}\}), X^{1;\beta_y}\big)\big],
    \label{eq:train-mapping}
\end{equation}
where $\vu$ can be $\vu_D$~(\Eqref{eq:latent-states-discrete}) or $\vu_C$~(\Eqref{eq:latent-states-continuous}) depending on the fitting method, $\text{goodness}(\cdot, \cdot)$ is also a function depends on the choice of the fitting method, measuring how likely $\vu$ is a sample trajectory of $X^{1;\beta_y}$. In~\Secref{sec:method-fitting}, we will define this function alongside the fitting methods. 

\noindent
\textbf{Regularizing PETs with latent dynamics.}
After learning $g_\gamma$ with \Eqref{eq:train-mapping}, we freeze $\gamma$ and use the goodness function as the running cost in Eq.~\ref{eq:pet-optimal-control} for PETs on downstream tasks. The objective becomes
\begin{equation}
\small
\label{eq:overall-loss}
\begin{aligned}
    \mathcal{L}=\terminalcost(\bm{h}_o^{(L)},y) + \alpha\cdot \text{goodness}\big(\vu(g_\gamma, \{\vh_o^{(\cdot)}\}), X^{1;\beta_y}\big), 
\end{aligned}
\end{equation}
where the second term is the running cost and $\alpha$ is a hyper-parameter controlling the regularization intensity. By optimizing \Eqref{eq:overall-loss}, PETs learn to predict $y$ correctly and keep the latent states at the position $o$ conform to the diffusion bridge $X^{1;\beta_y}$. Note that introducing $g_\gamma$ as the regularizer does not increase the number of trainable parameters for PETs during the training stage since $\gamma$ is fixed, and since we only use the pre-training corpus, no extra information of downstream tasks is leaked. Moreover, the regularizer only helps in training better PETs and does not intervene the inference.

\subsection{Determining Endpoints for Bridges}
\label{sec:method-endpoint}

An intuitive approach to determine the endpoints for the diffusion bridges for each target token is to optimize the endpoints together with the mapping $g_\gamma$. However, optimizing endpoints and $g_\gamma$ jointly may admit a trivial solution: endpoints are both $\mathbf{0}\in\R^{r}$ and $g_\gamma$ always outputs $\mathbf{0}$. Since $\mathbf{0}$ is always a point in the sample path of such a degenerated diffusion bridge, the value of goodness function can be meaninglessly high. Although sophisticated constraints can be imposed here, as the first work that uses diffusion bridges as regularizers, we simply pre-determine the endpoints and keep them fixed. We leave introducing constraints as future work. 

Specifically, we apply principal component analysis (PCA) to the output token embedding matrix $\mV\in\R^{|\sV|\times d}$ of the PLM, obtaining a $r$-dimensional embedding matrix, and re-normalize each row to have a norm $\eta$. Let the resulting embedding matrix be $\bm{\beta}\in\R^{|\sV|\times r}$. We then use $\mathbf{0}\in\R^r$ as the heads for all the bridges, and $\bm{\beta}$ as the tails of the diffusion bridges, i.e., the $r$-dimensional embedding of $y$ in $\bm{\beta}$ is used as $\beta_y$ in $X^{1;\beta_y}$. 
The intuition for using $\bm{\beta}$ as the tails is that the trajectories of the intermediate states for similar target tokens should be close. In $\mV$, similar tokens are close, and $\bm{\beta}$ obtained by PCA can well preserve the token similarity after reducing dimensions. 

\subsection{Fitting the Mapping $g_{\gamma}$}
\label{sec:method-fitting}
We use the Brownian bridge to illustrate the fitting of $g_\gamma$. It can be analogous to OU bridge easily.

\noindent
\textbf{Method 1: Approximating the Transition PDF.}
Generalizing~\Eqref{eq:brown-transition} to high dimension, we can derive the transition PDF from $(0, \bm{0})$ to $(t_{i+1}, g_\gamma(\vh_o^{(i)}, \bar{\vh}^{(i)}))$ for $X^{1;\beta_y}$:
\begin{equation*}
\small
\begin{aligned}
&\quad p^{1;\beta_y}(t_{i+1}, g_\gamma(\vh_o^{(i)}, \bar{\vh}^{(i)})\mid 0, \bm{0})\\
&\propto\exp(\frac{\lVert g_\gamma(\vh_o^{(i)}, \bar{\vh}^{(i)})-t_{i+1}\beta_y\rVert^2}{2t_{i+1}(1-t_{i+1})}),\quad (i=0,\dots,L),
\end{aligned}
\end{equation*}
where $t_i$ has the same definition as that in $\vu_D$~(\Eqref{eq:latent-states-discrete}). To make $g_\gamma$ approximate the transition PDF, we maximize the sum of log-probability of $\vu_D$ under the Brownian bridge $X^{1;\beta_y}$:
\begin{equation}
\small
\begin{aligned}
    \text{goodness}=\sum_{i=0}^{L}\log\big[p^{1;\beta_y}(t_{i+1}, g_\gamma(\vh_o^{(i)}, \bar{\vh}^{(i)})\mid 0, \bm{0})\big]+C,
\end{aligned}
    \label{eq:goodness-pdf}
\end{equation}
where $C$ is a constant. Here, $g_\gamma$ can be seen as a mapping from the hidden state space to the latent space by predicting the expectation of the Brownian bridge $X^{1;\beta_y}$ at $\{t_{i+1}\}_{i=0}^L$. 

\noindent
\textbf{Method 2: Approximating the SDE.}
Since the Brownian bridge is the solution to the SDE in~\Eqref{eq:brown-sde}, we let $g_\gamma$ approximate the SDE. Solving the SDE requires continuous latent states, while we only have $L+1$ discrete observations, we thus use the continuous representation $\vu_C$ introduced in~\Eqref{eq:latent-states-continuous}. Generalizing~\Eqref{eq:brown-sde} to high dimension, the SDE approximated by $g_\gamma$ can be defined as:
\begin{equation}
\label{eq:approximate-sde}
\small
\begin{aligned}
    dZ_{t}=g_\gamma(\vh_o^{(x)}, \bar{\vh}^{(x)}, t)dt+dB_t,\  x=(L+2)t-1,
\end{aligned}
\end{equation}
where $x$ is the same as that in~\Eqref{eq:latent-states-continuous}, $B:[0,1]\rightarrow \R^r$ is a standard $r$-dimensional Brownian motion. Here, we additionally introduce the dependence on $t$ for $g_\gamma$, since time information is shown to be important in previous neural differential equation works~\citep{DBLP:conf/icml/ZhangGUA20,DBLP:conf/nips/DupontDT19}. Following~\citet{DBLP:conf/aistats/LiWCD20}, when two SDEs share the same diffusion function, the KL divergence between the probability measures induced by the two SDEs is finite. Since the diffusion function $\sigma\equiv \bm{I}$ for~\Eqref{eq:approximate-sde} and the multi-dimensional generalization of~\Eqref{eq:brown-sde}, the KL divergence between the probability measures $\mu_Y$ of~\Eqref{eq:approximate-sde} and $\mu_X$ of generalized~\Eqref{eq:brown-sde} can be estimated by:
\begin{equation*}
\small
\begin{aligned}
    \KL(\mu_X||\mu_Y)&=\E_{Z}\big[\int_0^T\frac{1}{2}\lVert u(t, \gamma)\rVert_2^2\big], \\
    u(t, \gamma) &=\sigma^{-1}\big(g_\gamma(\vh_o^{(x)}, \bar{\vh}^{(x)}, t) -\mu(t, Z_{t})\big)\\
    &=g_\gamma(\vh_o^{(x)}, \bar{\vh}^{(x)}, t)-(\beta_y-Z_{t})/(1-t),
    \end{aligned}
\end{equation*}
where $\mu(\cdot, \cdot)$ is the drift function of the pre-determined Brownian bridge $X^{1;\beta_y}$. We use the KL divergence as the goodness function to optimize the mapping $g_\gamma$. Here, $g_\gamma$ can be seen as a mapping from the hidden state space to the latent state space by approximating the drift vector field of the underlying Brownian bridge $X^{1;\beta_y}$.
\section{Experiments}
\label{sec:experiments}

To verify the effectiveness and generality of the regularizers built on stochastic bridges, we conduct experiments on (1) different PLMs: $\megatronbert$ (340M)~\citep{DBLP:conf/naacl/DevlinCLT19} and $\deberta$ (750M)~\citep{DBLP:conf/iclr/HeLGC21}; (2) different PETs: Prompt tuning, LoRA, BitFit and Adapter; (3) different diffusion bridges: Brownian bridge and OU bridge. We show that the regularizers effectively improve the performance on GLUE~\citep{DBLP:conf/iclr/WangSMHLB19} under both full-set and few-shot settings. 

\subsection{Experimental Setups}
\label{sec:experiments-setups}
\textbf{Datasets.} Since both $\megatronbert$ and $\deberta$ use Wikipedia and BookCorpus~\citep{DBLP:conf/iccv/ZhuKZSUTF15} for pre-training, we thus use these two corpora to train $g_\gamma$. We report F1 for MRPC and QQP; Matthews correlation for CoLA; and accuracy for other tasks. We report the average performance and the standard deviation on the development set over 3 different runs. We append \texttt{[MASK]} to each sequence, and require the PLM to output the label word at \texttt{[MASK]} (e.g., \textit{negative} or \textit{positive} for SST-2). We exclude STS-B for it is a regression task.

\noindent
\textbf{Models and PETs.} We use the checkpoint released by~\citet{DBLP:journals/corr/abs-1909-08053} for $\megatronbert$, and the official v1 checkpoint for $\deberta$. We use a simple three-layer MLP to build $g_\gamma$. For Prompt tuning, we use a soft prompt of length 20, and append it to the end of each sequence. For LoRA, we apply it to the query and value of attention modules. For Adapter, we apply it to the output of attention and feed-forward modules. For BitFit, we tune all the bias terms in linear layers and layer normalization modules. Hereafter, we use \textbf{\textsc{PDF} regularizer} to refer to using $g_\gamma$ fitted by approximating the transition PDF, and \textbf{\textsc{SDE} regularizer} to refer to using $g_\gamma$ fitted by approximating the SDE, \textbf{vanilla} $x$ to refer to the PET $x$ without using regularizers.

\noindent
\textbf{Few-shot Experiments.} We randomly sample $2\times k$ examples from the original training set $\mathcal{D}_{\text{train}}$ for each class. The sampling is performed 5 times with different seeds to form 5 training sets and development sets $\{\tilde{\mathcal{D}}_{\text{train}}^{(i)}, \tilde{\mathcal{D}}_{\text{dev}}^{(i)}\}_{i=1}^5$ with each being $k$-shot. Each time we train PETs on  $\tilde{\mathcal{D}}^{(i)}_\text{train}$, we select the best model on $\tilde{\mathcal{D}}^{(i)}_\text{dev}$, and report its performance on the original development set $\mathcal{D}_\text{dev}$. 

\begin{table*}[t!]
    \centering
    \renewcommand{\arraystretch}{0.7}
    \resizebox{0.85\linewidth}{!}{
    \begin{tabular}{l *{9}{r}}
        \toprule
            PET & MNLI & QQP & QNLI & SST-2 & MRPC & CoLA & RTE & Average & $\Delta$\\ 
        \midrule
            \textsc{Prompt} & 84.4$_{0.1}$ & 85.3$_{0.3}$ & 91.5$_{0.1}$ & 95.5$_{0.1}$ & 73.9$_{2.4}$ & 55.5$_{3.4}$ & 60.8$_{1.5}$ & 78.1$_{0.6}$&-\\
            \textsc{+Brown\_PDF} & 84.7$_{0.2}$ & \textbf{85.5}$_{0.0}$ & \textbf{91.8}$_{0.6}$ & 95.7$_{0.1}$ & 75.4$_{0.5}$ & 56.4$_{3.3}$ & 61.5$_{2.2}$ & 78.7$_{0.4}$&0.6\\
            \textsc{+Brown\_SDE} & \textbf{84.9}$_{0.2}$ & 85.4$_{0.1}$ & \textbf{91.8}$_{0.4}$ & \textbf{95.8}$_{0.3}$ & \textbf{78.8}$_{1.2}$ & \textbf{61.4}$_{2.9}$ & \textbf{64.7}$_{1.1}$ & \textbf{80.4}$_{0.2}$&\textbf{2.3}\\
        \midrule
            \textsc{LoRA} & 88.8$_{0.1}$ & 89.2$_{0.2}$ & 93.5$_{0.2}$ & 95.5$_{0.1}$ & 84.6$_{0.4}$ & 62.8$_{1.6}$ & 78.9$_{1.6}$ & 84.8$_{0.3}$&-\\
            \textsc{+Brown\_PDF} & \textbf{88.9}$_{0.1}$ & \textbf{89.6}$_{0.1}$ & \textbf{93.9}$_{0.1}$ & 95.6$_{0.2}$ & 85.1$_{0.7}$ & 63.7$_{0.5}$ & 80.0$_{0.5}$ & 85.2$_{0.1}$&0.4\\
            \textsc{+Brown\_SDE} & \textbf{88.9}$_{0.1}$ & 89.5$_{0.1}$ & 93.7$_{0.1}$ & \textbf{95.7}$_{0.1}$ & \textbf{86.5}$_{1.2}$ & \textbf{63.9}$_{0.4}$ & \textbf{80.9}$_{0.8}$ & \textbf{85.6}$_{0.1}$& \textbf{0.8}\\ 
        \midrule
            \textsc{BitFit} & \textbf{87.9}$_{0.2}$ & 87.6$_{0.1}$ & 92.7$_{0.2}$ & 95.6$_{0.1}$ & 79.4$_{2.3}$ & 60.2$_{0.8}$ & 77.0$_{1.5}$ & 82.9$_{0.3}$&-\\
            \textsc{+Brown\_PDF} & \textbf{87.9}$_{0.1}$ & \textbf{87.8}$_{0.0}$ & \textbf{93.0}$_{0.2}$ & \textbf{95.7}$_{0.1}$ & 83.1$_{0.8}$ & 60.3$_{0.6}$ & \textbf{78.3}$_{0.9}$ & 83.7$_{0.2}$&0.8\\
            \textsc{+Brown\_SDE} & \textbf{87.9}$_{0.2}$ & 87.7$_{0.0}$ & 92.8$_{0.1}$ & \textbf{95.7}$_{0.1}$ & \textbf{83.3}$_{0.8}$ & \textbf{61.1}$_{1.2}$ & 77.7$_{1.5}$ & \textbf{83.8}$_{0.3}$& \textbf{0.9}\\ 
        \midrule
            \textsc{Adapter} & 88.8$_{0.1}$ & 89.6$_{0.3}$ & 93.7$_{0.1}$ & 95.6$_{0.1}$ & 83.6$_{0.1}$ & 60.4$_{1.2}$ & 79.5$_{1.2}$ & 84.5$_{0.3}$&-\\
            \textsc{+Brown\_PDF} & \textbf{89.0}$_{0.1}$ & 89.7$_{0.2}$ & 93.8$_{0.2}$ & \textbf{95.8}$_{0.1}$ & \textbf{86.5}$_{1.1}$ & \textbf{62.6}$_{0.7}$ & \textbf{83.2}$_{0.2}$ & \textbf{85.8}$_{0.2}$& \textbf{1.3}\\ 
            \textsc{+Brown\_SDE} & 88.9$_{0.1}$ & \textbf{89.8}$_{0.1}$ & \textbf{93.9}$_{0.2}$ & \textbf{95.8}$_{0.1}$ & 85.9$_{0.4}$ & 62.3$_{1.8}$ & 82.2$_{0.2}$ & 85.5$_{0.2}$&1.0\\
        \bottomrule
    \end{tabular}}
    \caption{The results on GLUE for $\megatronbert$. The values are the average value of the best performances over three different runs, and the subscripts are the standard deviations. The $\Delta$ column shows the difference of the average performance between the vanilla PETs regularized PETs.}
    \label{tab:megatron-glue-full}
\end{table*}
\noindent
\textbf{Hyper-parameters.} Hyper-parameters are listed in~\Appref{sec:appendix-hyperparameters}. We focus on the difference in performance between vanilla PETs and regularized PETs. Therefore, we set the hyper-parameters to common values from previous works and do not perform much hyper-parameter search. But we ensure the hyper-parameters for vanilla PETs and regularized PETs are the same for a fair comparison.

\subsection{Full-set Results}
\label{sec:experiments-glue}

The experimental results for $\megatronbert$ and $\deberta$ are reported in~\Tabref{tab:megatron-glue-full} and ~\Appref{sec:appendix-glue-all} respectively. Due to space limitation, see~\Tabref{tab:megatron-glue-full-complete} for the complete results including OU bridge regularizers. The first line of each block in the table is the performance of vanilla PETs, and the rest of the lines are the performances of the regularized PETs. 

In general, both Brownian and OU bridges, and both \textsc{PDF} and \textsc{SDE} regularizers are able to improve the performance of PETs, showing the effectiveness of our proposed regularizers. Particularly, for Prompt tuning, the \textsc{SDE} regularizer with both diffusion bridges yield an average performance improvement of more than 2\%. We assume that it is because Prompt tuning has far less trainable parameters than other PETs, and it only acts at the input layer, which is far from the supervision signals of the terminal cost $\terminalcost$. Therefore, when provided with the regularization on the hidden states, the prompts receive more guidance and eventually reaching a better local optimal. 

Overall, the two diffusion bridges in our experiments do not show much difference. As for the two fitting methods, \textsc{SDE} regularizer is generally more effective, especially for Prompt tuning where the number of trainable parameters is restricted. However, we also observe that \textsc{SDE} regularizer is about 3 times slower than \textsc{PDF} regularizer, which brings the trade-off between performance and efficiency. One can expect a better performance by leveraging more sophisticated underlying stochastic bridges, exploring more reasonable endpoints for bridges and designing better mapping $g_\gamma$. As the first work using latent stochastic bridges as regularizers, we mainly consider the most simple cases and aim to show the potential of the approach.

\subsection{Few-shot Results}
\label{sec:experiments-glue-fewshot}

\begin{table*}[t!]
    \centering
    \renewcommand{\arraystretch}{0.7}
    \resizebox{0.85\linewidth}{!}{
    \begin{tabular}{l *{8}{r}}
    \toprule
         PET & MNLI & QQP & QNLI & SST-2 & MRPC & RTE & Average & $\Delta$\\ 
    \midrule
        \textsc{Prompt} & 38.1$_{1.5}$ & 53.0$_{3.1}$ & 51.6$_{1.4}$ & 70.1$_{4.9}$ & 50.1$_{3.0}$ & 48.0$_{1.3}$ & 51.8$_{0.9}$ &-\\
        \textsc{+Brown\_PDF} & 38.7$_{2.3}$ & 54.9$_{2.8}$ & 52.1$_{1.1}$ & 75.0$_{11.0}$ & \textbf{52.8}$_{2.2}$ & 50.8$_{3.3}$ & 54.0$_{1.8}$ &2.2\\
        \textsc{+Brown\_SDE} & \textbf{40.6}$_{0.8}$ & \textbf{55.4}$_{2.1}$ & \textbf{52.9}$_{1.6}$ & \textbf{80.0}$_{10.9}$ & 51.9$_{3.6}$ & \textbf{51.7}$_{3.1}$ & \textbf{55.4}$_{1.5}$ &\textbf{3.6}\\
    \midrule
        \textsc{LoRA} & 48.7$_{4.5}$ & 59.9$_{5.5}$ & 53.2$_{1.2}$ & 90.2$_{1.1}$ & 53.6$_{3.4}$ & 64.2$_{0.9}$ & 61.6$_{0.6}$ &-\\
        \textsc{+Brown\_PDF} & 52.0$_{1.3}$ & 62.7$_{2.2}$ & 55.1$_{3.8}$ & \textbf{91.3}$_{0.1}$ & 57.4$_{5.0}$ & 65.3$_{1.5}$ & 64.0$_{0.7}$ &2.4\\
        \textsc{+Brown\_SDE} & \textbf{54.1}$_{0.9}$ & \textbf{65.5}$_{1.4}$ & \textbf{64.3}$_{5.1}$ & 91.2$_{0.3}$ & \textbf{60.2}$_{3.1}$ & \textbf{65.8}$_{1.4}$ & \textbf{66.8}$_{0.6}$ & \textbf{5.2}\\ 
    \midrule
        \textsc{BitFit} & 48.4$_{1.6}$ & 56.0$_{6.1}$ & 51.7$_{2.5}$ & 90.8$_{0.8}$ & 52.0$_{2.3}$ & 61.7$_{1.4}$ & 60.1$_{1.1}$ &-\\
        \textsc{+Brown\_PDF} & 48.5$_{1.9}$ & 56.0$_{6.0}$ & 53.5$_{2.0}$ & \textbf{91.0}$_{0.4}$ & 53.9$_{2.3}$ & 63.2$_{1.6}$ & 61.0$_{0.8}$ &0.9\\
        \textsc{+Brown\_SDE} & \textbf{52.3}$_{0.5}$ & \textbf{61.2}$_{2.9}$ & \textbf{58.8}$_{4.7}$ & 90.8$_{0.4}$ & \textbf{54.8}$_{3.0}$ & \textbf{63.9}$_{2.5}$ & \textbf{63.6}$_{0.8}$ & \textbf{3.5}\\ 
    \midrule
        \textsc{Adapter} & 47.4$_{3.7}$ & 57.0$_{7.2}$ & 55.8$_{2.9}$ & 91.0$_{0.4}$ & 55.8$_{2.5}$ & 62.7$_{2.0}$ & 61.6$_{1.2}$ &-\\
        \textsc{+Brown\_PDF} & 49.0$_{4.8}$ & 58.5$_{7.4}$ & 56.9$_{3.1}$ & 91.4$_{0.2}$ & 57.2$_{4.9}$ & 63.2$_{3.0}$ & 62.7$_{1.5}$ &1.1\\
        \textsc{+Brown\_SDE} & \textbf{52.3}$_{2.2}$ & \textbf{62.4}$_{2.9}$ & \textbf{64.8}$_{4.5}$ & \textbf{91.9}$_{0.4}$ & \textbf{57.3}$_{4.1}$ & \textbf{63.8}$_{1.8}$ & \textbf{65.4}$_{1.3}$ & \textbf{3.8}\\ 
    \bottomrule
    \end{tabular}}
    \caption{The results on GLUE for $\megatronbert$ under the 16-shot setting. We exclude CoLA because all PETs fail to give reasonable results under the few-shot setting.}
    \label{tab:glue-16shot}
\end{table*}
In~\Tabref{tab:megatron-glue-full}, the improvements are more substantial on small datasets MRPC, CoLA and RTE. This is probably because in large datasets, the abundant data has provided enough information to train high quality PETs; while in small datasets, the data is insufficient and the regularizer can offer additional supervision. To validate this, we conduct the experiments under the few-shot setting on GLUE. 

The 16-shot results are shown in~\Tabref{tab:glue-16shot}, and the results for the OU bridge, results for 4-, 8- and 32-shot and results for $\deberta$ are placed in~\Appref{sec:appendix-glue-all}. For all PETs, the \textsc{SDE} regularizer yields an improvement of more than 3\%. Particularly, the \textsc{SDE} regularizer on LoRA brings an improvement of 5.2\%. Also, there is now a substantial boost on what was originally a rich-resource dataset, such as MNLI, QQP and QNLI. The PDF regularizer also gives modest improvements. Though slightly inferior to the \textsc{SDE} regularizer, it is still satisfying, considering that the \textsc{PDF} regularizer brings such a performance improvement with little computational cost introduced. We additionally observe that the improvement is more significant on $\deberta$ in~\Tabref{tab:glue-deberta-16shot}, demonstrating the potential of our regularizers on larger models.

\section{Analyses}
\label{sec:analysis}
To better understand the role played by our regularizers, we analyze the hidden states of the PETs with and without regularizers. We choose Prompt tuning as a representative. By varying the hyper-parameter $\alpha$ in~\Eqref{eq:overall-loss}, we show that as the regularization intensity gets stronger, the clusters of hidden states corresponding to different labels become more distinguishable. Also, we show that the hidden states of vanilla PETs spontaneously approach the latent bridges in the latent space without knowing the bridges, indicating that there may exist intrinsically diffusion-bridge-like latent dynamics for PETs.

\subsection{Distances between Labels are Widen}
\label{sec:analysis-centroid-distance}
We use the different prompts obtained with or without regularizers on the full-set GLUE, and record the intermediate hidden states $\{\vh^{(i)}_{\text{[MASK]}}\}_{i=1}^{L}$. We vary the regularization intensity by adjusting the coefficient $\alpha$ in~\Eqref{eq:overall-loss} to inspect the impact of the regularization intensity on the hidden states. Note that when $\alpha=0$, it degenerates to the vanilla PET. 

\begin{figure*}[t!]
    \centering
    \begin{subfigure}[b]{0.2\textwidth}
        \centering
        \includegraphics[width=\textwidth]{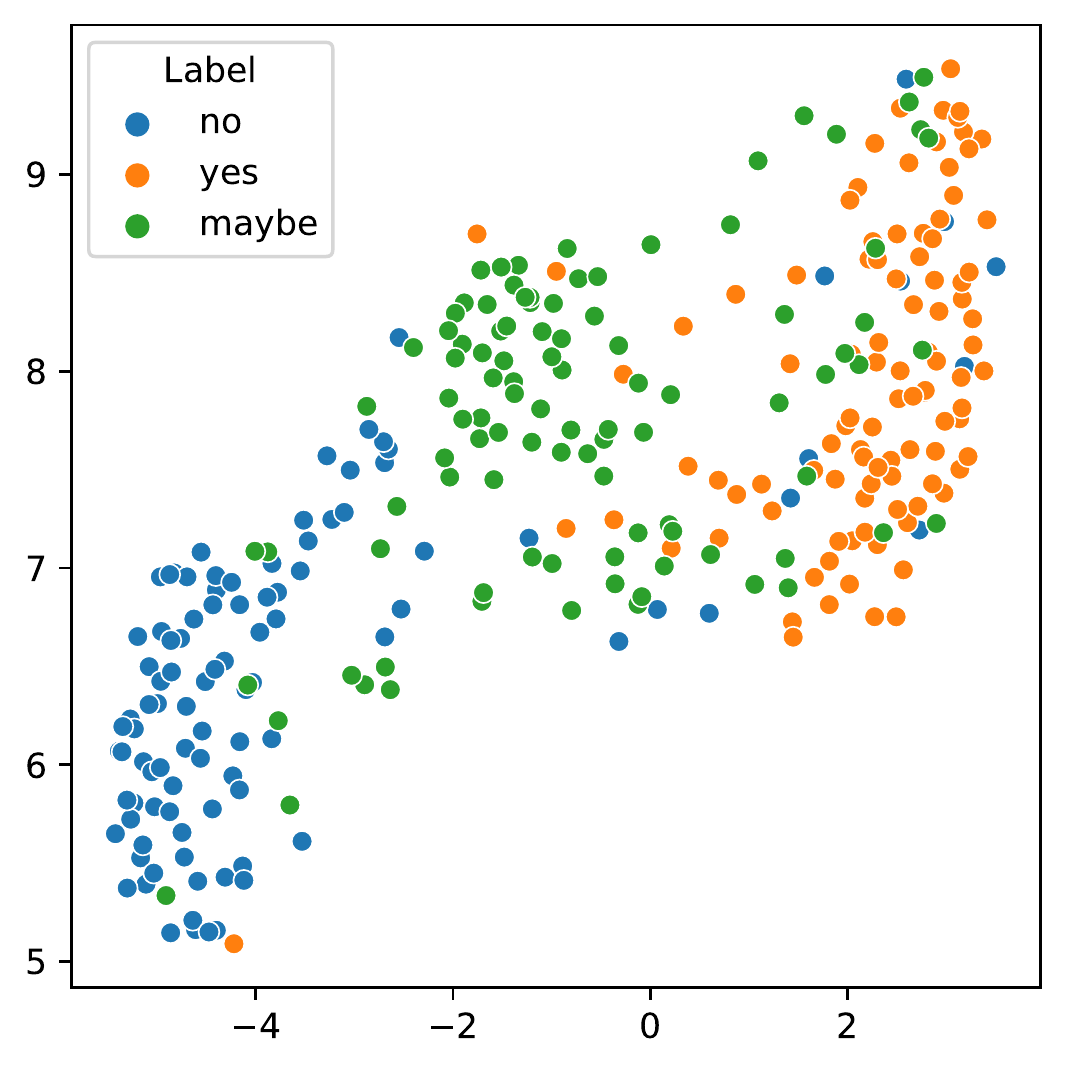}
        \caption{w/o regularizer}
        \label{fig:analyze-mnli-sde-0}
    \end{subfigure}
    \hfill
    \begin{subfigure}[b]{0.2\textwidth}
        \centering
        \includegraphics[width=\textwidth]{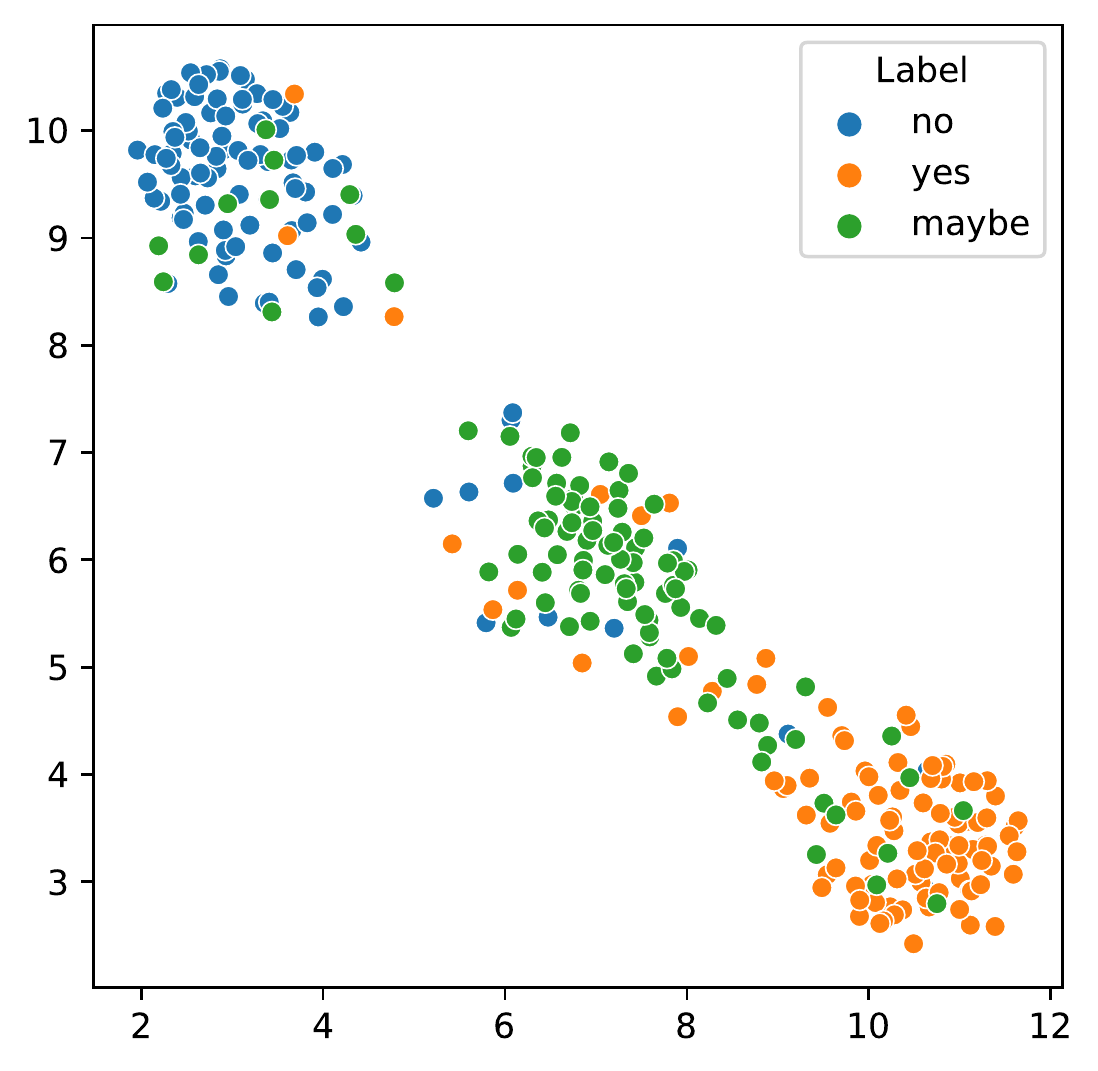}
        \caption{\textsc{SDE} $\alpha=0.001$}
        \label{fig:analyze-mnli-sde-0.001}
    \end{subfigure}
    \hfill
    \begin{subfigure}[b]{0.2\textwidth}
        \centering
        \includegraphics[width=\textwidth]{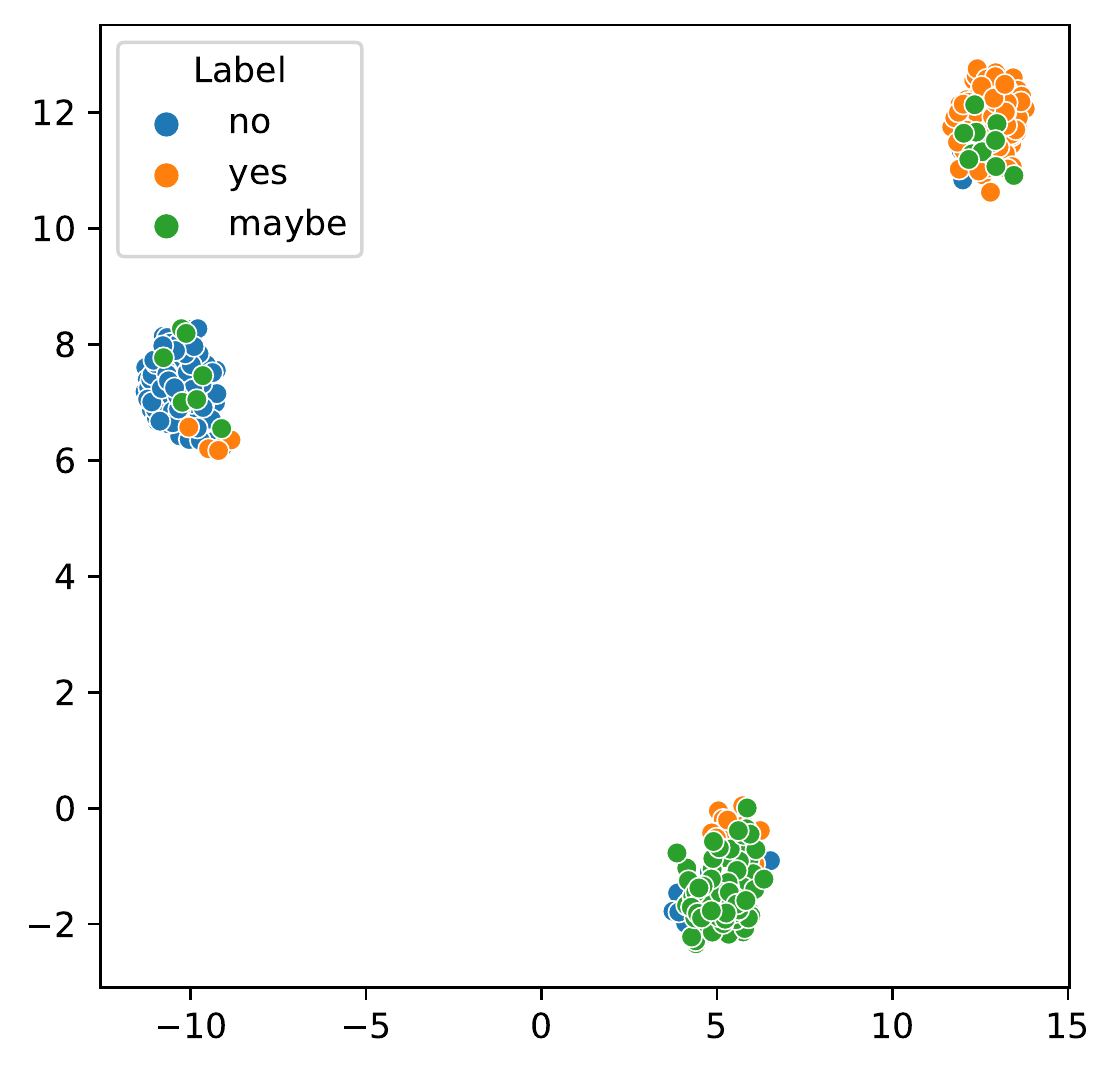}
        \caption{\textsc{SDE} $\alpha=0.01$}
        \label{fig:analyze-mnli-sde-0.01}
    \end{subfigure}
    \hfill
    \begin{subfigure}[b]{0.2\textwidth}
        \centering
        \includegraphics[width=\textwidth]{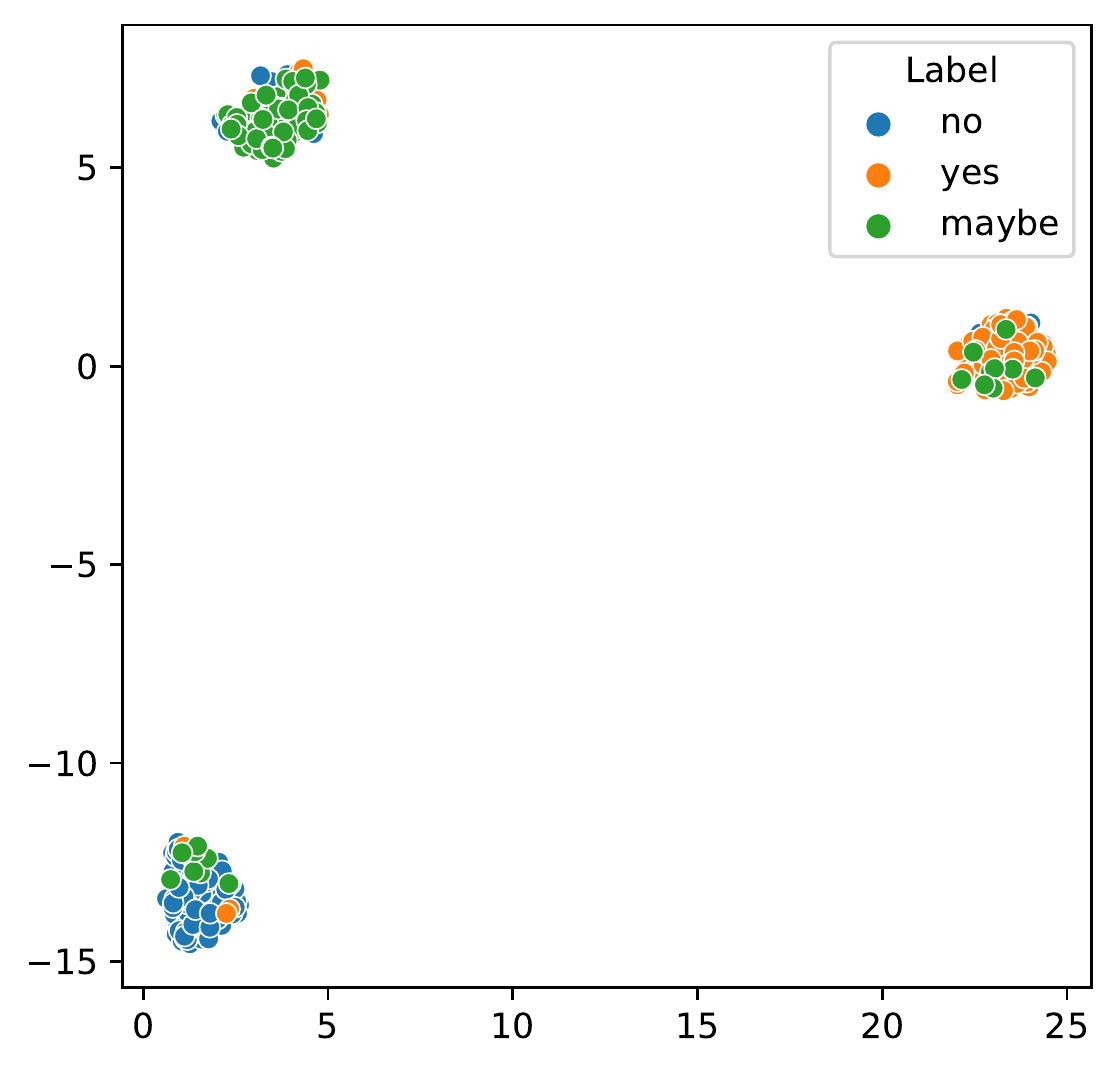}
        \caption{\textsc{SDE} $\alpha=0.1$}
        \label{fig:analyze-mnli-sde-0.1}
    \end{subfigure}
    \hfill
    \begin{subfigure}[b]{0.2\textwidth}
        \centering
        \includegraphics[width=\textwidth]{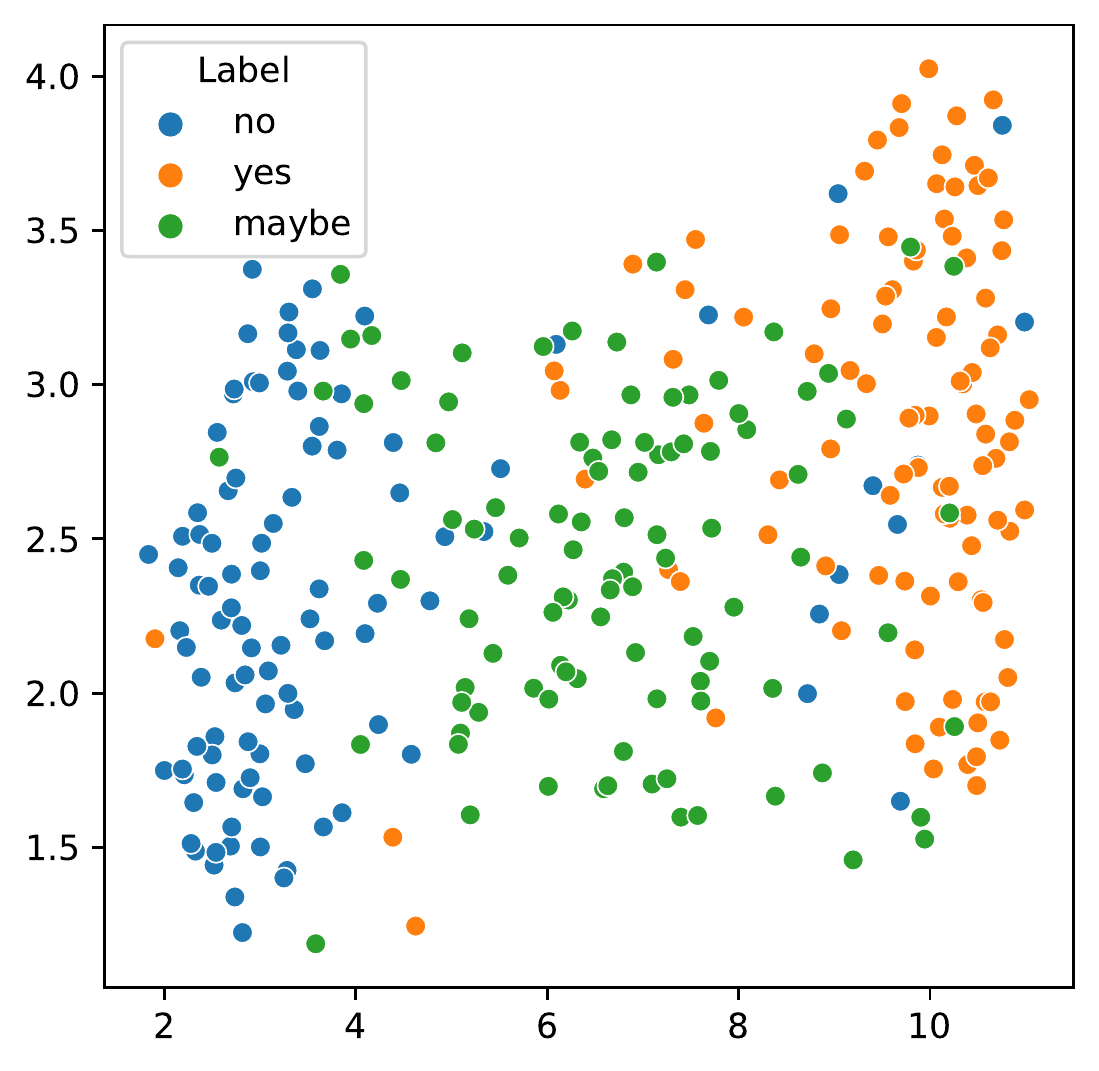}
        \caption{\textsc{PDF} $\alpha=0.01$}
        \label{fig:analyze-mnli-brown-0.01}
    \end{subfigure}
    \hfill
    \begin{subfigure}[b]{0.2\textwidth}
        \centering
        \includegraphics[width=\textwidth]{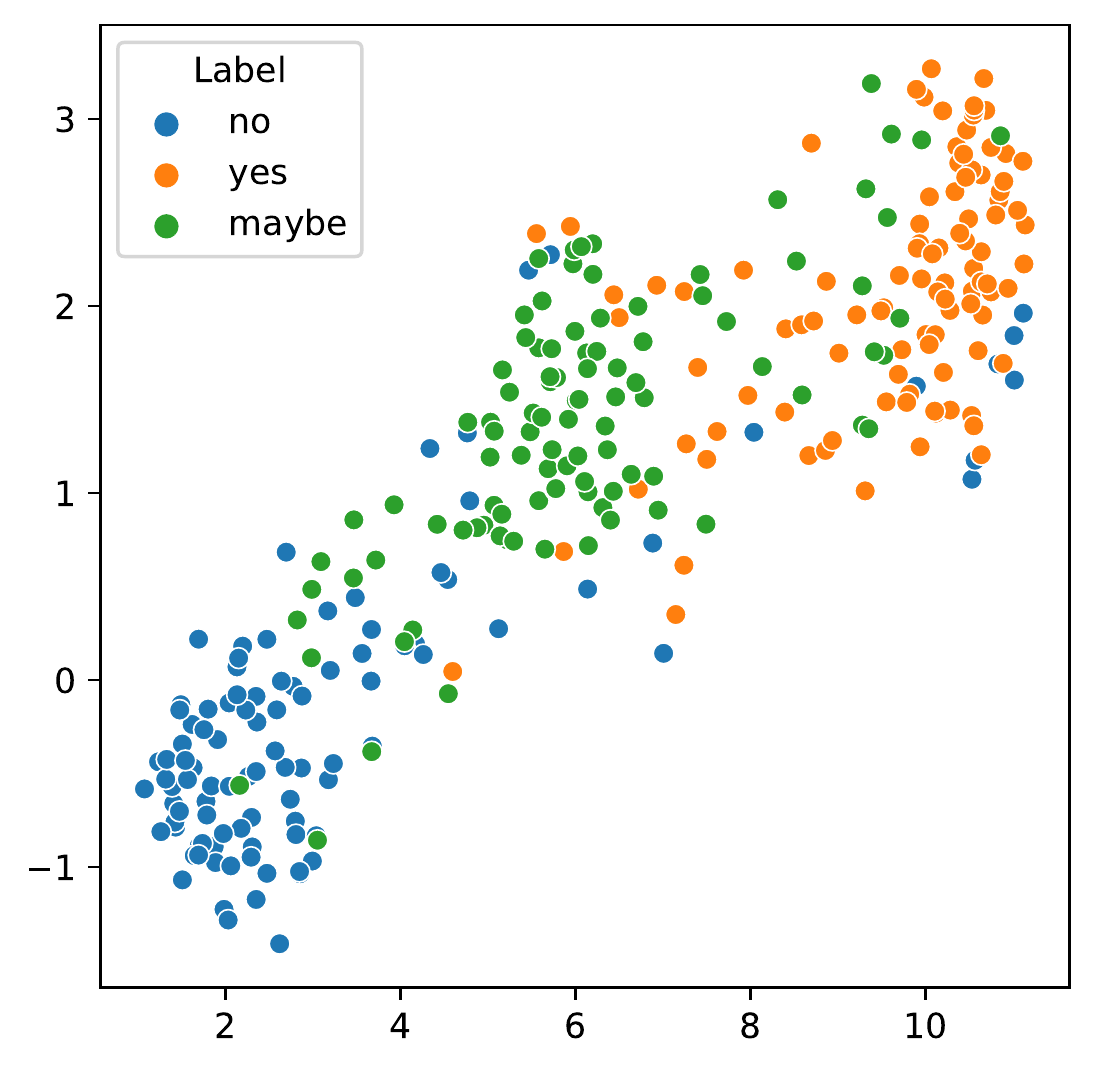}
        \caption{\textsc{PDF} $\alpha=0.1$}
        \label{fig:analyze-mnli-brown-0.1}
    \end{subfigure}
    \hfill
    \begin{subfigure}[b]{0.2\textwidth}
        \centering
        \includegraphics[width=\textwidth]{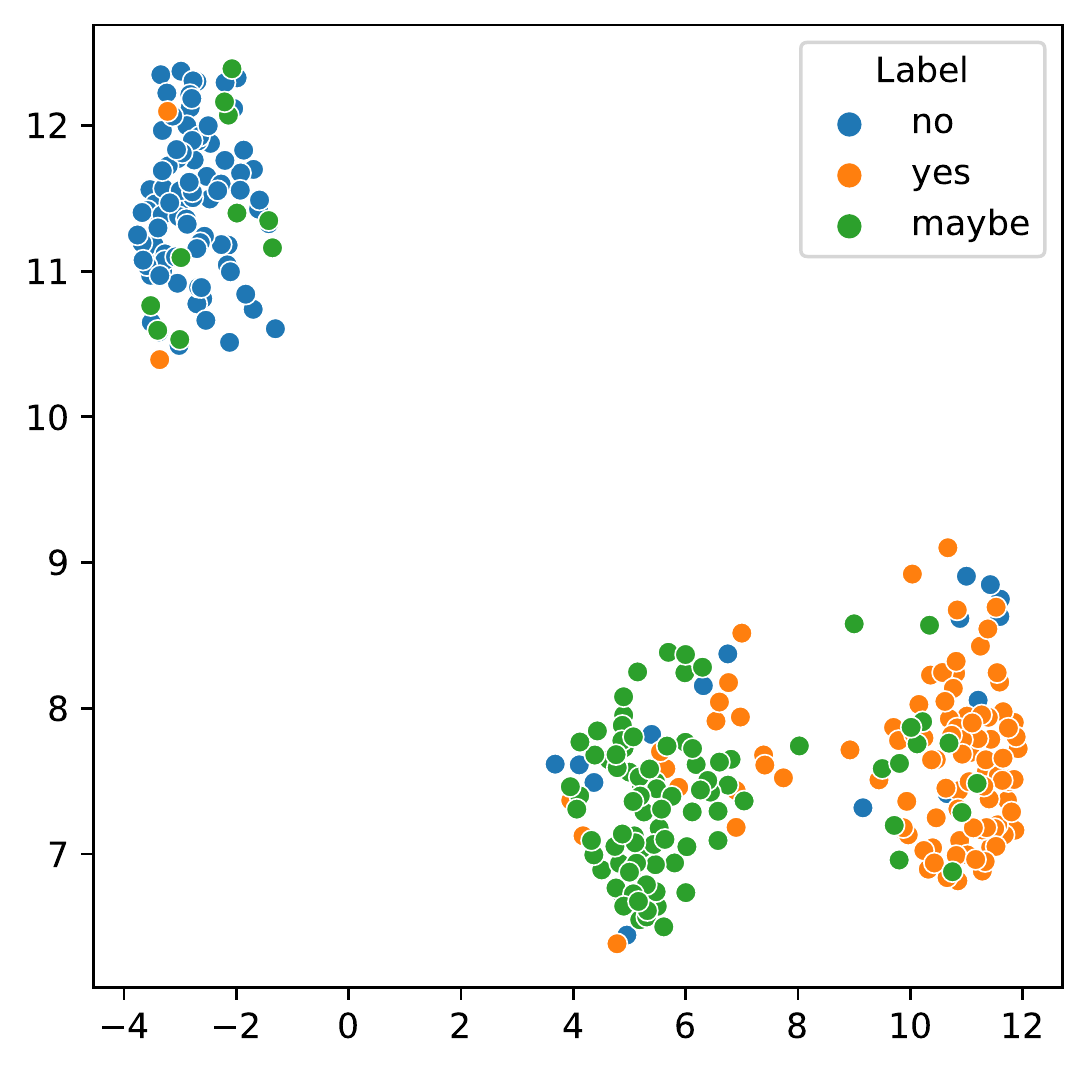}
        \caption{\textsc{PDF} $\alpha=1.0$}
        \label{fig:analyze-mnli-brown-1.0}
    \end{subfigure}
    \hfill
    \begin{subfigure}[b]{0.2\textwidth}
    \centering
    \renewcommand\arraystretch{1.1}
    \resizebox{\linewidth}{!}{
    \begin{tabular}{l r@{}l r@{}l}
        \toprule
        Dataset & \multicolumn{2}{c}{\textsc{PDF}} & \multicolumn{2}{c}{\textsc{SDE}}\\
        \midrule
        MNLI & 0&.928$^{***}$ & 0&.894$^{***}$ \\
        QQP & 0&.715$^{*}$ & 0&.897$^{***}$ \\
        QNLI & 0&.971$^{***}$ & 0&.837$^{**}$ \\
        SST-2 & 0&.966$^{***}$ & 0&.867$^{***}$ \\
        MRPC & 0&.226 & 0&.433 \\
        CoLA & 0&.633$^{*}$ & 0&.807$^{**}$ \\
        RTE & 0&.440 & 0&.589$^{*}$ \\
    \bottomrule
    \end{tabular}}
    \caption{Pearson's correlation}
    \label{tab:centroid-correlation}
    \end{subfigure}
    \caption{The visualization of the last layer's hidden states on MNLI using the prompt that is trained (a) without regularizer (b-d) with the \textsc{SDE} regularizer (e-g) with the \textsc{PDF} regularizer. And the table in (h) is the Pearson's correlation between the regularization strength $\alpha$ and the average distance between the centroids of different label clusters. $^{***}$: $\mathit{p}<.001$,  $^{**}$: $\mathit{p}<.01$, $^{*}$: $\mathit{p}<.05$.}
    \label{fig:analyze-mnli}
\end{figure*}
We randomly sample 100 samples for each label in MNLI, use UMAP~\citep{DBLP:journals/jossw/McInnesHSG18} to reduce the dimension of the last layer's hidden states of Prompt tuning and plot them in~\Figref{fig:analyze-mnli}. It shows clearly that for both regularizers, as the regularization intensity gets stronger, the hidden states of the last layer become more distinguishable among labels. By looking at the axes of these plots, we find that the distances between the clusters generally increase when the regularization intensity is increased. We also notice that the \textsc{SDE} regularizer better helps separate the hidden states of the last layer by substantially enlarging the distance between the centroids of different labels, which could be one of the reasons why the \textsc{SDE} regularizer has better effectiveness in almost all experiments. 

We also calculate the Pearson's correlation between the $\alpha$ and the average distance between the centroids of different clusters. The results are shown in Tab.~\subref{tab:centroid-correlation}. On all the datasets, the $\alpha$ has a positive correlation to the average centroid distance, and on most of the datasets, the correlations are significant ($\mathit{p}$-value <.05). This indicates that as the regularization intensity gets stronger, the centroids of different label clusters become more distant, which is a desired effect because the regularizer encourages the hidden states for different labels to conform to different latent bridges. 

\subsection{Hidden States Spontaneously Approach the Latent Bridges}
\label{sec:analysis-bridge-distance}
An interesting phenomenon we observe is that the vanilla PETs' intermediate hidden states spontaneously approach our latent bridges when they are projected by our mapping $g_\gamma$. That is, applying our mapping $g_\gamma$ to the hidden states of vanilla PETs, we find that when the performance of vanilla PETs becomes better, the average distance from $g_\gamma(\{\vh_o^{(\cdot)}, \bar{\vh}^{(\cdot)}\})$ to our latent bridge gets closer. Here, similar to~\citet{DBLP:conf/iclr/WangDGH22}, we define the distance from $g_\gamma(\vh_o^{(\cdot)}, \bar{\vh}^{(\cdot)})$ to its corresponding latent bridge $X_y$ using~\Eqref{eq:goodness-pdf} without the constant. Note that the vanilla PETs have no access to $g_\gamma$ and the latent bridges during the training process, and $g_\gamma$ also has no access to the PETs during its fitting.

We show the above phenomenon by conducting analyses in few-shot scenarios with \textsc{PDF} regularizer, and reporting in~\Tabref{tab:bridgeloss} the correlation between (1) the number of shots and the average distance from latent hidden states to latent bridges (2) the performance and the average distance from latent hidden states to latent bridges. We report Kendall's rank correlation
for (1), and Pearson's correlation for (2). 
See~\Appref{sec:appendix-correlation} for the detailed setup.

From~\Tabref{tab:bridgeloss}, the number of shots has a negative correlation to the distance, and the correlation is significant on 4 out of 6 datasets. This indicates that as the amount of available data increases for vanilla PETs, its intermediate hidden states in latent space spontaneously approach latent bridges even without knowing the mapping $g_\gamma$ and the bridges. Additionally, the results in~\Tabref{tab:bridgeloss} show the negative correlation between the performance of vanilla PETs and the distance to the latent bridges, and it is significant on 3 out of 6 datasets.

\begin{table}
    \centering
    \renewcommand{\arraystretch}{0.7}
    \resizebox{0.8\linewidth}{!}{
    \begin{tabular}{l r@{}l r@{}l r@{}l r@{}l}
        \toprule
        & \multicolumn{4}{c}{Dist-Shot} & \multicolumn{4}{c}{Dist-Perf}\\
        \cmidrule(lr){2-5}\cmidrule(lr){6-9}
        Dataset & \multicolumn{2}{c}{Coef.} & \multicolumn{2}{c}{$\mathit{p}$-value} & \multicolumn{2}{c}{Coef.} & \multicolumn{2}{c}{$\mathit{p}$-value} \\
        \midrule
        MNLI & -0&.39 & .026&$^{*}$ & -0&.09 & .715& \\
        QQP & -0&.36 & .037&$^{*}$ & -0&.42 & .062& \\
        QNLI & -0&.32 & .069& & -0&.77 & $<$.001&$^{***}$ \\
        SST-2 & -0&.50 & .005&$^{**}$ & -0&.69 & $<$.001&$^{***}$ \\
        MRPC & -0&.21 & .225& & -0&.13 & .591& \\
        RTE & -0&.34 & .049&$^{*}$ & -0&.87 & $<$.001&$^{***}$ \\
    \bottomrule
    \end{tabular}}
    \caption{Kendall's rank correlation between the number of shots and the distance to the latent bridges (Dist-Shot), and Pearson's correlation between the performance and the distance to the latent bridges (Dist-Perf).}
    \label{tab:bridgeloss}
\end{table}
Altogether, the two findings on correlation show that as the PETs' performance improve, their intermediate hidden states projected by $g_\gamma$ spontaneously approach our bridges in the latent space. This implies that there exists intrinsically diffusion-bridge-like latent dynamics for PETs, and also justifies our use of diffusion bridges as regularizers. 
\section{Related Works}
\label{sec:related-works}

Recent years have witnessed the success of PLMs~\citep{DBLP:journals/jmlr/RaffelSRLNMZLL20,DBLP:conf/nips/BrownMRSKDNSSAA20}.
However, as the sizes of PLMs continue to grow, it becomes increasingly impractical to perform fine-tuning on downstream tasks.
Many efforts have been devoted to PETs, aiming to tune only a few parameters rather than the whole PLM on downstream tasks. For example, Prompt tuning~\citep{DBLP:conf/emnlp/LesterAC21} prepends tunable embeddings to the input, Adapter~\citep{DBLP:conf/icml/HoulsbyGJMLGAG19} inserts small modules into each layer, BitFit~\citep{DBLP:conf/acl/ZakenGR22} tunes only the bias terms, and LoRA~\citep{DBLP:conf/iclr/HuSWALWWC22} decomposes the weight updates into low-rank matrices. 
In this paper, based on the theoretical grounding of PETs on optimal control~\citep{DBLP:conf/iclr/YangL22a,DBLP:journals/corr/abs-2203-06904}, we propose stochastic bridges as the regularizer on intermediate hidden states and introduce regularized PETs.

Our work also closely relates to continuous-time neural differential equations (NDEs)~\citep{DBLP:conf/nips/ChenRBD18,DBLP:conf/nips/RubanovaCD19,DBLP:conf/aabi/LiWCD19,DBLP:conf/icml/KidgerFLL21}. Continuous-time NDEs model the dynamics of the hidden states with ODEs or SDEs parameterized by neural networks. 
Inspired by these works, we use SDEs to represent the latent dynamics of PETs in the latent space. Our work differs from them in that we focus on using neural SDEs as regularizers for intermediate hidden states, rather than feature extractors.
We also notice that~\citet{DBLP:conf/iclr/WangDGH22} explore the use of Brownian bridge in regularizing the model dynamics across time. Our work differs from theirs in that we regularize the dynamics of intermediate hidden states across model layers. We additionally show that other diffusion bridges can be easily applied as the regularizer. As far as we know, we are the first to show the diffusion-bridge-like dynamics for hidden states across PLM layers and use diffusion bridges as regularizers on intermediate hidden states.
\section{Conclusion}
\label{sec:conclusion}
Starting from the optimal control perspective of PETs, we notice that the existing PETs lack a running cost that regularizes the intermediate hidden states. We thus propose to use stochastic bridges in a latent space as the regularizers for PETs. 
Experimental results on different models, tasks and PETs show that the proposed regularizers effectively improve the PETs' performance. Our analyses further show that the hidden states of the vanilla PETs spontaneously approach our diffusion bridges, indicating that there may exists intrinsically diffusion-bridge-like dynamics in PETs. As the first work using stochastic bridges as regularizers, we show its effectiveness and generality. We believe it will be a promising direction.
\section{Limitations}
\label{sec:limitations}

Introducing the regularizers inevitably incurs additional computational cost in the training of PETs. To show their impact on the training speed, we plot the time-performance curves for both \textsc{PDF} and \textsc{SDE} regularizers on full-set GLUE in~\Quadfigref{fig:speed_prompt}{fig:speed_lora}{fig:speed_bias}{fig:speed_adapter}.

On different PETs, the regularized PETs with \textsc{PDF} regularizer has similar running time to the vanilla PETs. On the two large datasets, QQP and MNLI, regularized PETs with \textsc{SDE} regularizer take about 2 to 3 times longer to achieve the best performance than vanilla PETs. However, on medium-sized (QNLI, SST-2) and small datasets (CoLA, MRPC, RTE), the time to achieve the best results with \textsc{SDE} regularizer is comparable to vanilla PETs.

Overall, the \textsc{PDF} regularizer can effectively improve the performance of PETs without introducing much computational cost. In scenarios where there is relatively more focus on the inference performance of PETs and less concern about the slightly longer training time, or when the dataset is small, \textsc{SDE} regularizer should be a good choice.

Our method does not introduce additional risk to the original risks of PETs.
\section*{Acknowledgement}
This work is supported by the National Natural Science Foundation of China (No. 62236004, No. 62236011), Institute Guo Qiang at Tsinghua University.

\bibliography{iclr2023_conference}

\begin{thebibliography}{28}
\expandafter\ifx\csname natexlab\endcsname\relax\def\natexlab#1{#1}\fi

\bibitem[{Brown et~al.(2020)Brown, Mann, Ryder, Subbiah, Kaplan, Dhariwal,
  Neelakantan, Shyam, Sastry, Askell, Agarwal, Herbert{-}Voss, Krueger,
  Henighan, Child, Ramesh, Ziegler, Wu, Winter, Hesse, Chen, Sigler, Litwin,
  Gray, Chess, Clark, Berner, McCandlish, Radford, Sutskever, and
  Amodei}]{DBLP:conf/nips/BrownMRSKDNSSAA20}
Tom~B. Brown, Benjamin Mann, Nick Ryder, Melanie Subbiah, Jared Kaplan,
  Prafulla Dhariwal, Arvind Neelakantan, Pranav Shyam, Girish Sastry, Amanda
  Askell, Sandhini Agarwal, Ariel Herbert{-}Voss, Gretchen Krueger, Tom
  Henighan, Rewon Child, Aditya Ramesh, Daniel~M. Ziegler, Jeffrey Wu, Clemens
  Winter, Christopher Hesse, Mark Chen, Eric Sigler, Mateusz Litwin, Scott
  Gray, Benjamin Chess, Jack Clark, Christopher Berner, Sam McCandlish, Alec
  Radford, Ilya Sutskever, and Dario Amodei. 2020.
\newblock \href
  {https://proceedings.neurips.cc/paper/2020/hash/1457c0d6bfcb4967418bfb8ac142f64a-Abstract.html}
  {Language models are few-shot learners}.
\newblock In \emph{Advances in Neural Information Processing Systems 33: Annual
  Conference on Neural Information Processing Systems 2020, NeurIPS 2020,
  December 6-12, 2020, virtual}.

\bibitem[{Chen et~al.(2018)Chen, Rubanova, Bettencourt, and
  Duvenaud}]{DBLP:conf/nips/ChenRBD18}
Tian~Qi Chen, Yulia Rubanova, Jesse Bettencourt, and David Duvenaud. 2018.
\newblock \href
  {https://proceedings.neurips.cc/paper/2018/hash/69386f6bb1dfed68692a24c8686939b9-Abstract.html}
  {Neural ordinary differential equations}.
\newblock In \emph{Advances in Neural Information Processing Systems 31: Annual
  Conference on Neural Information Processing Systems 2018, NeurIPS 2018,
  December 3-8, 2018, Montr{\'{e}}al, Canada}, pages 6572--6583.

\bibitem[{Devlin et~al.(2019)Devlin, Chang, Lee, and
  Toutanova}]{DBLP:conf/naacl/DevlinCLT19}
Jacob Devlin, Ming{-}Wei Chang, Kenton Lee, and Kristina Toutanova. 2019.
\newblock \href {https://doi.org/10.18653/v1/n19-1423} {{BERT:} pre-training of
  deep bidirectional transformers for language understanding}.
\newblock In \emph{Proceedings of the 2019 Conference of the North American
  Chapter of the Association for Computational Linguistics: Human Language
  Technologies, {NAACL-HLT} 2019, Minneapolis, MN, USA, June 2-7, 2019, Volume
  1 (Long and Short Papers)}, pages 4171--4186. Association for Computational
  Linguistics.

\bibitem[{Ding et~al.(2022)Ding, Qin, Yang, Wei, Yang, Su, Hu, Chen, Chan,
  Chen, Yi, Zhao, Wang, Liu, Zheng, Chen, Liu, Tang, Li, and
  Sun}]{DBLP:journals/corr/abs-2203-06904}
Ning Ding, Yujia Qin, Guang Yang, Fuchao Wei, Zonghan Yang, Yusheng Su,
  Shengding Hu, Yulin Chen, Chi{-}Min Chan, Weize Chen, Jing Yi, Weilin Zhao,
  Xiaozhi Wang, Zhiyuan Liu, Hai{-}Tao Zheng, Jianfei Chen, Yang Liu, Jie Tang,
  Juanzi Li, and Maosong Sun. 2022.
\newblock \href {https://doi.org/10.48550/arXiv.2203.06904} {Delta tuning: {A}
  comprehensive study of parameter efficient methods for pre-trained language
  models}.
\newblock \emph{CoRR}, abs/2203.06904.

\bibitem[{Dosovitskiy et~al.(2021)Dosovitskiy, Beyer, Kolesnikov, Weissenborn,
  Zhai, Unterthiner, Dehghani, Minderer, Heigold, Gelly, Uszkoreit, and
  Houlsby}]{DBLP:conf/iclr/DosovitskiyB0WZ21}
Alexey Dosovitskiy, Lucas Beyer, Alexander Kolesnikov, Dirk Weissenborn,
  Xiaohua Zhai, Thomas Unterthiner, Mostafa Dehghani, Matthias Minderer, Georg
  Heigold, Sylvain Gelly, Jakob Uszkoreit, and Neil Houlsby. 2021.
\newblock \href {https://openreview.net/forum?id=YicbFdNTTy} {An image is worth
  16x16 words: Transformers for image recognition at scale}.
\newblock In \emph{9th International Conference on Learning Representations,
  {ICLR} 2021, Virtual Event, Austria, May 3-7, 2021}. OpenReview.net.

\bibitem[{Dupont et~al.(2019)Dupont, Doucet, and
  Teh}]{DBLP:conf/nips/DupontDT19}
Emilien Dupont, Arnaud Doucet, and Yee~Whye Teh. 2019.
\newblock \href
  {https://proceedings.neurips.cc/paper/2019/hash/21be9a4bd4f81549a9d1d241981cec3c-Abstract.html}
  {Augmented neural odes}.
\newblock In \emph{Advances in Neural Information Processing Systems 32: Annual
  Conference on Neural Information Processing Systems 2019, NeurIPS 2019,
  December 8-14, 2019, Vancouver, BC, Canada}, pages 3134--3144.

\bibitem[{Gao et~al.(2021)Gao, Fisch, and Chen}]{DBLP:conf/acl/GaoFC20}
Tianyu Gao, Adam Fisch, and Danqi Chen. 2021.
\newblock \href {https://doi.org/10.18653/v1/2021.acl-long.295} {Making
  pre-trained language models better few-shot learners}.
\newblock In \emph{Proceedings of the 59th Annual Meeting of the Association
  for Computational Linguistics and the 11th International Joint Conference on
  Natural Language Processing, {ACL/IJCNLP} 2021, (Volume 1: Long Papers),
  Virtual Event, August 1-6, 2021}, pages 3816--3830. Association for
  Computational Linguistics.

\bibitem[{He et~al.(2021)He, Liu, Gao, and Chen}]{DBLP:conf/iclr/HeLGC21}
Pengcheng He, Xiaodong Liu, Jianfeng Gao, and Weizhu Chen. 2021.
\newblock \href {https://openreview.net/forum?id=XPZIaotutsD} {Deberta:
  decoding-enhanced bert with disentangled attention}.
\newblock In \emph{9th International Conference on Learning Representations,
  {ICLR} 2021, Virtual Event, Austria, May 3-7, 2021}. OpenReview.net.

\bibitem[{Houlsby et~al.(2019)Houlsby, Giurgiu, Jastrzebski, Morrone,
  de~Laroussilhe, Gesmundo, Attariyan, and
  Gelly}]{DBLP:conf/icml/HoulsbyGJMLGAG19}
Neil Houlsby, Andrei Giurgiu, Stanislaw Jastrzebski, Bruna Morrone, Quentin
  de~Laroussilhe, Andrea Gesmundo, Mona Attariyan, and Sylvain Gelly. 2019.
\newblock \href {http://proceedings.mlr.press/v97/houlsby19a.html}
  {Parameter-efficient transfer learning for {NLP}}.
\newblock In \emph{Proceedings of the 36th International Conference on Machine
  Learning, {ICML} 2019, 9-15 June 2019, Long Beach, California, {USA}},
  volume~97 of \emph{Proceedings of Machine Learning Research}, pages
  2790--2799. {PMLR}.

\bibitem[{Hu et~al.(2022)Hu, Shen, Wallis, Allen{-}Zhu, Li, Wang, Wang, and
  Chen}]{DBLP:conf/iclr/HuSWALWWC22}
Edward~J. Hu, Yelong Shen, Phillip Wallis, Zeyuan Allen{-}Zhu, Yuanzhi Li,
  Shean Wang, Lu~Wang, and Weizhu Chen. 2022.
\newblock \href {https://openreview.net/forum?id=nZeVKeeFYf9} {Lora: Low-rank
  adaptation of large language models}.
\newblock In \emph{The Tenth International Conference on Learning
  Representations, {ICLR} 2022, Virtual Event, April 25-29, 2022}.
  OpenReview.net.

\bibitem[{Kidger et~al.(2021)Kidger, Foster, Li, and
  Lyons}]{DBLP:conf/icml/KidgerFLL21}
Patrick Kidger, James Foster, Xuechen Li, and Terry~J. Lyons. 2021.
\newblock \href {http://proceedings.mlr.press/v139/kidger21b.html} {Neural sdes
  as infinite-dimensional gans}.
\newblock In \emph{Proceedings of the 38th International Conference on Machine
  Learning, {ICML} 2021, 18-24 July 2021, Virtual Event}, volume 139 of
  \emph{Proceedings of Machine Learning Research}, pages 5453--5463. {PMLR}.

\bibitem[{Lester et~al.(2021)Lester, Al{-}Rfou, and
  Constant}]{DBLP:conf/emnlp/LesterAC21}
Brian Lester, Rami Al{-}Rfou, and Noah Constant. 2021.
\newblock \href {https://doi.org/10.18653/v1/2021.emnlp-main.243} {The power of
  scale for parameter-efficient prompt tuning}.
\newblock In \emph{Proceedings of the 2021 Conference on Empirical Methods in
  Natural Language Processing, {EMNLP} 2021, Virtual Event / Punta Cana,
  Dominican Republic, 7-11 November, 2021}, pages 3045--3059. Association for
  Computational Linguistics.

\bibitem[{Li and Liang(2021)}]{DBLP:conf/acl/LiL20}
Xiang~Lisa Li and Percy Liang. 2021.
\newblock \href {https://doi.org/10.18653/v1/2021.acl-long.353} {Prefix-tuning:
  Optimizing continuous prompts for generation}.
\newblock In \emph{Proceedings of the 59th Annual Meeting of the Association
  for Computational Linguistics and the 11th International Joint Conference on
  Natural Language Processing, {ACL/IJCNLP} 2021, (Volume 1: Long Papers),
  Virtual Event, August 1-6, 2021}, pages 4582--4597. Association for
  Computational Linguistics.

\bibitem[{Li et~al.(2019)Li, Wong, Chen, and Duvenaud}]{DBLP:conf/aabi/LiWCD19}
Xuechen Li, Ting{-}Kam~Leonard Wong, Ricky T.~Q. Chen, and David Duvenaud.
  2019.
\newblock \href {http://proceedings.mlr.press/v118/li20a.html} {Scalable
  gradients and variational inference for stochastic differential equations}.
\newblock In \emph{Symposium on Advances in Approximate Bayesian Inference,
  {AABI} 2019, Vancouver, BC, Canada, December 8, 2019}, volume 118 of
  \emph{Proceedings of Machine Learning Research}, pages 1--28. {PMLR}.

\bibitem[{Li et~al.(2020)Li, Wong, Chen, and
  Duvenaud}]{DBLP:conf/aistats/LiWCD20}
Xuechen Li, Ting{-}Kam~Leonard Wong, Ricky T.~Q. Chen, and David Duvenaud.
  2020.
\newblock \href {http://proceedings.mlr.press/v108/li20i.html} {Scalable
  gradients for stochastic differential equations}.
\newblock In \emph{The 23rd International Conference on Artificial Intelligence
  and Statistics, {AISTATS} 2020, 26-28 August 2020, Online [Palermo, Sicily,
  Italy]}, volume 108 of \emph{Proceedings of Machine Learning Research}, pages
  3870--3882. {PMLR}.

\bibitem[{McInnes et~al.(2018)McInnes, Healy, Saul, and
  Gro{\ss}berger}]{DBLP:journals/jossw/McInnesHSG18}
Leland McInnes, John Healy, Nathaniel Saul, and Lukas Gro{\ss}berger. 2018.
\newblock \href {https://doi.org/10.21105/joss.00861} {{UMAP:} uniform manifold
  approximation and projection}.
\newblock \emph{J. Open Source Softw.}, 3(29):861.

\bibitem[{Raffel et~al.(2020)Raffel, Shazeer, Roberts, Lee, Narang, Matena,
  Zhou, Li, and Liu}]{DBLP:journals/jmlr/RaffelSRLNMZLL20}
Colin Raffel, Noam Shazeer, Adam Roberts, Katherine Lee, Sharan Narang, Michael
  Matena, Yanqi Zhou, Wei Li, and Peter~J. Liu. 2020.
\newblock \href {http://jmlr.org/papers/v21/20-074.html} {Exploring the limits
  of transfer learning with a unified text-to-text transformer}.
\newblock \emph{J. Mach. Learn. Res.}, 21:140:1--140:67.

\bibitem[{Rubanova et~al.(2019)Rubanova, Chen, and
  Duvenaud}]{DBLP:conf/nips/RubanovaCD19}
Yulia Rubanova, Tian~Qi Chen, and David Duvenaud. 2019.
\newblock \href
  {https://proceedings.neurips.cc/paper/2019/hash/42a6845a557bef704ad8ac9cb4461d43-Abstract.html}
  {Latent ordinary differential equations for irregularly-sampled time series}.
\newblock In \emph{Advances in Neural Information Processing Systems 32: Annual
  Conference on Neural Information Processing Systems 2019, NeurIPS 2019,
  December 8-14, 2019, Vancouver, BC, Canada}, pages 5321--5331.

\bibitem[{Shoeybi et~al.(2019)Shoeybi, Patwary, Puri, LeGresley, Casper, and
  Catanzaro}]{DBLP:journals/corr/abs-1909-08053}
Mohammad Shoeybi, Mostofa Patwary, Raul Puri, Patrick LeGresley, Jared Casper,
  and Bryan Catanzaro. 2019.
\newblock \href {http://arxiv.org/abs/1909.08053} {Megatron-lm: Training
  multi-billion parameter language models using model parallelism}.
\newblock \emph{CoRR}, abs/1909.08053.

\bibitem[{Smith et~al.(2022)Smith, Patwary, Norick, LeGresley, Rajbhandari,
  Casper, Liu, Prabhumoye, Zerveas, Korthikanti, Zheng, Child, Aminabadi,
  Bernauer, Song, Shoeybi, He, Houston, Tiwary, and
  Catanzaro}]{DBLP:journals/corr/abs-2201-11990}
Shaden Smith, Mostofa Patwary, Brandon Norick, Patrick LeGresley, Samyam
  Rajbhandari, Jared Casper, Zhun Liu, Shrimai Prabhumoye, George Zerveas,
  Vijay Korthikanti, Elton Zheng, Rewon Child, Reza~Yazdani Aminabadi, Julie
  Bernauer, Xia Song, Mohammad Shoeybi, Yuxiong He, Michael Houston, Saurabh
  Tiwary, and Bryan Catanzaro. 2022.
\newblock \href {http://arxiv.org/abs/2201.11990} {Using deepspeed and megatron
  to train megatron-turing {NLG} 530b, {A} large-scale generative language
  model}.
\newblock \emph{CoRR}, abs/2201.11990.

\bibitem[{Sobczyk(2013)}]{sobczyk2013stochastic}
K~Sobczyk. 2013.
\newblock \emph{Stochastic Differential Equations: With Applications to Physics
  and Engineering}, volume~40.
\newblock Springer Science \& Business Media.

\bibitem[{Wang et~al.(2019)Wang, Singh, Michael, Hill, Levy, and
  Bowman}]{DBLP:conf/iclr/WangSMHLB19}
Alex Wang, Amanpreet Singh, Julian Michael, Felix Hill, Omer Levy, and
  Samuel~R. Bowman. 2019.
\newblock \href {https://openreview.net/forum?id=rJ4km2R5t7} {{GLUE:} {A}
  multi-task benchmark and analysis platform for natural language
  understanding}.
\newblock In \emph{7th International Conference on Learning Representations,
  {ICLR} 2019, New Orleans, LA, USA, May 6-9, 2019}. OpenReview.net.

\bibitem[{Wang et~al.(2022)Wang, Durmus, Goodman, and
  Hashimoto}]{DBLP:conf/iclr/WangDGH22}
Rose~E. Wang, Esin Durmus, Noah~D. Goodman, and Tatsunori Hashimoto. 2022.
\newblock \href {https://openreview.net/forum?id=pMQwKL1yctf} {Language
  modeling via stochastic processes}.
\newblock In \emph{The Tenth International Conference on Learning
  Representations, {ICLR} 2022, Virtual Event, April 25-29, 2022}.
  OpenReview.net.

\bibitem[{Wang and Sloan(2011)}]{DBLP:journals/ior/WangS11}
Xiaoqun Wang and Ian~H. Sloan. 2011.
\newblock \href {https://doi.org/10.1287/opre.1100.0853} {Quasi-monte carlo
  methods in financial engineering: An equivalence principle and dimension
  reduction}.
\newblock \emph{Oper. Res.}, 59(1):80--95.

\bibitem[{Yang and Liu(2022)}]{DBLP:conf/iclr/YangL22a}
Zonghan Yang and Yang Liu. 2022.
\newblock \href {https://openreview.net/forum?id=eBCmOocUejf} {On robust
  prefix-tuning for text classification}.
\newblock In \emph{The Tenth International Conference on Learning
  Representations, {ICLR} 2022, Virtual Event, April 25-29, 2022}.
  OpenReview.net.

\bibitem[{Zaken et~al.(2022)Zaken, Goldberg, and
  Ravfogel}]{DBLP:conf/acl/ZakenGR22}
Elad~Ben Zaken, Yoav Goldberg, and Shauli Ravfogel. 2022.
\newblock \href {https://doi.org/10.18653/v1/2022.acl-short.1} {Bitfit: Simple
  parameter-efficient fine-tuning for transformer-based masked
  language-models}.
\newblock In \emph{Proceedings of the 60th Annual Meeting of the Association
  for Computational Linguistics (Volume 2: Short Papers), {ACL} 2022, Dublin,
  Ireland, May 22-27, 2022}, pages 1--9. Association for Computational
  Linguistics.

\bibitem[{Zhang et~al.(2020)Zhang, Gao, Unterman, and
  Arodz}]{DBLP:conf/icml/ZhangGUA20}
Han Zhang, Xi~Gao, Jacob Unterman, and Tom Arodz. 2020.
\newblock \href {http://proceedings.mlr.press/v119/zhang20h.html}
  {Approximation capabilities of neural odes and invertible residual networks}.
\newblock In \emph{Proceedings of the 37th International Conference on Machine
  Learning, {ICML} 2020, 13-18 July 2020, Virtual Event}, volume 119 of
  \emph{Proceedings of Machine Learning Research}, pages 11086--11095. {PMLR}.

\bibitem[{Zhu et~al.(2015)Zhu, Kiros, Zemel, Salakhutdinov, Urtasun, Torralba,
  and Fidler}]{DBLP:conf/iccv/ZhuKZSUTF15}
Yukun Zhu, Ryan Kiros, Richard~S. Zemel, Ruslan Salakhutdinov, Raquel Urtasun,
  Antonio Torralba, and Sanja Fidler. 2015.
\newblock \href {https://doi.org/10.1109/ICCV.2015.11} {Aligning books and
  movies: Towards story-like visual explanations by watching movies and reading
  books}.
\newblock In \emph{2015 {IEEE} International Conference on Computer Vision,
  {ICCV} 2015, Santiago, Chile, December 7-13, 2015}, pages 19--27. {IEEE}
  Computer Society.

\end{thebibliography}
\bibliographystyle{acl_natbib}

\appendix

\begin{figure*}[t!]
    \centering
    \begin{subfigure}[b]{0.49\textwidth}
        \centering
        \includegraphics[width=\textwidth]{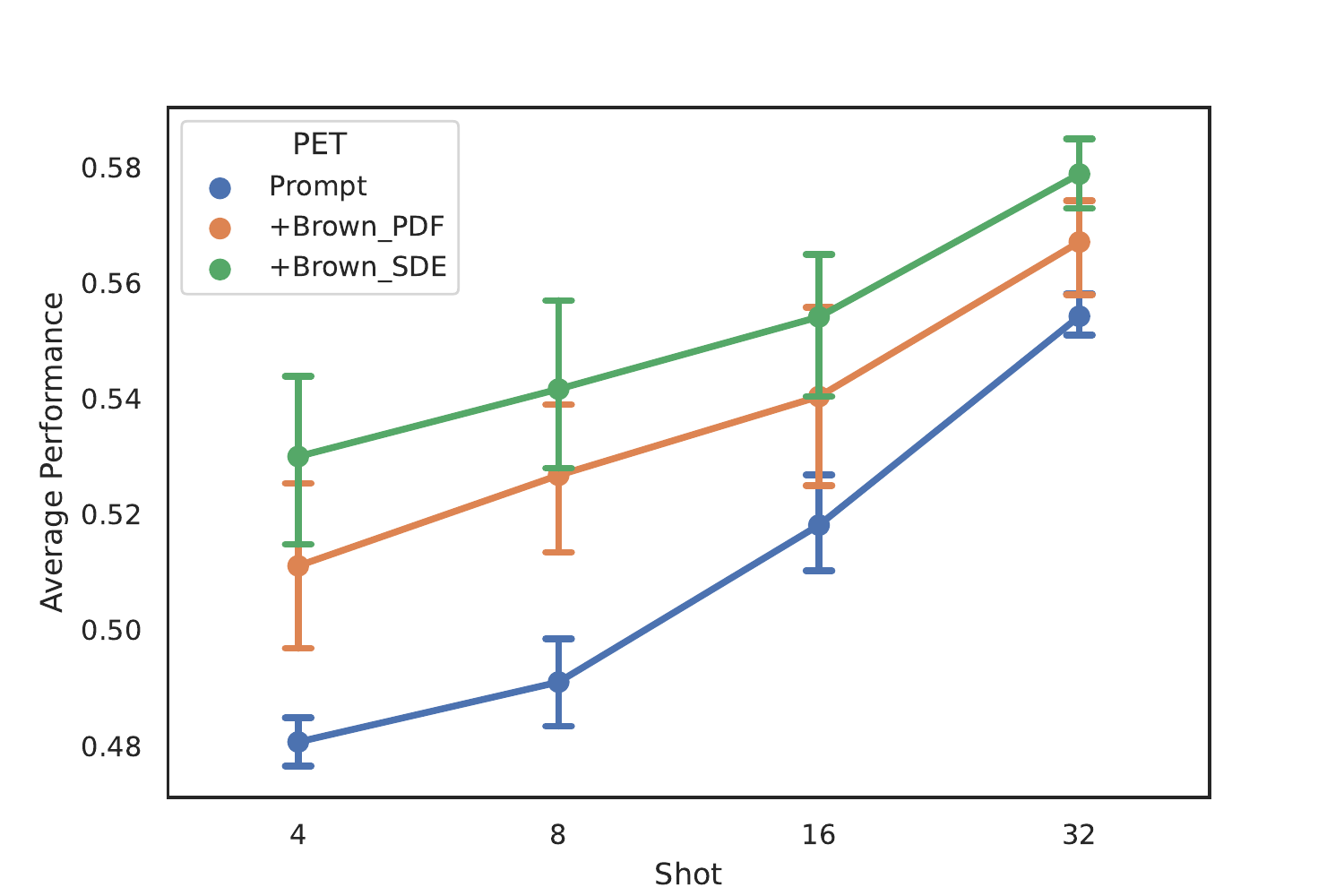}
        \caption{\textsc{Prompt Tuning}}
        \label{fig:shot-perf-prompt}
    \end{subfigure}
    \hfill
    \begin{subfigure}[b]{0.49\textwidth}
        \centering
        \includegraphics[width=\textwidth]{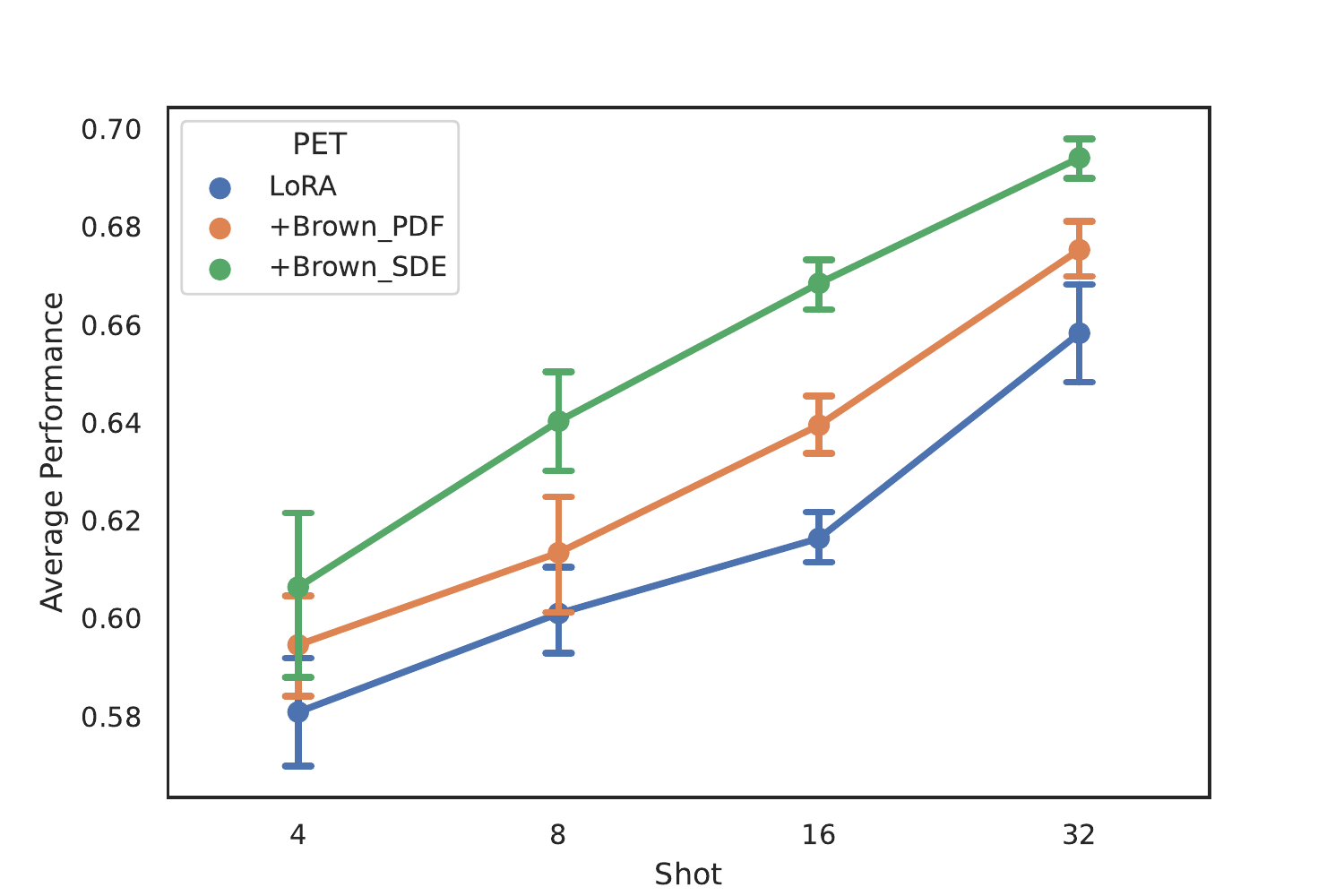}
        \caption{\textsc{LoRA}}
        \label{fig:shot-perf-lora}
    \end{subfigure}
    \hfill
    \begin{subfigure}[b]{0.49\textwidth}
        \centering
        \includegraphics[width=\textwidth]{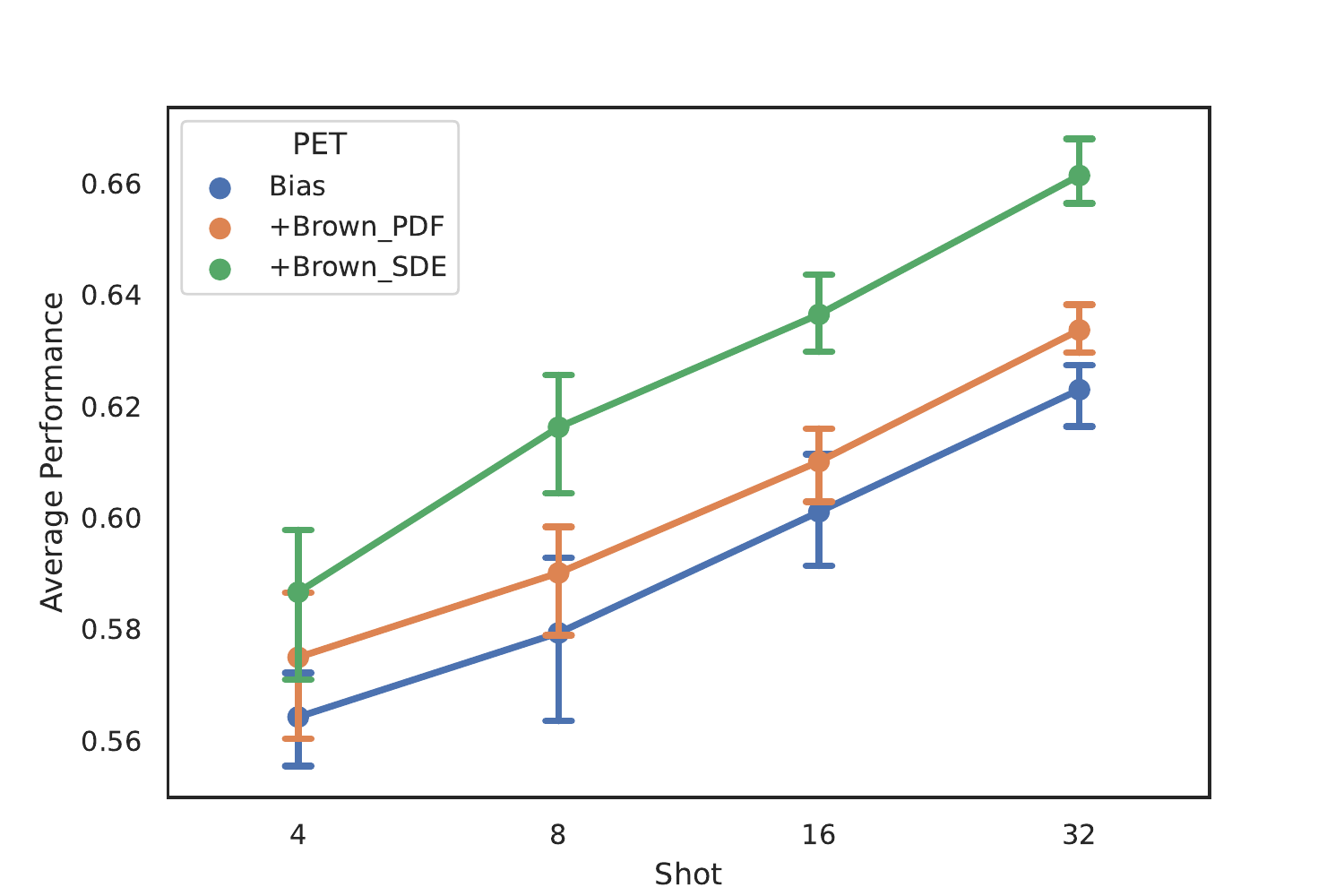}
        \caption{\textsc{Bias Tuning}}
        \label{fig:shot-perf-bias}
    \end{subfigure}
    \hfill
    \begin{subfigure}[b]{0.49\textwidth}
        \centering
        \includegraphics[width=\textwidth]{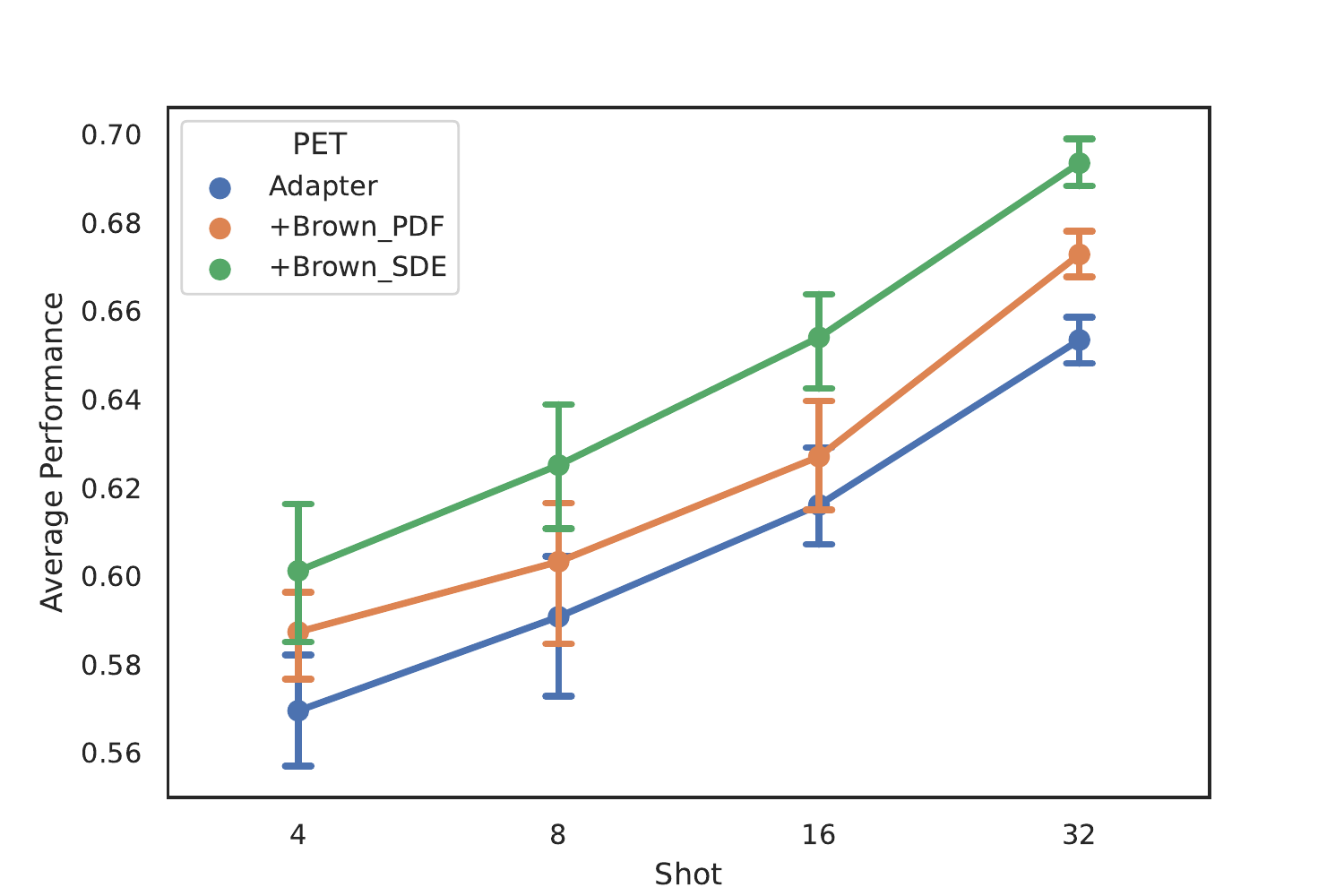}
        \caption{\textsc{Adapter}}
        \label{fig:shot-perf-adapter}
    \end{subfigure}
    \caption{The average $\megatronbert$ few-shot GLUE results trained with different PETs under different shots. The results are averaged across 5 different seeds and the error bars indicate the 95\% confidence. \textsc{SDE} regularizer consistently outperforms the baseline \textsc{PDF} regularizer.}
    \label{fig:shot-perf}
\end{figure*}
\begin{figure*}[t!]
    \centering
    \begin{subfigure}[b]{0.49\textwidth}
        \centering
        \includegraphics[width=\textwidth]{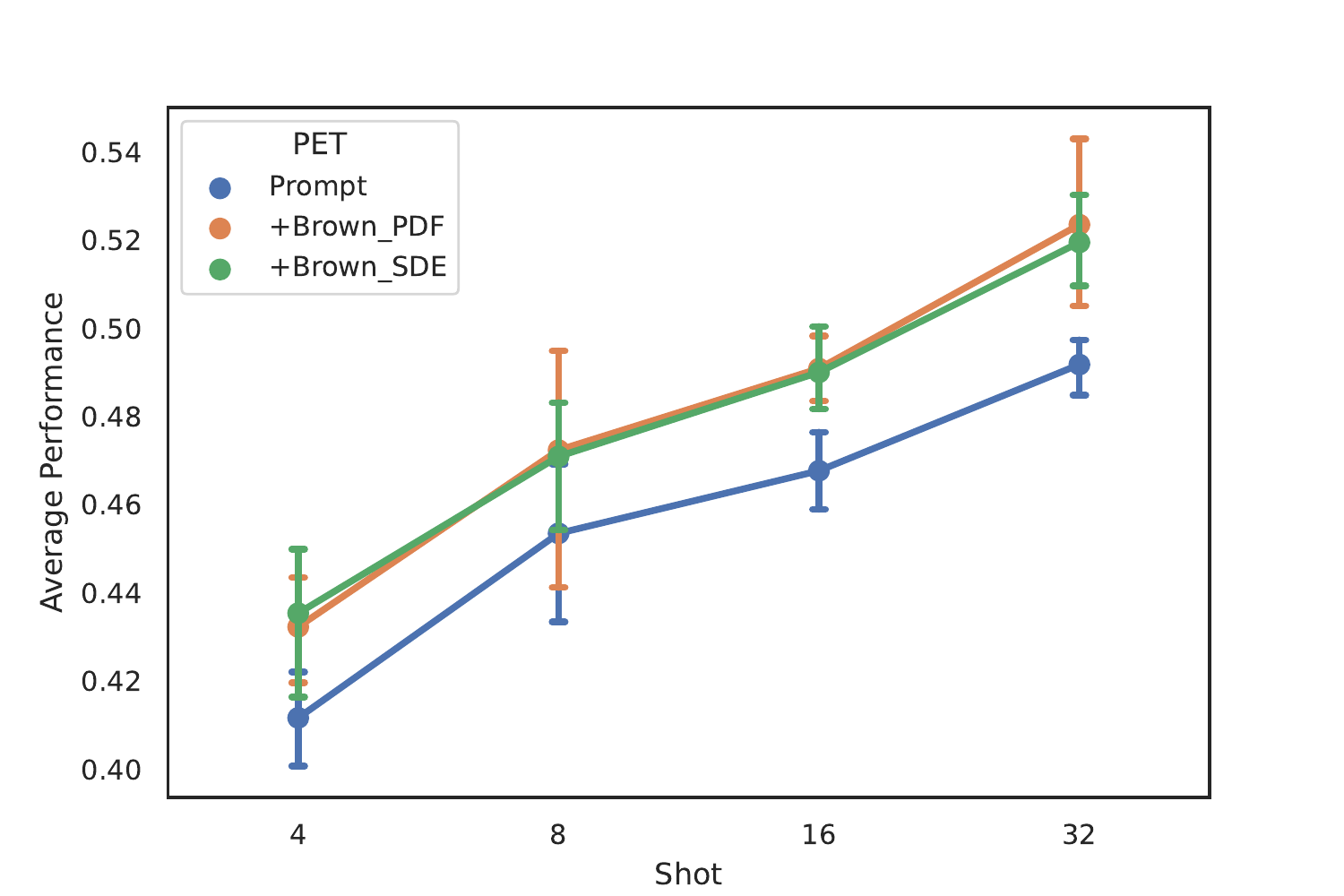}
        \caption{\textsc{Prompt Tuning}}
        \label{fig:deberta-shot-perf-prompt}
    \end{subfigure}
    \hfill
    \begin{subfigure}[b]{0.49\textwidth}
        \centering
        \includegraphics[width=\textwidth]{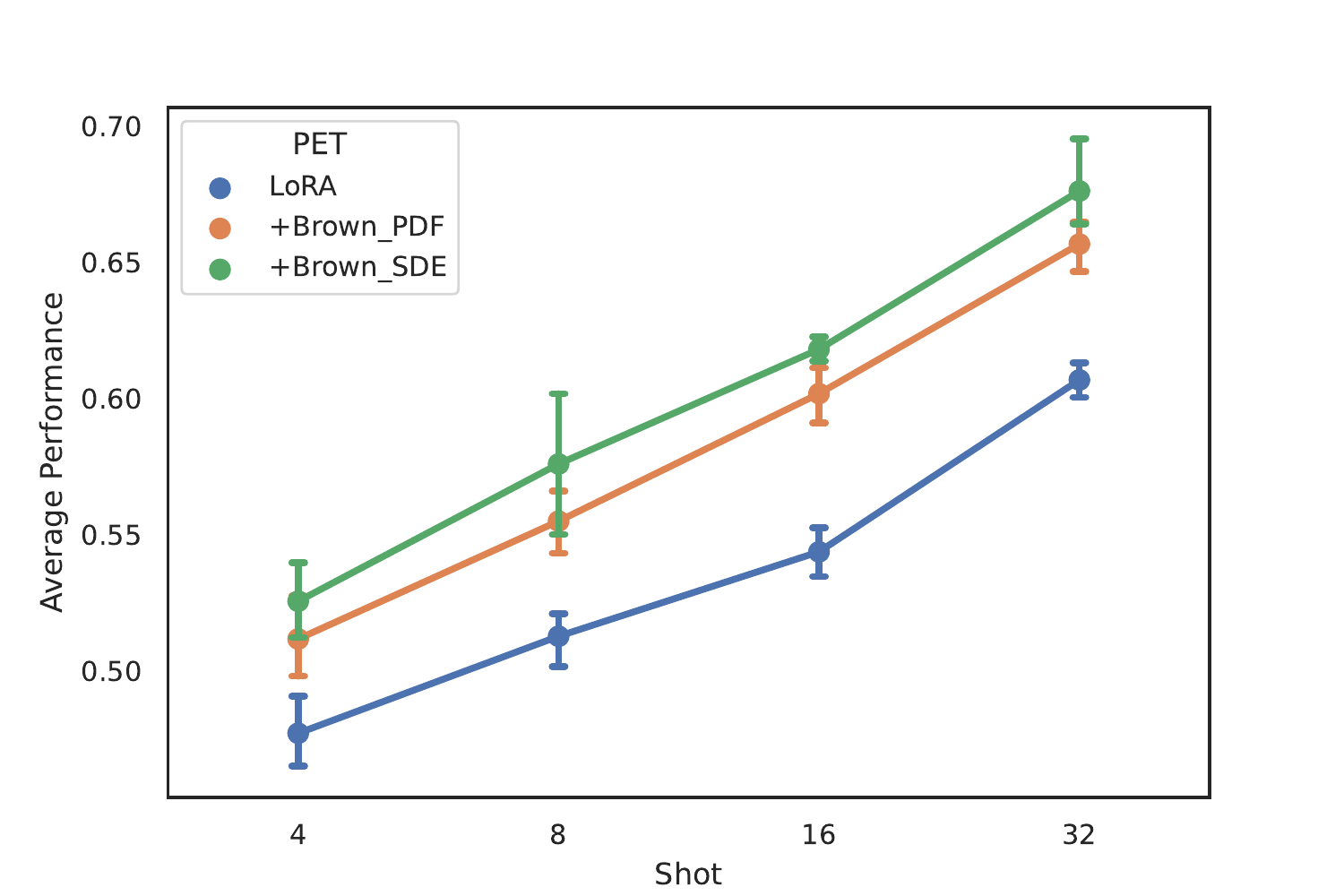}
        \caption{\textsc{LoRA}}
        \label{fig:deberta-shot-perf-lora}
    \end{subfigure}
    \hfill
    \begin{subfigure}[b]{0.49\textwidth}
        \centering
        \includegraphics[width=\textwidth]{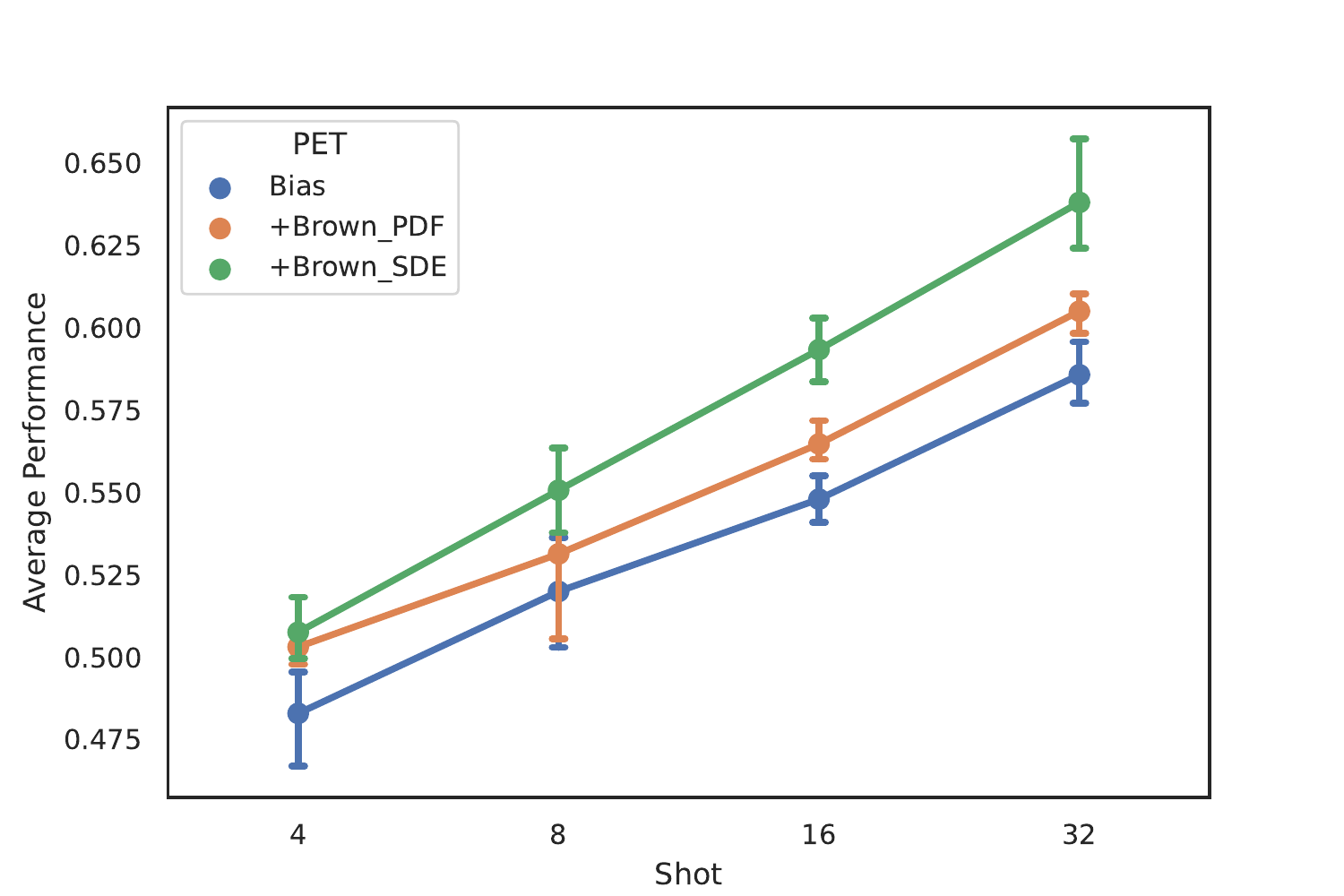}
        \caption{\textsc{BitFit}}
        \label{fig:deberta-shot-perf-bias}
    \end{subfigure}
    \hfill
    \begin{subfigure}[b]{0.49\textwidth}
        \centering
        \includegraphics[width=\textwidth]{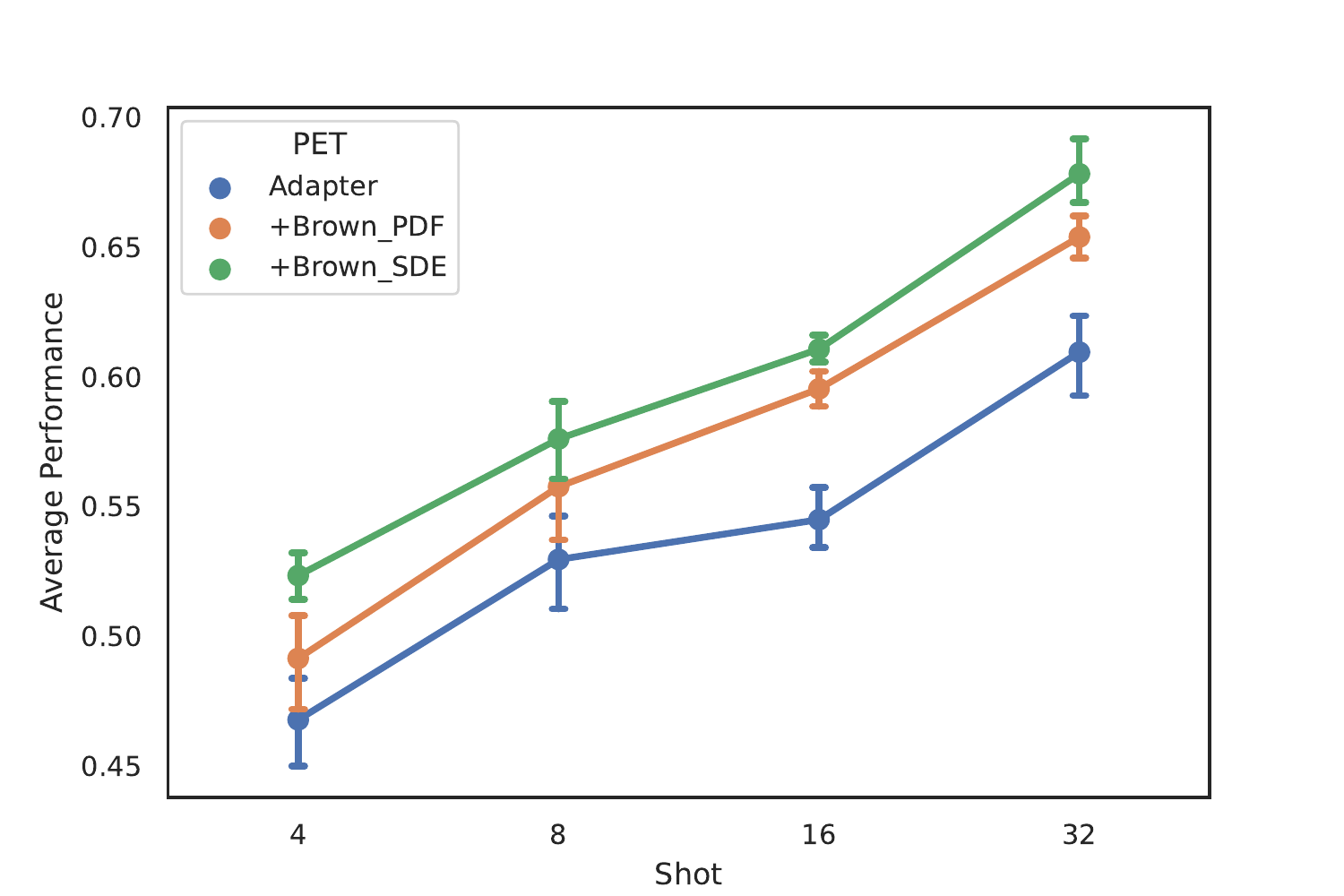}
        \caption{\textsc{Adapter}}
        \label{fig:deberta-shot-perf-adapter}
    \end{subfigure}
    \caption{The average $\deberta$ few-shot GLUE results trained with different PETs under different shots. The results are averaged across 5 different seeds and the error bars indicate the 95\% confidence.}
    \label{fig:deberta-shot-perf}
\end{figure*}
\section{Background for Parameter-Efficient Tuning Methods}
\label{sec:appendix-pet}
The large number of parameters in the PLMs makes fine-tuning impractical, therefore different PETs are proposed to mitigate the problem. The current PETs can be categorized into three groups: addition-based, specification-based and reparameterization-based~\citep{DBLP:journals/corr/abs-2203-06904}. To verify the generality of our method, we include one or two PETs from each category in this work, and we give a brief review to these PETs.

\textbf{Prompt Tuning} is an addition-based PET. It prepends or appends trainable virtual tokens $\mP\in\R^{m\times d}$ to each sequence $\vx\in\R^{n\times d}$ to form a new input sequence $[\mP; \vx]$ or $[\vx; \mP]\in\R^{(n+m)\times d}$, where $n, m$ are length of original sequence and virtual tokens respectively, $d$ is the embedding dimension. The virtual tokens $\mP$ can be either continuous~\cite{DBLP:conf/emnlp/LesterAC21} or be restricted to be embeddings of discrete tokens in vocabulary~\citep{DBLP:conf/acl/GaoFC20}.

\textbf{Adapter}~\citep{DBLP:conf/icml/HoulsbyGJMLGAG19} is an addition-based PET. It inserts two-layer MLPs after the attention module and feed-forward module at each layer. Denote $\vh\in\R^d$ as the input of Adapter, $r$ as the intermediate dimension of Adapter's MLP, $\mW_d\in\R^{r\times d}, \mW_u\in\R^{d\times r}$ as the down-projection and up-projection of Adapter, and $\sigma$ as the activation function. Then the computation of Adapter can be formulated as
\begin{align*}
    \vh\leftarrow \mW_u \sigma(\mW_d\vx)+\vh
\end{align*}

\textbf{BitFit}~\citep{DBLP:conf/acl/ZakenGR22} is a specification-based PET. It specifies the bias terms in layer normalization modules and linear transformation modules as trainable.

\textbf{LoRA}~\citep{DBLP:conf/iclr/HuSWALWWC22} is a reparameterization-based PET. It assumes that when training the model, the updates $\Delta\mW$ for model's pre-trained parameters $\mW\in\R^{d\times k}$ are low-rank, and thus reparameterize the $\Delta\mW$ of each matrix in attention module with a low-rank decomposition $\Delta\mW=\mB\mA$, where $\mB\in\R^{d\times r}, \mA\in\R^{r\times k}$. For a forward pass $\vh=\mW\vx$, the computation of LoRA can be written as
\begin{align*}
    \vh=(\mW+\Delta\mW)\vx=\mW\vx+\mB\mA\vx
\end{align*}

\begin{table}[t]
    \centering
    \begin{tabular}{l r r}
    \toprule
        & \textsc{PDF} & \textsc{SDE}\\
    \midrule
        Learning rate & 1e-3 & 1e-3\\
        Weight decay & 0 & 0\\
        Batch size & 128 & 128\\
        Grad norm & 1.0 & 1.0\\
        Max steps & 100k & 500k \\
        Warmup ratio & 0.01 & 0.01\\
    \bottomrule
    \end{tabular}
    \caption{Hyper-parameters for training $g_\gamma$}
    \label{tab:hyperparameter-g}
\end{table}
\section{Properties for Ornstein-Uhlenbeck Bridge}
\label{sec:appendix-ou-property}
\begin{proposition}[Properties of Ornstein-Uhlenbeck Bridge]
A Ornstein-Uhlenbeck $X^{T;\beta}$ pinned at $X^{T;\beta}_0=0$ and $X^{T;\beta}_T=\beta$ is the solution to the following SDE:
\begin{equation}
\label{eq:ou-sde}
\small
\begin{aligned}
    d{\tilde{X}}_{t}&=q\mu_tdt+\sigma dB_t,\\
    \mu_t&=-\coth\left[q(T-t)\right]{\tilde{X}}_{t}+\frac{\beta}{\sinh\left[q(T-t)\right]},\\
    \tilde{X}_0&=0,
\end{aligned}
\end{equation}
where $q$ is the diffusion coefficient and $\sigma$ is the diffusion for the OU process. The transition probability density function reads as:
\begin{equation}
\label{eq:ou-transition}
\small
\begin{aligned}
p^{T;\beta}(t, y\mid 0, 0)&=\frac{1}{\sqrt{2\pi\sigma(s,t)}}\;\exp\left\{-\frac{\left(y-\mu(t)\right)^{2}}{2\sigma(s,t)}\right\},\\
\mu(t)&=\frac{\sinh(q(T-t))}{\sinh(q(T-s))}\beta\\
\sigma(s,t)&=\frac{\sigma^{2}}{q}\;\frac{\sinh(q(T-t))\sinh(q(t-s))}{\sinh(q(T-s))}.
\end{aligned}
\end{equation}
\end{proposition}

\section{Other Results for GLUE Experiments}
\label{sec:appendix-glue-all}
In this section, we present the complete results including OU bridge regularizer for~\Tabref{tab:megatron-glue-full} and~\Tabref{tab:glue-16shot}. We also report the results for $\deberta$, and the results on few-shot GLUE for both $\megatronbert$ and $\deberta$ under 4-, 8-, 16-, and 32-shot. We observe that $\megatronbert$ cannot give reasonable answers on CoLA and the Matthews correlations are around $0$ for all the PETs and all the shots we have experienced with. However, the situation gets better for the larger model $\deberta$. Therefore, we exclude CoLA for $\megatronbert$ and keep it for $\deberta$. We only select the Brownian bridge as the representative in this section, since the Brownian bridge and Ornstein-Uhlenbeck bridge have no significant difference in~\Tabref{tab:megatron-glue-full}.

\begin{table*}[t!]
    \centering
    \begin{tabular}{l *{8}{r}}
    \toprule
         PET & MNLI & QQP & QNLI & SST-2 & MRPC & RTE & Average & $\Delta$\\ 
    \midrule
        \textsc{Prompt} & 38.1$_{1.5}$ & 53.0$_{3.1}$ & 51.6$_{1.4}$ & 70.1$_{4.9}$ & 50.1$_{3.0}$ & 48.0$_{1.3}$ & 51.8$_{0.9}$ &-\\
        \textsc{+Brown\_PDF} & 38.7$_{2.3}$ & 54.9$_{2.8}$ & 52.1$_{1.1}$ & 75.0$_{11.0}$ & 52.8$_{2.2}$ & 50.8$_{3.3}$ & 54.0$_{1.8}$ &2.2\\
        \textsc{+Brown\_SDE} & \textbf{40.6}$_{0.8}$ & 55.4$_{2.1}$ & 52.9$_{1.6}$ & 80.0$_{10.9}$ & 51.9$_{3.6}$ & 51.7$_{3.1}$ & 55.4$_{1.5}$ &3.6\\
        \textsc{+OU\_PDF} & 38.9$_{1.6}$ & 54.9$_{2.5}$ & 51.8$_{2.0}$ & 71.4$_{9.3}$ & \textbf{54.3}$_{2.6}$ & 51.5$_{3.5}$ & 53.8$_{1.9}$ &2.0\\
        \textsc{+OU\_SDE} & 40.0$_{1.7}$ & \textbf{56.8}$_{1.5}$ & \textbf{53.5}$_{2.2}$ & \textbf{82.7}$_{3.2}$ & 52.7$_{2.1}$ & \textbf{52.0}$_{1.9}$ & \textbf{56.3}$_{0.5}$ & \textbf{4.5}\\ 
    \midrule
        \textsc{LoRA} & 48.7$_{4.5}$ & 59.9$_{5.5}$ & 53.2$_{1.2}$ & 90.2$_{1.1}$ & 53.6$_{3.4}$ & 64.2$_{0.9}$ & 61.6$_{0.6}$ &-\\
        \textsc{+Brown\_PDF} & 52.0$_{1.3}$ & 62.7$_{2.2}$ & 55.1$_{3.8}$ & 91.3$_{0.1}$ & 57.4$_{5.0}$ & 65.3$_{1.5}$ & 64.0$_{0.7}$ &2.4\\
        \textsc{+Brown\_SDE} & \textbf{54.1}$_{0.9}$ & \textbf{65.5}$_{1.4}$ & 64.3$_{5.1}$ & 91.2$_{0.3}$ & 60.2$_{3.1}$ & 65.8$_{1.4}$ & \textbf{66.8}$_{0.6}$ & \textbf{5.2}\\ 
        \textsc{+OU\_PDF} & 51.6$_{2.4}$ & 62.5$_{2.3}$ & 55.3$_{3.4}$ & \textbf{91.4}$_{0.5}$ & 56.9$_{5.2}$ & 65.6$_{0.4}$ & 63.9$_{1.1}$ &2.3\\
        \textsc{+OU\_SDE} & 52.9$_{2.6}$ & 63.1$_{1.5}$ & \textbf{64.4}$_{5.1}$ & 91.3$_{0.3}$ & \textbf{62.2}$_{2.7}$ & \textbf{66.2}$_{1.1}$ & 66.7$_{0.7}$ &5.1\\
    \midrule
        \textsc{BitFit} & 48.4$_{1.6}$ & 56.0$_{6.1}$ & 51.7$_{2.5}$ & 90.8$_{0.8}$ & 52.0$_{2.3}$ & 61.7$_{1.4}$ & 60.1$_{1.1}$ &-\\
        \textsc{+Brown\_PDF} & 48.5$_{1.9}$ & 56.0$_{6.0}$ & 53.5$_{2.0}$ & 91.0$_{0.4}$ & 53.9$_{2.3}$ & 63.2$_{1.6}$ & 61.0$_{0.8}$ &0.9\\
        \textsc{+Brown\_SDE} & \textbf{52.3}$_{0.5}$ & \textbf{61.2}$_{2.9}$ & 58.8$_{4.7}$ & 90.8$_{0.4}$ & 54.8$_{3.0}$ & \textbf{63.9}$_{2.5}$ & \textbf{63.6}$_{0.8}$ & \textbf{3.5}\\ 
        \textsc{+OU\_PDF} & 49.0$_{1.8}$ & 56.1$_{6.1}$ & 52.8$_{2.2}$ & 90.8$_{0.7}$ & 53.5$_{2.2}$ & 62.5$_{1.3}$ & 60.8$_{0.8}$ &0.7\\
        \textsc{+OU\_SDE} & 51.8$_{1.2}$ & 58.4$_{2.2}$ & \textbf{58.9}$_{4.1}$ & \textbf{91.1}$_{0.4}$ & \textbf{56.4}$_{3.6}$ & 63.5$_{1.4}$ & 63.4$_{0.8}$ &3.3\\
    \midrule
        \textsc{Adapter} & 47.4$_{3.7}$ & 57.0$_{7.2}$ & 55.8$_{2.9}$ & 91.0$_{0.4}$ & 55.8$_{2.5}$ & 62.7$_{2.0}$ & 61.6$_{1.2}$ &-\\
        \textsc{+Brown\_PDF} & 49.0$_{4.8}$ & 58.5$_{7.4}$ & 56.9$_{3.1}$ & 91.4$_{0.2}$ & 57.2$_{4.9}$ & 63.2$_{3.0}$ & 62.7$_{1.5}$ &1.1\\
        \textsc{+Brown\_SDE} & \textbf{52.3}$_{2.2}$ & 62.4$_{2.9}$ & \textbf{64.8}$_{4.5}$ & \textbf{91.9}$_{0.4}$ & 57.3$_{4.1}$ & \textbf{63.8}$_{1.8}$ & \textbf{65.4}$_{1.3}$ & \textbf{3.8}\\ 
        \textsc{+OU\_PDF} & 48.4$_{4.3}$ & 58.4$_{7.4}$ & 57.7$_{3.7}$ & 91.3$_{0.5}$ & 57.1$_{3.3}$ & 63.0$_{1.8}$ & 62.6$_{1.3}$ &1.0\\
        \textsc{+OU\_SDE} & 48.3$_{4.5}$ & \textbf{62.8}$_{1.8}$ & 63.8$_{4.7}$ & 91.1$_{0.4}$ & \textbf{59.8}$_{2.3}$ & 63.6$_{1.9}$ & 64.9$_{0.6}$ &3.3\\
    \bottomrule
    \end{tabular}
    \caption{The complete results on GLUE for $\megatronbert$ under 16-shot setting. We exclude CoLA because all PETs fail to give reasonable result in few-shot setting.}
    \label{tab:glue-16shot-complete}
\end{table*}
\begin{table*}[t!]
    \centering
    \setlength{\tabcolsep}{4.3pt}
    \begin{tabular}{l *{9}{r}}
        \toprule
            PET & MNLI & QQP & QNLI & SST-2 & MRPC & CoLA & RTE & Average & $\Delta$\\ 
        \midrule
            \textsc{Prompt} & 84.4$_{0.1}$ & 85.3$_{0.3}$ & 91.5$_{0.1}$ & 95.5$_{0.1}$ & 73.9$_{2.4}$ & 55.5$_{3.4}$ & 60.8$_{1.5}$ & 78.1$_{0.6}$&-\\
            \textsc{+Brown\_PDF} & 84.7$_{0.2}$ & \textbf{85.5}$_{0.0}$ & 91.8$_{0.6}$ & 95.7$_{0.1}$ & 75.4$_{0.5}$ & 56.4$_{3.3}$ & 61.5$_{2.2}$ & 78.7$_{0.4}$&0.6\\
            \textsc{+OU\_PDF} & 84.7$_{0.1}$ & 85.4$_{0.1}$ & 91.8$_{0.3}$ & 95.6$_{0.2}$ & 76.9$_{1.0}$ & 57.1$_{1.2}$ & 60.5$_{3.0}$ & 78.9$_{0.2}$&0.8\\
            \textsc{+Brown\_SDE} & \textbf{84.9}$_{0.2}$ & 85.4$_{0.1}$ & 91.8$_{0.4}$ & \textbf{95.8}$_{0.3}$ & 78.8$_{1.2}$ & 61.4$_{2.9}$ & 64.7$_{1.1}$ & 80.4$_{0.2}$&2.3\\
            \textsc{+OU\_SDE} & 84.7$_{0.2}$ & 85.3$_{0.1}$ & \textbf{92.1}$_{0.3}$ & 95.5$_{0.2}$ & \textbf{80.2}$_{0.5}$ & \textbf{61.5}$_{3.5}$ & \textbf{65.9}$_{2.7}$ & \textbf{80.7}$_{0.8}$& \textbf{2.6}\\ 
        \midrule
            \textsc{LoRA} & 88.8$_{0.1}$ & 89.2$_{0.2}$ & 93.5$_{0.2}$ & 95.5$_{0.1}$ & 84.6$_{0.4}$ & 62.8$_{1.6}$ & 78.9$_{1.6}$ & 84.8$_{0.3}$&-\\
            \textsc{+Brown\_PDF} & \textbf{88.9}$_{0.1}$ & \textbf{89.6}$_{0.1}$ & \textbf{93.9}$_{0.1}$ & 95.6$_{0.2}$ & 85.1$_{0.7}$ & 63.7$_{0.5}$ & 80.0$_{0.5}$ & 85.2$_{0.1}$&0.4\\
            \textsc{+OU\_PDF} & \textbf{88.9}$_{0.2}$ & 89.4$_{0.1}$ & 93.7$_{0.1}$ & \textbf{95.7}$_{0.3}$ & 86.0$_{0.5}$ & 63.6$_{0.6}$ & 80.5$_{0.8}$ & 85.4$_{0.1}$&0.6\\
            \textsc{+Brown\_SDE} & \textbf{88.9}$_{0.1}$ & 89.5$_{0.1}$ & 93.7$_{0.1}$ & \textbf{95.7}$_{0.1}$ & \textbf{86.5}$_{1.2}$ & \textbf{63.9}$_{0.4}$ & \textbf{80.9}$_{0.8}$ & \textbf{85.6}$_{0.1}$& \textbf{0.8}\\ 
            \textsc{+OU\_SDE} & 88.8$_{0.0}$ & 89.5$_{0.1}$ & 93.7$_{0.2}$ & \textbf{95.7}$_{0.3}$ & 86.3$_{1.2}$ & 63.7$_{0.6}$ & 80.1$_{0.9}$ & 85.4$_{0.1}$&0.6\\
        \midrule
            \textsc{BitFit} & \textbf{87.9}$_{0.2}$ & 87.6$_{0.1}$ & 92.7$_{0.2}$ & 95.6$_{0.1}$ & 79.4$_{2.3}$ & 60.2$_{0.8}$ & 77.0$_{1.5}$ & 82.9$_{0.3}$&-\\
            \textsc{+Brown\_PDF} & \textbf{87.9}$_{0.1}$ & \textbf{87.8}$_{0.0}$ & \textbf{93.0}$_{0.2}$ & 95.7$_{0.1}$ & 83.1$_{0.8}$ & 60.3$_{0.6}$ & 78.3$_{0.9}$ & 83.7$_{0.2}$&0.8\\
            \textsc{+OU\_PDF} & \textbf{87.9}$_{0.1}$ & \textbf{87.8}$_{0.1}$ & \textbf{93.0}$_{0.1}$ & 95.7$_{0.1}$ & 82.6$_{0.8}$ & 59.8$_{1.0}$ & \textbf{78.8}$_{1.4}$ & 83.6$_{0.4}$&0.7\\
            \textsc{+Brown\_SDE} & \textbf{87.9}$_{0.2}$ & 87.7$_{0.0}$ & 92.8$_{0.1}$ & 95.7$_{0.1}$ & 83.3$_{0.8}$ & \textbf{61.1}$_{1.2}$ & 77.7$_{1.5}$ & \textbf{83.8}$_{0.3}$& \textbf{0.9}\\ 
            \textsc{+OU\_SDE} & \textbf{87.9}$_{0.1}$ & 87.6$_{0.1}$ & 92.7$_{0.1}$ & \textbf{95.8}$_{0.1}$ & \textbf{83.7}$_{0.8}$ & 60.8$_{1.6}$ & 77.1$_{0.9}$ & 83.7$_{0.4}$&0.8\\
        \midrule
            \textsc{Adapter} & 88.8$_{0.1}$ & 89.6$_{0.3}$ & 93.7$_{0.1}$ & 95.6$_{0.1}$ & 83.6$_{0.1}$ & 60.4$_{1.2}$ & 79.5$_{1.2}$ & 84.5$_{0.3}$&-\\
            \textsc{+Brown\_PDF} & \textbf{89.0}$_{0.1}$ & 89.7$_{0.2}$ & 93.8$_{0.2}$ & \textbf{95.8}$_{0.1}$ & 86.5$_{1.1}$ & \textbf{62.6}$_{0.7}$ & \textbf{83.2}$_{0.2}$ & \textbf{85.8}$_{0.2}$& \textbf{1.3}\\ 
            \textsc{+OU\_PDF} & 88.9$_{0.1}$ & 89.7$_{0.0}$ & 93.8$_{0.1}$ & \textbf{95.8}$_{0.1}$ & \textbf{86.8}$_{0.6}$ & 61.9$_{0.2}$ & 82.0$_{0.6}$ & 85.6$_{0.1}$&1.1\\
            \textsc{+Brown\_SDE} & 88.9$_{0.1}$ & \textbf{89.8}$_{0.1}$ & \textbf{93.9}$_{0.2}$ & \textbf{95.8}$_{0.1}$ & 85.9$_{0.4}$ & 62.3$_{1.8}$ & 82.2$_{0.2}$ & 85.5$_{0.2}$&1.0\\
            \textsc{+OU\_SDE} & 88.9$_{0.1}$ & \textbf{89.8}$_{0.1}$ & 93.7$_{0.1}$ & 95.7$_{0.1}$ & 85.9$_{0.7}$ & 62.5$_{1.2}$ & 82.7$_{0.5}$ & 85.6$_{0.2}$&1.1\\
        \bottomrule
    \end{tabular}
    \caption{The complete results on GLUE for $\megatronbert$. The values are the average value of the best performances over three different runs, and the subscripts are the standard deviations. The $\Delta$ column shows the difference of the average performance between the PETs with and without our regularizers.}
    \label{tab:megatron-glue-full-complete}
\end{table*}
\begin{table*}[t!]
    \centering
    \setlength{\tabcolsep}{4.3pt}
    \begin{tabular}{l *{9}{r}}
        \toprule
            PET & MNLI & QQP & QNLI & SST-2 & MRPC & CoLA & RTE & Average & $\Delta$\\ 
        \midrule
            \textsc{Prompt} & 87.2$_{0.1}$ & 86.5$_{0.1}$ & 93.8$_{0.1}$ & \textbf{96.8}$_{0.1}$ & 75.4$_{2.5}$ & 64.2$_{3.8}$ & 78.8$_{3.5}$ & 83.2$_{0.5}$&-\\
            \textsc{+Brown\_PDF} & \textbf{87.6}$_{0.1}$ & \textbf{86.8}$_{0.3}$ & \textbf{94.2}$_{0.1}$ & \textbf{96.8}$_{0.1}$ & 80.3$_{2.9}$ & \textbf{65.5}$_{1.3}$ & \textbf{79.8}$_{0.5}$ & 84.5$_{0.6}$&1.3\\
            \textsc{+Brown\_SDE} & \textbf{87.6}$_{0.1}$ & \textbf{86.8}$_{0.1}$ & 94.0$_{0.1}$ & \textbf{96.8}$_{0.1}$ & \textbf{84.4}$_{0.6}$ & 64.8$_{0.5}$ & 79.5$_{1.3}$ & \textbf{84.8}$_{0.2}$& \textbf{1.6}\\ 
        \midrule
            \textsc{LoRA} & \textbf{91.1}$_{0.1}$ & 90.3$_{0.1}$ & 95.1$_{0.1}$ & 96.8$_{0.1}$ & 88.7$_{0.7}$ & 68.0$_{1.3}$ & 83.4$_{1.1}$ & 87.6$_{0.3}$&-\\
            \textsc{+Brown\_PDF} & \textbf{91.1}$_{0.0}$ & \textbf{90.5}$_{0.0}$ & \textbf{95.2}$_{0.0}$ & \textbf{97.0}$_{0.1}$ & 90.1$_{0.8}$ & 68.6$_{0.8}$ & \textbf{85.9}$_{1.3}$ & 88.3$_{0.2}$&0.7\\
            \textsc{+Brown\_SDE} & \textbf{91.1}$_{0.1}$ & 90.4$_{0.0}$ & 95.1$_{0.0}$ & 96.9$_{0.2}$ & \textbf{90.5}$_{0.6}$ & \textbf{69.6}$_{1.1}$ & 85.6$_{0.9}$ & \textbf{88.5}$_{0.1}$& \textbf{0.9}\\ 
        \midrule
            \textsc{BitFit} & 90.0$_{0.1}$ & \textbf{88.4}$_{0.0}$ & \textbf{95.0}$_{0.0}$ & \textbf{96.6}$_{0.1}$ & 87.3$_{0.6}$ & 66.9$_{0.2}$ & 82.4$_{0.6}$ & 86.7$_{0.1}$&-\\
            \textsc{+Brown\_PDF} & \textbf{90.2}$_{0.0}$ & 88.3$_{0.1}$ & \textbf{95.0}$_{0.1}$ & \textbf{96.6}$_{0.1}$ & 89.8$_{0.5}$ & \textbf{67.9}$_{0.8}$ & 82.9$_{0.6}$ & 87.2$_{0.1}$&0.5\\
            \textsc{+Brown\_SDE} & 90.1$_{0.1}$ & 88.3$_{0.0}$ & 94.8$_{0.0}$ & \textbf{96.6}$_{0.1}$ & \textbf{90.4}$_{0.5}$ & \textbf{67.9}$_{0.4}$ & \textbf{83.8}$_{0.5}$ & \textbf{87.4}$_{0.1}$& \textbf{0.7}\\ 
        \midrule
            \textsc{Adapter} & 91.1$_{0.1}$ & 90.0$_{0.1}$ & 95.2$_{0.0}$ & 96.8$_{0.2}$ & 87.9$_{0.5}$ & 68.8$_{1.8}$ & 85.0$_{0.6}$ & 87.8$_{0.2}$&-\\
            \textsc{+Brown\_PDF} & \textbf{91.2}$_{0.1}$ & 90.0$_{0.0}$ & \textbf{95.3}$_{0.1}$ & \textbf{96.9}$_{0.2}$ & 89.2$_{0.8}$ & 70.1$_{1.0}$ & \textbf{86.9}$_{1.5}$ & 88.5$_{0.5}$&0.7\\
            \textsc{+Brown\_SDE} & \textbf{91.2}$_{0.2}$ & \textbf{90.1}$_{0.1}$ & 95.2$_{0.0}$ & \textbf{96.9}$_{0.2}$ & \textbf{90.3}$_{0.9}$ & \textbf{70.8}$_{1.1}$ & 86.3$_{1.4}$ & \textbf{88.7}$_{0.4}$& \textbf{0.9}\\ 
        \bottomrule
    \end{tabular}
    \caption{The results on GLUE for $\deberta$.}
    \label{tab:deberta-glue-full}
\end{table*}
\begin{table*}[t!]
    \centering
    \setlength{\tabcolsep}{4pt}
    \begin{tabular}{l *{9}{r}}
    \toprule
         PET & MNLI & QQP & QNLI & SST-2 & MRPC & CoLA & RTE & Average & $\Delta$\\ 
    \midrule
        \textsc{Prompt} & 34.4$_{1.2}$ & 53.2$_{5.1}$ & 51.7$_{1.7}$ & 73.3$_{8.7}$ & 50.2$_{3.2}$ & 2.5$_{2.5}$ & 52.2$_{4.9}$ & 45.4$_{2.1}$ &-\\
        \textsc{+Brown\_PDF} & 35.8$_{0.9}$ & 57.7$_{2.4}$ & \textbf{53.5}$_{1.5}$ & \textbf{87.5}$_{3.0}$ & 52.2$_{1.9}$ & 2.8$_{2.9}$ & \textbf{54.1}$_{1.8}$ & \textbf{49.1}$_{0.8}$ & \textbf{3.7}\\ 
        \textsc{+Brown\_SDE} & \textbf{35.9}$_{1.6}$ & \textbf{59.6}$_{6.4}$ & 53.1$_{1.9}$ & 82.1$_{4.2}$ & \textbf{55.4}$_{1.4}$ & \textbf{2.9}$_{3.0}$ & \textbf{54.1}$_{1.3}$ & 49.0$_{1.1}$ &3.6\\
    \midrule
        \textsc{LoRA} & 43.1$_{3.6}$ & 68.4$_{2.9}$ & 60.1$_{5.3}$ & \textbf{91.8}$_{1.1}$ & 57.6$_{1.5}$ & 3.1$_{3.8}$ & 56.6$_{2.3}$ & 54.4$_{1.0}$ &-\\
        \textsc{+Brown\_PDF} & \textbf{52.1}$_{3.4}$ & 70.2$_{2.9}$ & \textbf{73.3}$_{6.3}$ & 91.7$_{1.1}$ & 59.5$_{4.5}$ & 14.2$_{5.6}$ & 60.4$_{4.2}$ & 60.2$_{1.2}$ &5.8\\
        \textsc{+Brown\_SDE} & 49.6$_{4.3}$ & \textbf{70.6}$_{1.5}$ & 72.4$_{6.0}$ & 90.7$_{1.2}$ & \textbf{59.8}$_{4.5}$ & \textbf{28.9}$_{1.5}$ & \textbf{60.6}$_{3.1}$ & \textbf{61.8}$_{0.5}$ & \textbf{7.4}\\ 
    \midrule
        \textsc{BitFit} & 41.9$_{3.8}$ & 67.7$_{2.6}$ & 60.3$_{4.2}$ & \textbf{91.8}$_{0.8}$ & 54.9$_{2.5}$ & 9.4$_{2.4}$ & 57.6$_{2.0}$ & 54.8$_{0.8}$ &-\\
        \textsc{+Brown\_PDF} & 45.2$_{3.7}$ & \textbf{70.3}$_{1.2}$ & 65.4$_{6.7}$ & 90.9$_{0.8}$ & 55.6$_{2.1}$ & 8.2$_{2.5}$ & \textbf{59.6}$_{2.3}$ & 56.5$_{0.7}$ &1.7\\
        \textsc{+Brown\_SDE} & \textbf{45.7}$_{3.8}$ & 69.2$_{2.3}$ & \textbf{69.2}$_{6.3}$ & 89.7$_{1.3}$ & \textbf{57.8}$_{4.0}$ & \textbf{24.2}$_{4.6}$ & 59.4$_{2.8}$ & \textbf{59.3}$_{1.2}$ & \textbf{4.5}\\ 
    \midrule
        \textsc{Adapter} & 43.1$_{2.9}$ & 67.7$_{2.7}$ & 55.9$_{5.3}$ & \textbf{91.1}$_{0.9}$ & 56.1$_{2.1}$ & 8.6$_{5.6}$ & 59.0$_{2.3}$ & 54.5$_{1.4}$ &-\\
        \textsc{+Brown\_PDF} & \textbf{50.7}$_{3.0}$ & 70.1$_{1.6}$ & 70.9$_{5.2}$ & 90.6$_{1.7}$ & 57.6$_{4.3}$ & 16.1$_{7.4}$ & \textbf{60.6}$_{3.9}$ & 59.5$_{0.8}$ &5.0\\
        \textsc{+Brown\_SDE} & 47.1$_{1.6}$ & \textbf{72.0}$_{1.1}$ & \textbf{71.3}$_{4.3}$ & 91.0$_{1.0}$ & \textbf{59.7}$_{4.5}$ & \textbf{26.4}$_{5.5}$ & 60.0$_{5.7}$ & \textbf{61.1}$_{0.6}$ & \textbf{6.6}\\ 
    \bottomrule
    \end{tabular}
    \caption{The results on GLUE for $\deberta$ under 16-shot setting.}
    \label{tab:glue-deberta-16shot}
\end{table*}

In~\Tabref{tab:megatron-glue-full-complete} and~\Tabref{tab:glue-16shot-complete}, we report the performance of OU bridge regularizers. The experimental setups are the same as~\Tabref{tab:megatron-glue-full} and~\Tabref{tab:glue-16shot} respectively. The performances between OU bridge and Brownian bridge do not have a significant difference.
In~\Tabref{tab:deberta-glue-full}, we report the performance of $\deberta$ on full GLUE datasets. On all four PETs, the \textsc{SDE} regularizer outperforms the \textsc{PDF} regularizer, this is consistent with the results we see in~\Tabref{tab:megatron-glue-full}. The results for 4-, 8-, and 32-shot for $\megatronbert$ and $\deberta$ are plotted respectively in~\Figref{fig:shot-perf} and~\Figref{fig:deberta-shot-perf}. For simplicity, we only plot the average performance for each PET. We report the results for 16-shot experiments for $\deberta$ in~\Tabref{tab:glue-deberta-16shot}. The setup for the experiment is almost the same as the experiment in~\Secref{sec:experiments-glue-fewshot}, and the hyper-parameters are listed in~\Appref{sec:appendix-hyperparameters}. The \textsc{SDE} regularizer outperforms the \textsc{PDF} regularizer on most of the PETs except Prompt tuning. We notice that the \textsc{SDE} regularizer helps $\deberta$ substantially on CoLA for most of the PETs, indicating the \textsc{SDE} regularizer can effectively provide useful guidance when the data is scarce and the task is hard. 

\section{Calculation of Correlation in~\Secref{sec:analysis}}
\label{sec:appendix-correlation}
In this section, we elaborate on how we calculate the correlations reported in~\Tabref{tab:bridgeloss}.
\subsection{Correlation between Number of Shots and Distance to Bridge}
\begin{definition}[Tie]
A pair of observation $\{(x_i, y_i), (x_j, y_j)\}$ is defined as tied if $x_i=x_j$ or $y_i=y_j$.
\end{definition}
Since we generate the few-shot datasets using 5 random seeds for each shot, each PET has 5 results for each shot. This results in observations with ties, e.g., the two observations for distances on the first seed and second seed for 8-shot $\{(8, d_1), (8, d_2)\}$ are tied. To calculate the correlation for data with ties, the Tau-b of Kendall's rank correlation is more suitable than Pearson's correlation. We therefore report the Kendall's rank correlation for the correlation between the number of shots and the hidden states distance to the latent bridges.

\subsection{Correlation between Performance and Distance to Bridge}
We mix all the few-shot results for different shots and different seeds to form observations of performance and the hidden states distances to bridges, and then calculate the Pearson's correlation. 

\section{Hyper-parameters}
\label{sec:appendix-hyperparameters}
\subsection{Training $g_\gamma$}
we use simple 3-layer MLP for $g_\gamma$ in all of our experiments. For \textsc{PDF} regularizer, the output dimensions of each layer are 1024, 256, 128, and for \textsc{SDE} regularizer, the output dimensions of each layer are 1024, 256, 32. We observe that the running time increases noticeably with the final output dimension for \textsc{SDE} regularizer, we thus choose a smaller one for \textsc{SDE} regularizer. The hyper-parameters for training $g_\gamma$ are listed in~\Tabref{tab:hyperparameter-g}.

\subsection{Training PETs on full-set GLUE}
\begin{table*}[t]
    \centering
    \begin{tabular}{l *{7}{r}}
    \toprule
        & MNLI & QQP & QNLI & SST-2 & MRPC & CoLA & RTE\\
    \midrule
        \textit{BERT}$_\textit{large}$\\
        \textsc{Prompt} & 0.05 & 0.3 & 0.01 & 0.2 & 0.8 & 0.4 & 0.4\\
        \textsc{LoRA} & 0.2 & 0.8 & 0.5 & 0.1 & 0.1 & 0.5 & 0.3\\
        \textsc{Bias} & 0.05 & 0.3 & 0.6 & 0.05 & 0.8 & 0.05 & 0.7\\
        \textsc{Adapter} & 0.8 & 0.01 & 0.7 & 0.1 & 0.6 & 0.4 & 0.3\\
    \midrule
        \textit{Deberta}$_\textit{xlarge}$\\
        \textsc{Prompt} & 0.05 & 0.1 & 0.2 & 0.8 & 0.7 & 0.3 & 0.7\\
        \textsc{LoRA} & 0.1 & 0.1 & 0.05 & 0.01 & 0.1 & 0.01 & 0.2\\
        \textsc{Bias} & 0.05 & 0.01 & 0.01 & 0.01 & 0.8 & 0.01 & 0.8\\
        \textsc{Adapter} & 0.3 & 0.01 & 0.01 & 0.01 & 0.01 & 0.6 & 0.1\\
    \bottomrule
    \end{tabular}
    \caption{Best $\alpha$ for \textsc{PDF} regularizer on full-set GLUE}
    \label{tab:hyperparameter-fullglue-pdf-alpha}
\end{table*}
\begin{table*}[t]
    \centering
    \begin{tabular}{l *{7}{r}}
    \toprule
        & MNLI & QQP & QNLI & SST-2 & MRPC & CoLA & RTE\\
    \midrule
        \textit{BERT}$_\textit{large}$\\
        \textsc{Prompt} & 0.005 & 0.005 & 0.0005 & 0.0001 & 1.0 & 0.2 & 0.05\\
        \textsc{LoRA} & 0.0005 & 0.01 & 0.01 & 0.01 & 0.005 & 0.0005 & 0.0005\\
        \textsc{Bias} & 0.001 & 0.001 & 0.005 & 0.0001 & 1.0 & 1.0 & 0.005\\
        \textsc{Adapter} & 0.001 & 0.001 & 0.0005 & 0.001 & 0.005 & 0.5 & 0.005\\
    \midrule
        \textit{Deberta}$_\textit{xlarge}$\\
        \textsc{Prompt} & 0.0001 & 0.001 & 0.0001 & 0.0001 & 0.2 & 0.0001 & 0.0001\\
        \textsc{LoRA} & 0.0005 & 0.05 & 0.0001 & 0.0001 & 0.1 & 0.005 & 0.0001\\
        \textsc{Bias} & 0.0001 & 0.0005 & 0.0001 & 0.0001 & 0.05 & 0.2 & 0.001\\
        \textsc{Adapter} & 0.001 & 0.0005 & 0.0001 & 0.0005 & 0.1 & 0.001 & 0.0005\\
    \bottomrule
    \end{tabular}
    \caption{Best $\alpha$ for \textsc{SDE} regularizer on full-set GLUE}
    \label{tab:hyperparameter-fullglue-sde-alpha}
\end{table*}
We run all the experiments for 50k steps, and evaluate on the development set every 1k steps. For $\megatronbert$, we use 32 as the batch size while for $\deberta$, we use 16 as the batch size. We choose learning rate 1e-3 for Prompt tuning for both PLMs, and 1e-4 for other PETs for both PLMs. We use 0.01 weight decay, 1.0 maximum gradient norm and no learning rate warm-up for all the experiments. We search the best regularization strength $\alpha$ in \{0.01, 0.05, 0.1, 0.2, 0.3, 0.4, 0.5, 0.6, 0.7, 0.8, 0.9, 1.0\} for \textsc{PDF} regularizer, and in \{0.0001, 0.0005, 0.001, 0.005, 0.01, 0.05, 0.1, 0.2, 0.5, 1.0\} for \textsc{SDE} regularizer. The best $\alpha$ for \textsc{PDF} regularizer are listed in~\Tabref{tab:hyperparameter-fullglue-pdf-alpha}, and best $\alpha$ for \textsc{SDE} regularizer are listed in~\Tabref{tab:hyperparameter-fullglue-sde-alpha}.

\subsection{Training PETs on few-shot GLUE}
We run all the experiments for 1k steps, and evaluate on the development set every 50 steps. For all the shots for both regularizers and both models, we use a batch size of 2. Other hyper-parameters are kept the same as in the experiments on full-set GLUE.

\section{Performance of Regularizers Trained with Tiny Corpus}
\begin{table*}[t]
    \centering
    \resizebox{\linewidth}{!}{
    \begin{tabular}{l *{10}{r}}
        \toprule
            PET & MNLI & QQP & QNLI & SST-2 & MRPC & CoLA & RTE & Average & $\Delta$ & $\Delta_\text{whole}$\\ 
        \midrule
            \textsc{Prompt} & 84.4$_{0.1}$ & 85.3$_{0.3}$ & 91.5$_{0.1}$ & 95.5$_{0.1}$ & 73.9$_{2.4}$ & 55.5$_{3.4}$ & 60.8$_{1.5}$ & 78.1$_{0.6}$&-&-\\
            \textsc{+Brown\_PDF} & 84.7$_{0.0}$ & \textbf{85.4}$_{0.1}$ & \textbf{92.1}$_{0.1}$ & \textbf{95.8}$_{0.2}$ & 75.8$_{1.4}$ & 56.6$_{0.9}$ & 61.1$_{3.0}$ & 78.8$_{0.5}$&0.7&+0.1\\
            \textsc{+Brown\_SDE} & \textbf{84.8}$_{0.2}$ & \textbf{85.4}$_{0.1}$ & \textbf{92.1}$_{0.1}$ & \textbf{95.8}$_{0.2}$ & \textbf{79.0}$_{1.3}$ & \textbf{59.8}$_{8.2}$ & \textbf{65.5}$_{1.2}$ & \textbf{80.3}$_{1.2}$&\textbf{2.2}& -0.1\\ 
        \midrule
            \textsc{LoRA} & 88.8$_{0.1}$ & 89.2$_{0.2}$ & 93.5$_{0.2}$ & 95.5$_{0.1}$ & 84.6$_{0.4}$ & 62.8$_{1.6}$ & 78.9$_{1.6}$ & 84.8$_{0.3}$&-&-\\
            \textsc{+Brown\_PDF} & \textbf{88.9}$_{0.1}$ & \textbf{89.5}$_{0.1}$ & \textbf{93.8}$_{0.1}$ & \textbf{95.8}$_{0.2}$ & 86.0$_{0.6}$ & \textbf{64.4}$_{0.6}$ & 80.4$_{0.3}$ & \textbf{85.5}$_{0.1}$&\textbf{0.7}& +0.3\\ 
            \textsc{+Brown\_SDE} & \textbf{88.9}$_{0.0}$ & 89.4$_{0.1}$ & \textbf{93.8}$_{0.1}$ & 95.6$_{0.2}$ & \textbf{86.8}$_{0.5}$ & 63.5$_{0.9}$ & \textbf{80.9}$_{0.8}$ & \textbf{85.5}$_{0.2}$&\textbf{0.7}& -0.1\\ 
        \midrule
            \textsc{BitFit} & \textbf{87.9}$_{0.2}$ & 87.6$_{0.1}$ & 92.7$_{0.2}$ & 95.6$_{0.1}$ & 79.4$_{2.3}$ & 60.2$_{0.8}$ & 77.0$_{1.5}$ & 82.9$_{0.3}$&-&-\\
            \textsc{+Brown\_PDF} & \textbf{87.9}$_{0.1}$ & \textbf{87.7}$_{0.1}$ & \textbf{92.8}$_{0.1}$ & \textbf{95.7}$_{0.2}$ & \textbf{83.6}$_{0.7}$ & \textbf{61.2}$_{0.2}$ & \textbf{78.7}$_{1.1}$ & \textbf{84.0}$_{0.1}$&\textbf{1.1}& +0.3\\ 
            \textsc{+Brown\_SDE} & \textbf{87.9}$_{0.1}$ & \textbf{87.7}$_{0.1}$ & \textbf{92.8}$_{0.1}$ & \textbf{95.7}$_{0.1}$ & 82.8$_{0.7}$ & 61.1$_{0.8}$ & 77.4$_{0.6}$ & 83.6$_{0.3}$&0.7&-0.2\\
        \midrule
            \textsc{Adapter} & 88.8$_{0.1}$ & 89.6$_{0.3}$ & 93.7$_{0.1}$ & 95.6$_{0.1}$ & 83.6$_{0.1}$ & 60.4$_{1.2}$ & 79.5$_{1.2}$ & 84.5$_{0.3}$&-&-\\
            \textsc{+Brown\_PDF} & \textbf{88.9}$_{0.1}$ & \textbf{89.8}$_{0.1}$ & \textbf{93.8}$_{0.2}$ & \textbf{95.9}$_{0.2}$ & 85.4$_{0.8}$ & 60.9$_{0.5}$ & \textbf{82.7}$_{1.8}$ & 85.3$_{0.2}$&0.8&-0.5\\
            \textsc{+Brown\_SDE} & \textbf{88.9}$_{0.0}$ & \textbf{89.8}$_{0.1}$ & 93.7$_{0.1}$ & 95.8$_{0.3}$ & \textbf{85.7}$_{0.5}$ & \textbf{61.9}$_{0.9}$ & 82.4$_{0.3}$ & \textbf{85.5}$_{0.2}$&\textbf{1.0}& 0.0\\ 
        \bottomrule
    \end{tabular}
    }
    \caption{The results on GLUE for $\megatronbert$ \textit{with regularizers trained on 0.1\% of the pre-training corpus}. $\Delta_\text{whole}$ is the difference between the the average performance in this table and the average performance in~\Tabref{tab:megatron-glue-full}.}
    \label{tab:megatron-1wlines-glue-full}
\end{table*}
\begin{table*}[t]
    \centering
    \begin{tabular}{l *{9}{r}}
    \toprule
         PET & MNLI & QQP & QNLI & SST-2 & MRPC & RTE & Average & $\Delta$ & $\Delta_\text{whole}$\\ 
    \midrule
        \textsc{Prompt} & 38.1$_{1.5}$ & 53.0$_{3.1}$ & 51.6$_{1.4}$ & 70.1$_{4.9}$ & 50.1$_{3.0}$ & 48.0$_{1.3}$ & 51.8$_{0.9}$ &-&-\\
        \textsc{+Brown\_PDF} & 38.4$_{1.5}$ & 54.6$_{2.5}$ & 52.4$_{1.4}$ & 73.1$_{6.0}$ & \textbf{54.0}$_{1.4}$ & \textbf{51.1}$_{1.7}$ & 53.9$_{1.1}$ &2.1 & -0.1\\
        \textsc{+Brown\_SDE} & \textbf{39.3}$_{1.1}$ & \textbf{56.2}$_{3.0}$ & \textbf{52.8}$_{1.1}$ & \textbf{80.8}$_{7.1}$ & 53.0$_{6.0}$ & 51.0$_{2.5}$ & \textbf{55.5}$_{0.9}$ & \textbf{3.7} & +0.1\\ 
    \midrule
        \textsc{LoRA} & 48.7$_{4.5}$ & 59.9$_{5.5}$ & 53.2$_{1.2}$ & 90.2$_{1.1}$ & 53.6$_{3.4}$ & 64.2$_{0.9}$ & 61.6$_{0.6}$ &- &-\\
        \textsc{+Brown\_PDF} & 51.4$_{2.2}$ & 62.0$_{1.8}$ & 54.8$_{3.1}$ & \textbf{91.2}$_{0.3}$ & 57.4$_{4.0}$ & 65.7$_{0.7}$ & 63.7$_{0.9}$ &2.1 & -0.3\\
        \textsc{+Brown\_SDE} & \textbf{53.2}$_{1.9}$ & \textbf{65.4}$_{1.7}$ & \textbf{64.2}$_{5.0}$ & \textbf{91.2}$_{0.3}$ & \textbf{61.4}$_{3.6}$ & \textbf{66.3}$_{0.7}$ & \textbf{66.9}$_{0.7}$ & \textbf{5.3} & -0.1\\ 
    \midrule
        \textsc{BitFit} & 48.4$_{1.6}$ & 56.0$_{6.1}$ & 51.7$_{2.5}$ & 90.8$_{0.8}$ & 52.0$_{2.3}$ & 61.7$_{1.4}$ & 60.1$_{1.1}$ &-&-\\
        \textsc{+Brown\_PDF} & 49.1$_{1.8}$ & 56.0$_{6.0}$ & 53.1$_{2.4}$ & \textbf{91.1}$_{0.2}$ & 54.0$_{2.0}$ & 62.8$_{1.0}$ & 61.0$_{0.7}$ &0.9 & 0.0\\
        \textsc{+Brown\_SDE} & \textbf{51.4}$_{1.4}$ & \textbf{60.1}$_{1.7}$ & \textbf{56.7}$_{1.8}$ & \textbf{91.1}$_{0.3}$ & \textbf{57.3}$_{4.5}$ & \textbf{63.0}$_{1.2}$ & \textbf{63.2}$_{0.7}$ & \textbf{3.1} & -0.4\\ 
    \midrule
        \textsc{Adapter} & 47.4$_{3.7}$ & 57.0$_{7.2}$ & 55.8$_{2.9}$ & 91.0$_{0.4}$ & 55.8$_{2.5}$ & 62.7$_{2.0}$ & 61.6$_{1.2}$ &- &-\\
        \textsc{+Brown\_PDF} & 48.1$_{4.0}$ & 58.3$_{6.9}$ & 57.9$_{3.5}$ & 91.4$_{0.4}$ & 57.7$_{3.6}$ & 63.0$_{3.0}$ & 62.7$_{1.0}$ &1.1 & 0.0\\
        \textsc{+Brown\_SDE} & \textbf{50.3}$_{2.2}$ & \textbf{61.0}$_{6.1}$ & \textbf{63.4}$_{4.4}$ & \textbf{91.5}$_{0.3}$ & \textbf{58.4}$_{2.2}$ & \textbf{64.3}$_{1.9}$ & \textbf{64.8}$_{1.1}$ & \textbf{3.2} & -0.6\\ 
    \bottomrule
    \end{tabular}
    \caption{The results on GLUE for $\megatronbert$ under 16-shot setting \textit{with regularizers trained on 0.1\% of the pre-training corpus}. $\Delta_\text{whole}$ is the difference between the the average performance in this table and the average performance in~\Tabref{tab:glue-16shot}.}
    \label{tab:glue-1wlines-16shot}
\end{table*}
\label{sec:appendix-tiny-corpus}
Although we use the pre-training corpus to train our mapping $g_\gamma$, the training is actually fast and data-efficient. We show that when using only 10,000 documents in the pre-training corpus (about 0.1\% of the corpus), the obtained regularizers still perform great and are comparable to the regularizers trained on the whole pre-training corpus. We train the mapping $g_\gamma$ for 5,000 iterations with 128 batch size. On a single NVIDIA A100 GPU, the training can be done in 1 hour for PDF regularizer, and 3 hours for SDE regularizer. The cost of training our regularizer is quite small compared to the resources required for pre-training. 

We conduct the same experiments as~\Secref{sec:experiments-glue} and~\Secref{sec:experiments-glue-fewshot} with the regularizers trained with tiny corpus. The results are presented in~\Tabref{tab:megatron-1wlines-glue-full} and~\Tabref{tab:glue-1wlines-16shot} respectively.

On full-set GLUE, the PDF regularizer performs even better on three out of four PETs, and although its performance is affected on Adapter, it still outperforms the vanilla Adapter. The SDE regularizer is slightly affected on three out of four PETs, but it still brings substantial improvements on all the PETs.

On few-shot GLUE, the impact of the shrinkage of the corpus is relatively obvious. But overall, the regularizers still performs great on all the PETs. The drop in performances are relatively small compared to the boost they bring to vanilla PETs.

\begin{figure*}[t!]
    \centering
    \begin{subfigure}[b]{0.24\textwidth}
        \centering
        \includegraphics[width=\textwidth]{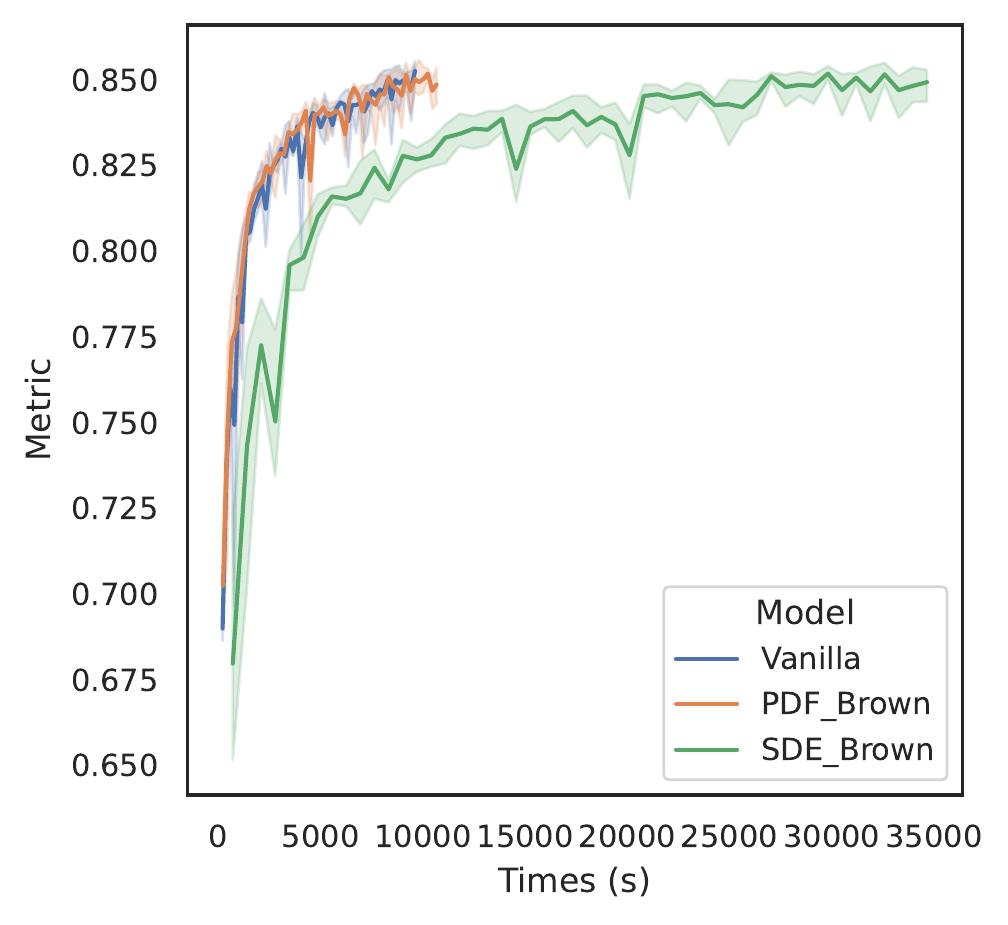}
        \caption{QQP}
    \end{subfigure}
    \hfill
    \begin{subfigure}[b]{0.24\textwidth}
        \centering
        \includegraphics[width=\textwidth]{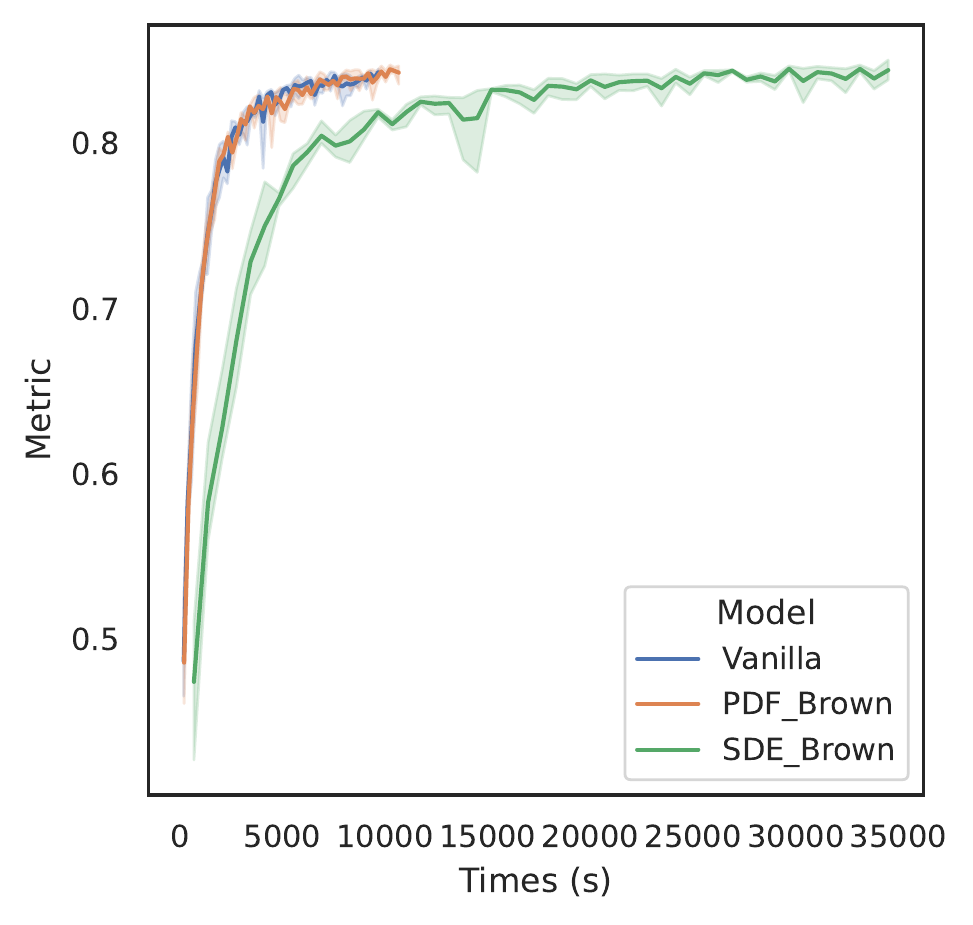}
        \caption{MNLI}
    \end{subfigure}
    \hfill
    \begin{subfigure}[b]{0.24\textwidth}
        \centering
        \includegraphics[width=\textwidth]{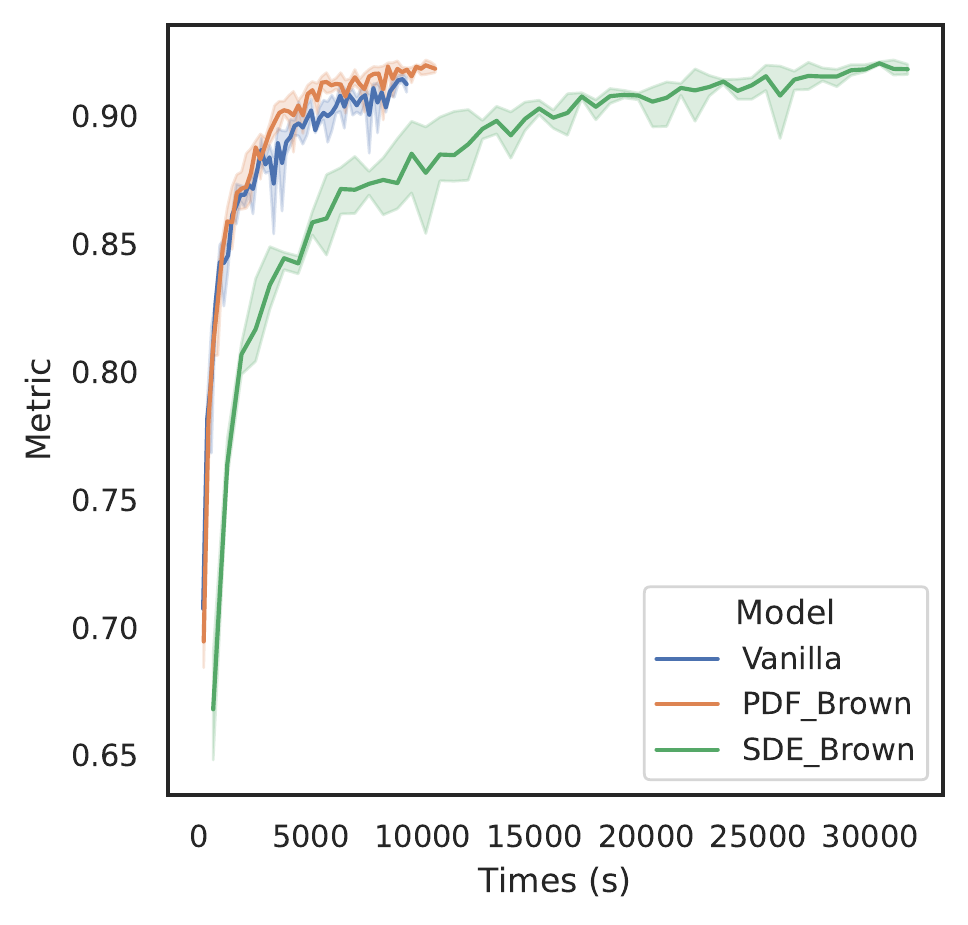}
        \caption{QNLI}
    \end{subfigure}
    \hfill
    \begin{subfigure}[b]{0.24\textwidth}
        \centering
        \includegraphics[width=\textwidth]{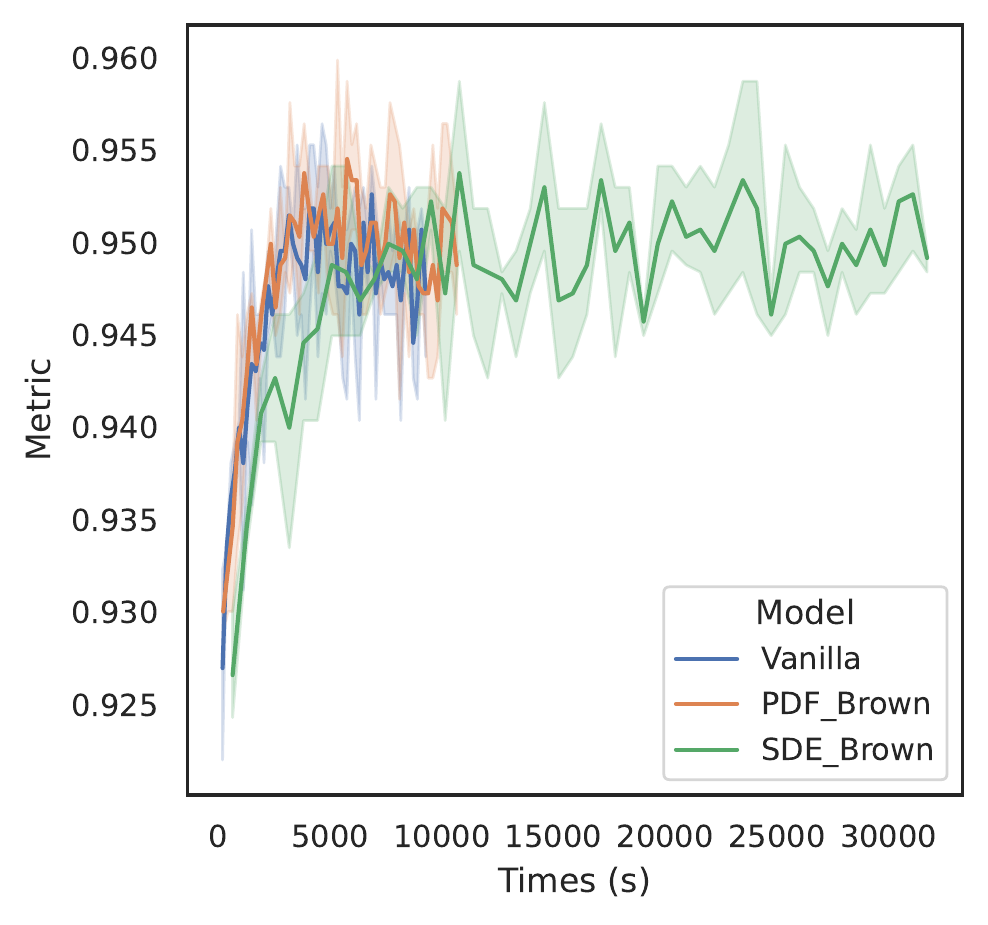}
        \caption{SST-2}
    \end{subfigure}
    \hfill
    \begin{subfigure}[b]{0.24\textwidth}
        \centering
        \includegraphics[width=\textwidth]{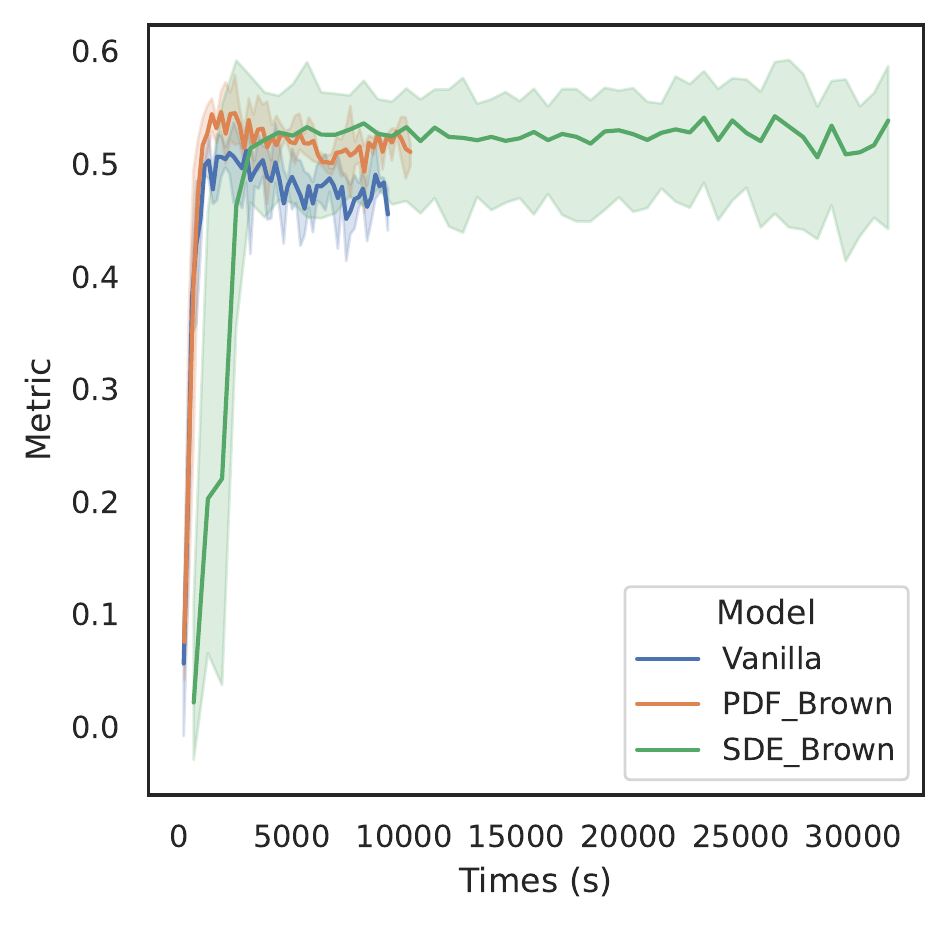}
        \caption{CoLA}
    \end{subfigure}
    \begin{subfigure}[b]{0.24\textwidth}
        \centering
        \includegraphics[width=\textwidth]{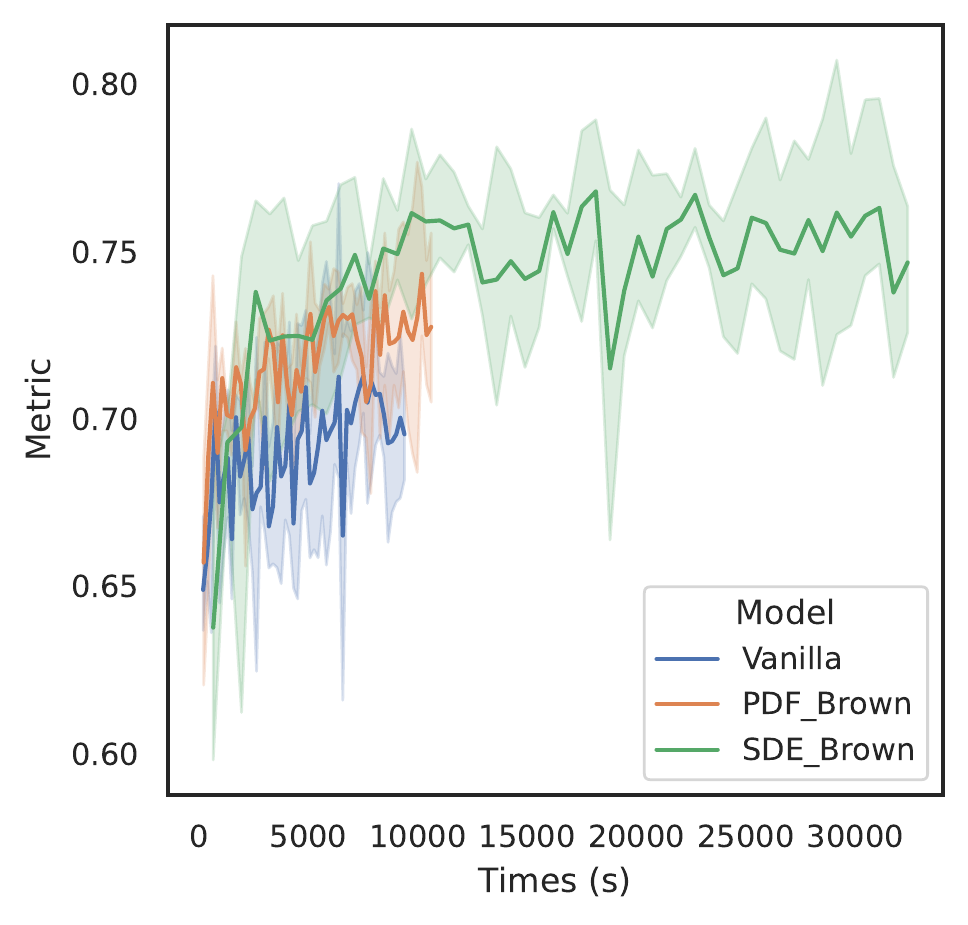}
        \caption{MRPC}
    \end{subfigure}
    \begin{subfigure}[b]{0.24\textwidth}
        \centering
        \includegraphics[width=\textwidth]{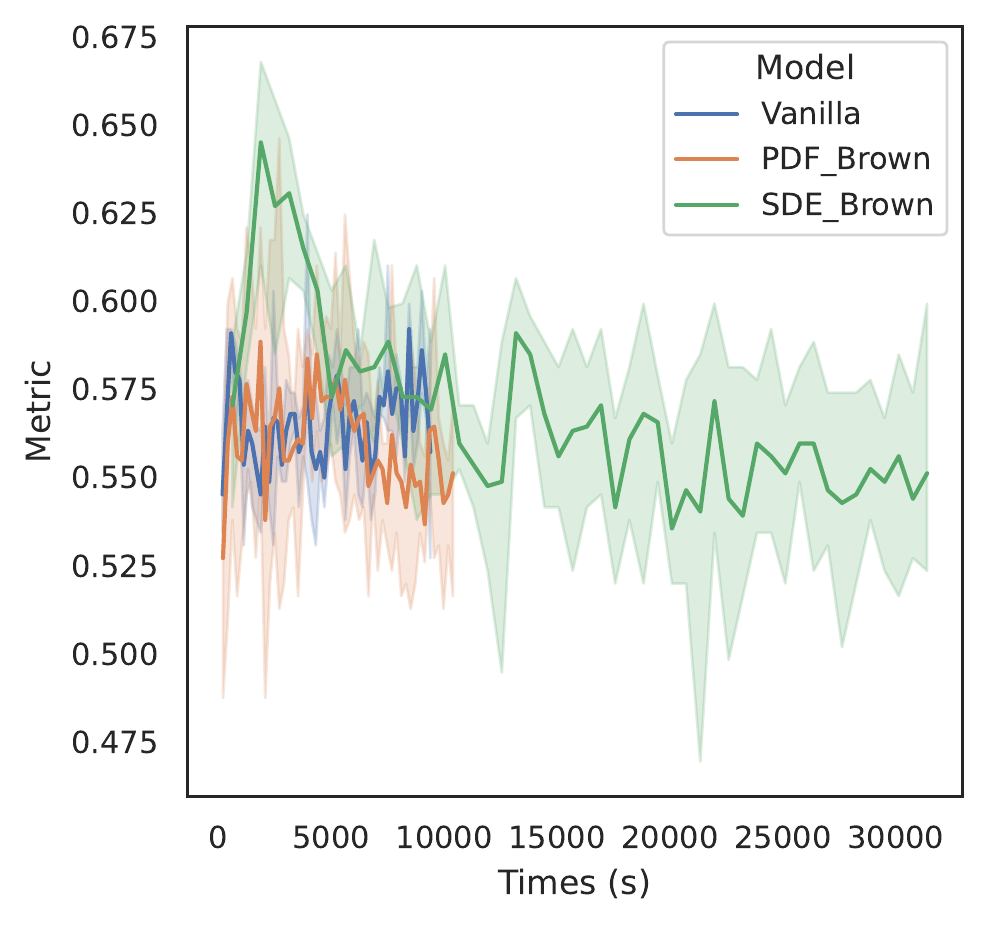}
        \caption{RTE}
    \end{subfigure}
    \hfill
    \caption{Time-Metric curve for regularizers on \emph{prompt tuning}.}
    \label{fig:speed_prompt}
\end{figure*}
\begin{figure*}[t!]
    \centering
    \begin{subfigure}[b]{0.24\textwidth}
        \centering
        \includegraphics[width=\textwidth]{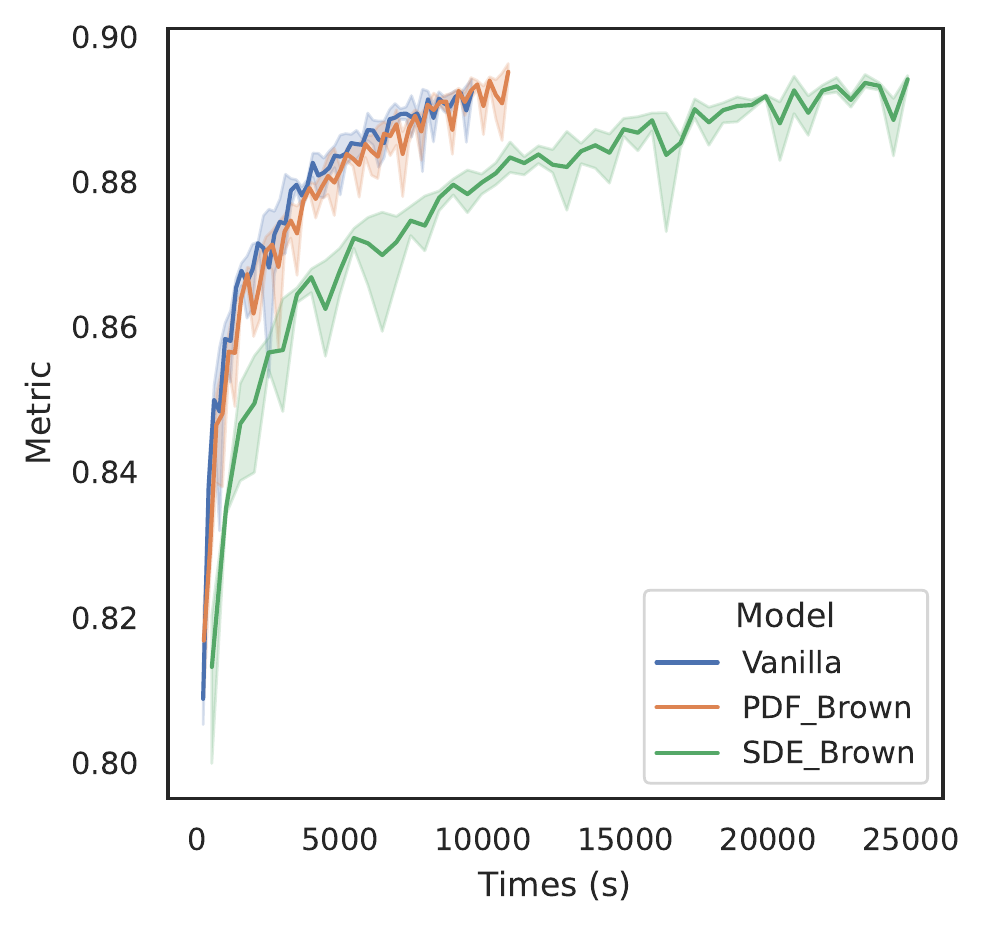}
        \caption{QQP}
    \end{subfigure}
    \hfill
    \begin{subfigure}[b]{0.24\textwidth}
        \centering
        \includegraphics[width=\textwidth]{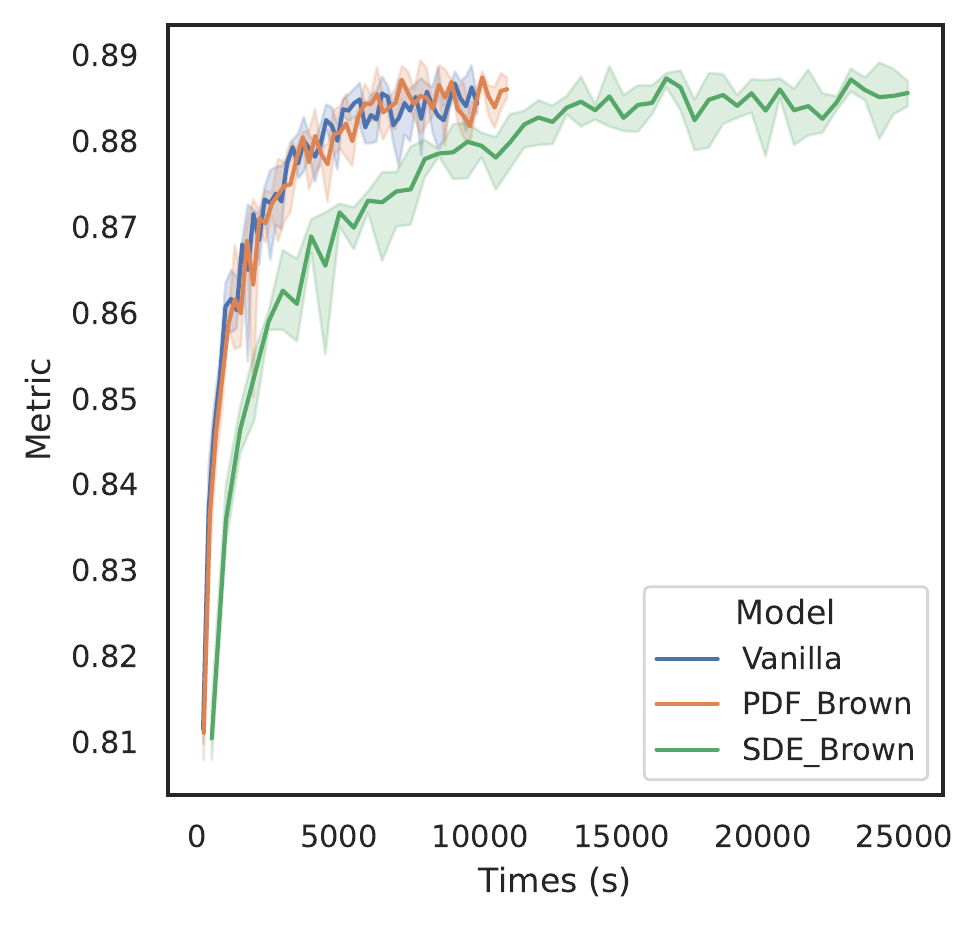}
        \caption{MNLI}
    \end{subfigure}
    \hfill
    \begin{subfigure}[b]{0.24\textwidth}
        \centering
        \includegraphics[width=\textwidth]{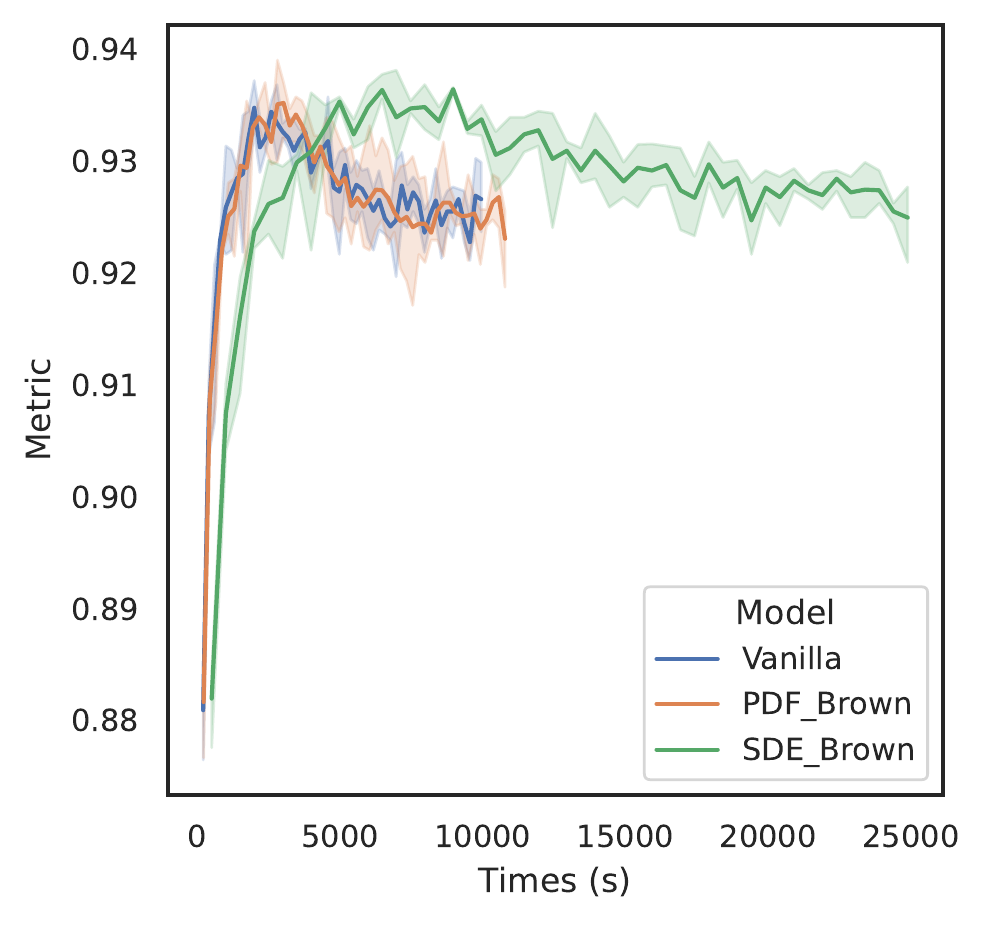}
        \caption{QNLI}
    \end{subfigure}
    \hfill
    \begin{subfigure}[b]{0.24\textwidth}
        \centering
        \includegraphics[width=\textwidth]{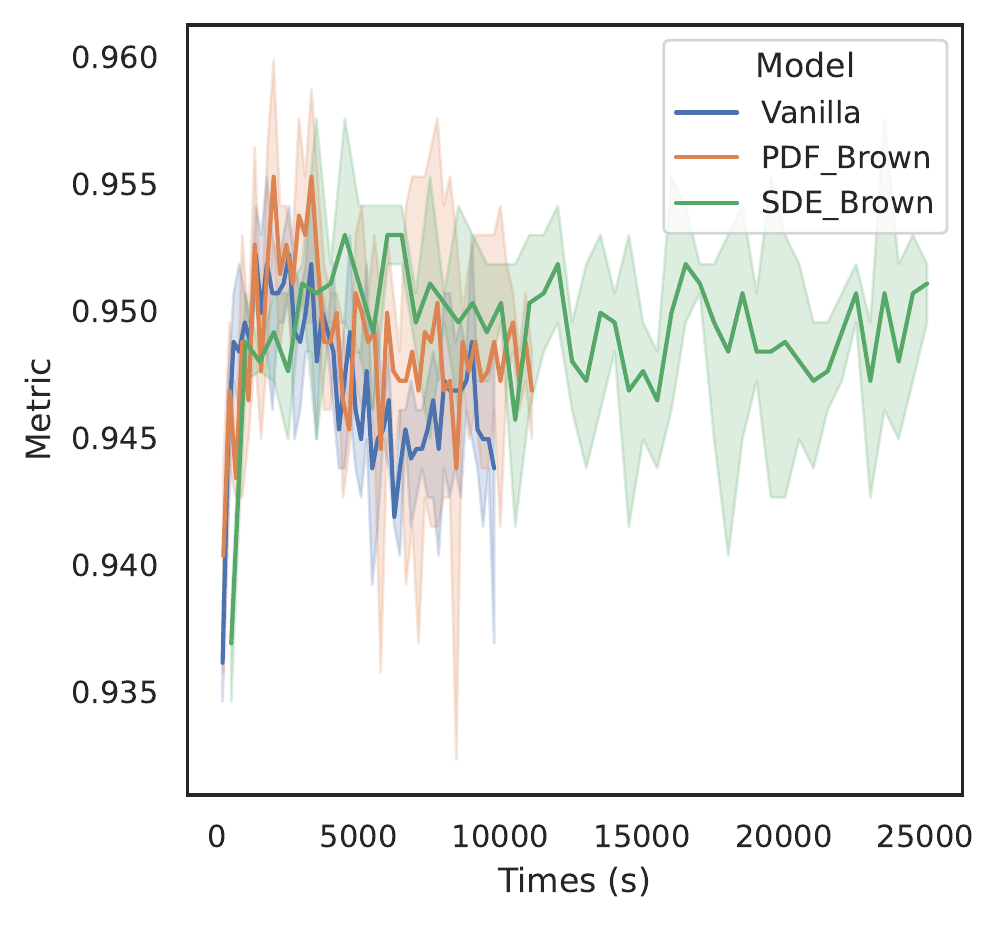}
        \caption{SST-2}
    \end{subfigure}
    \hfill
    \begin{subfigure}[b]{0.24\textwidth}
        \centering
        \includegraphics[width=\textwidth]{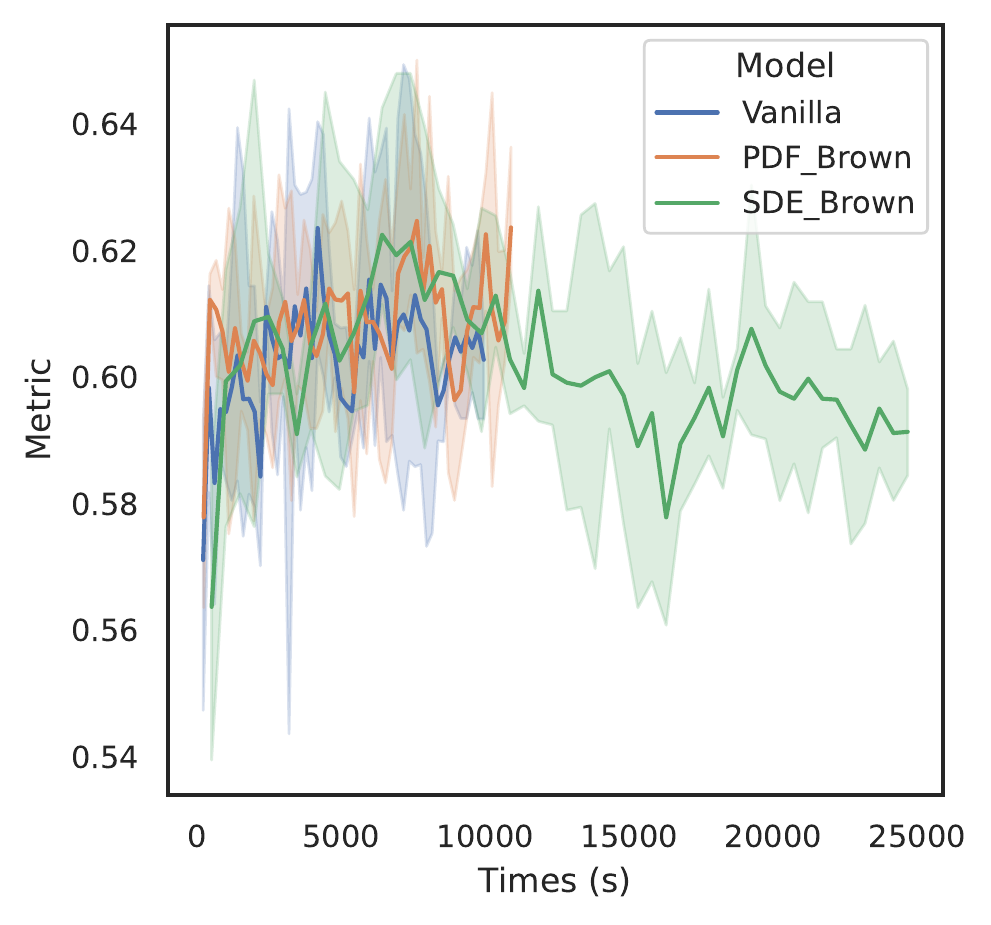}
        \caption{CoLA}
    \end{subfigure}
    \begin{subfigure}[b]{0.24\textwidth}
        \centering
        \includegraphics[width=\textwidth]{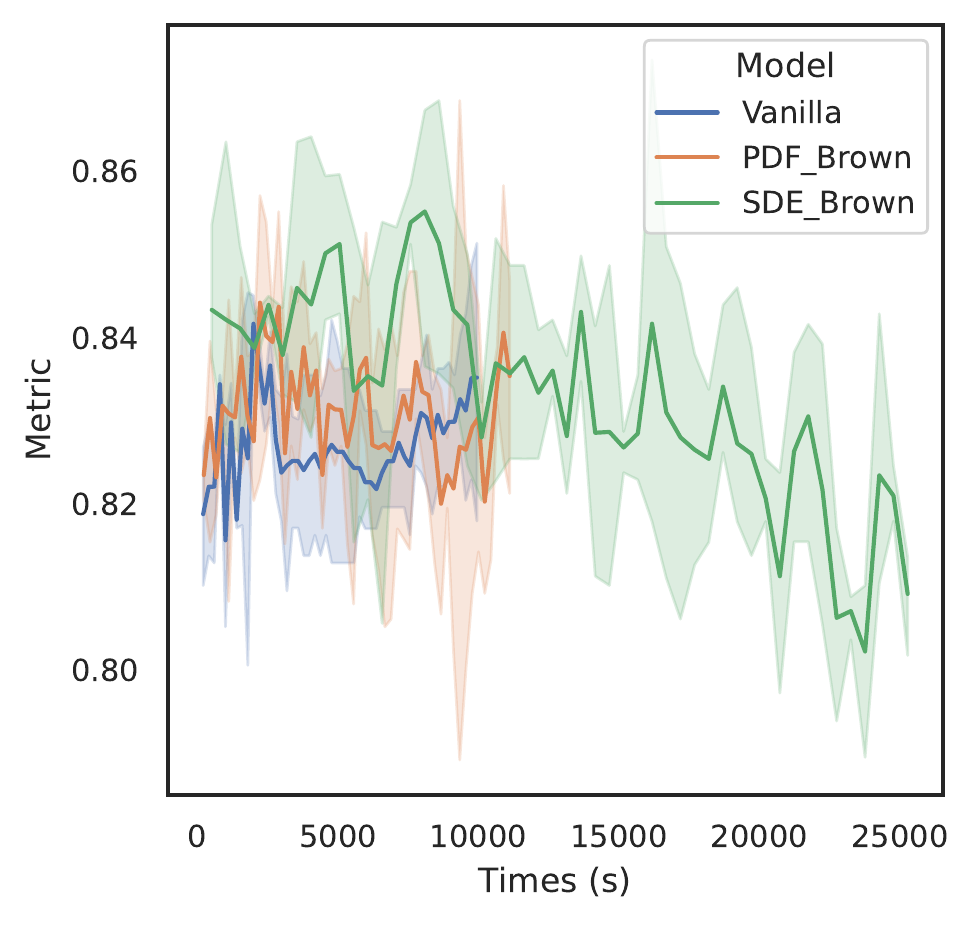}
        \caption{MRPC}
    \end{subfigure}
    \begin{subfigure}[b]{0.24\textwidth}
        \centering
        \includegraphics[width=\textwidth]{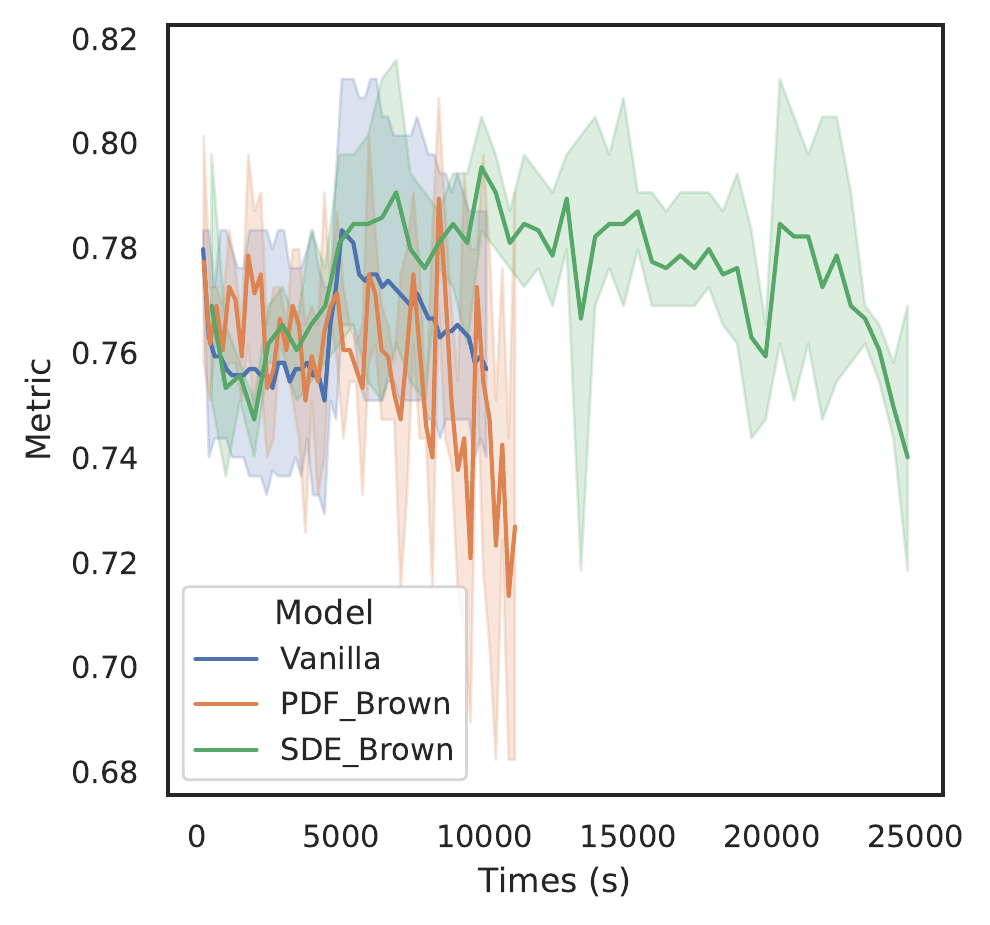}
        \caption{RTE}
    \end{subfigure}
    \hfill
     \caption{Time-Metric curve for regularizers on \emph{LoRA}.}
    \label{fig:speed_lora}
\end{figure*}
\begin{figure*}[t!]
    \centering
    \begin{subfigure}[b]{0.24\textwidth}
        \centering
        \includegraphics[width=\textwidth]{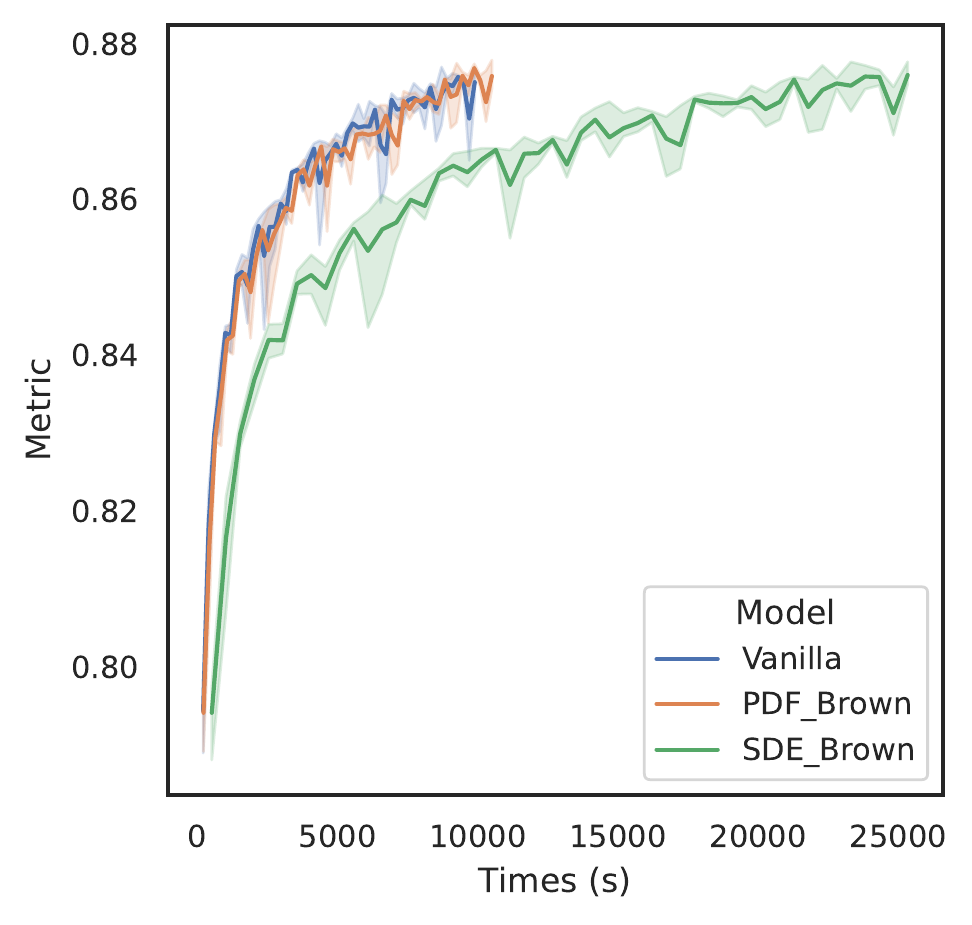}
        \caption{QQP}
    \end{subfigure}
    \hfill
    \begin{subfigure}[b]{0.24\textwidth}
        \centering
        \includegraphics[width=\textwidth]{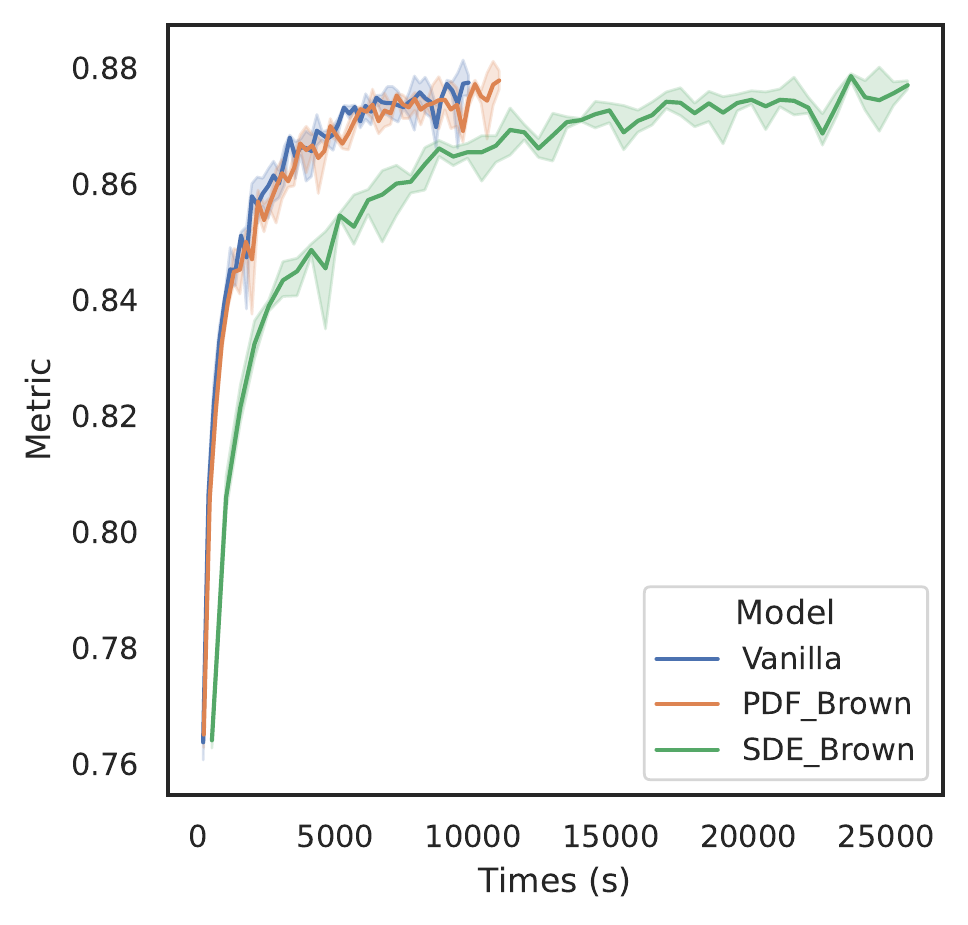}
        \caption{MNLI}
    \end{subfigure}
    \hfill
    \begin{subfigure}[b]{0.24\textwidth}
        \centering
        \includegraphics[width=\textwidth]{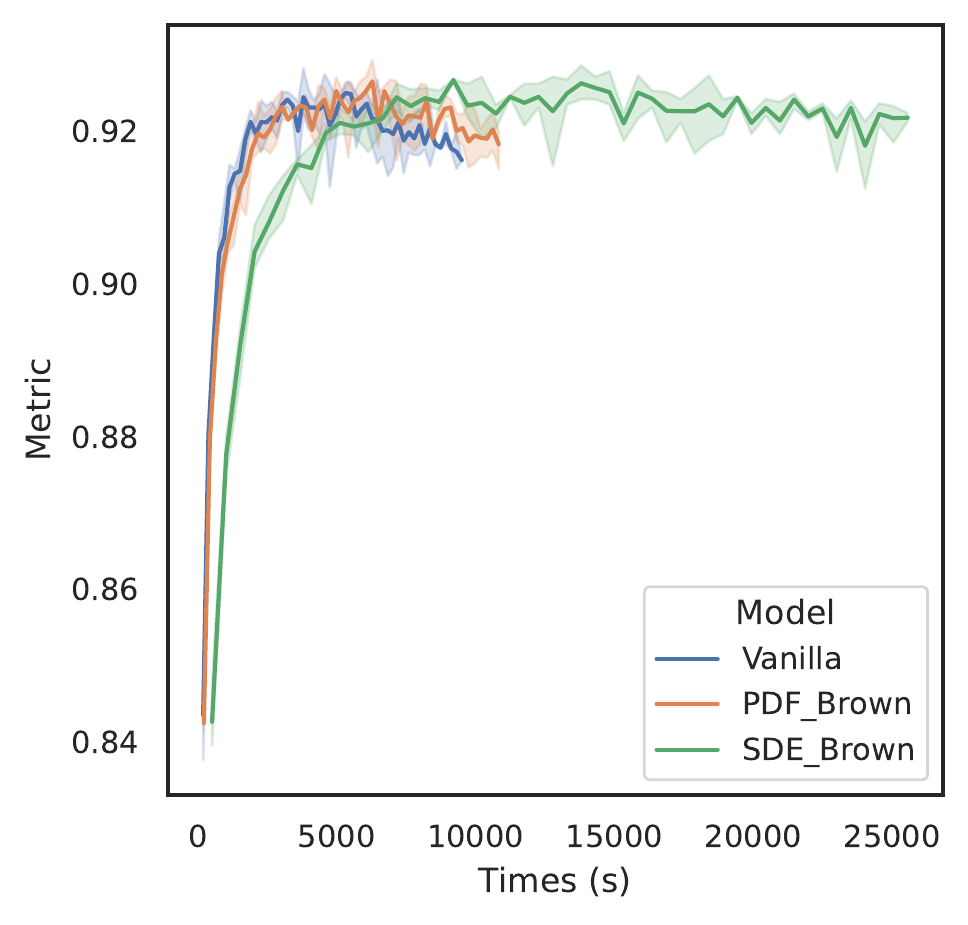}
        \caption{QNLI}
    \end{subfigure}
    \hfill
    \begin{subfigure}[b]{0.24\textwidth}
        \centering
        \includegraphics[width=\textwidth]{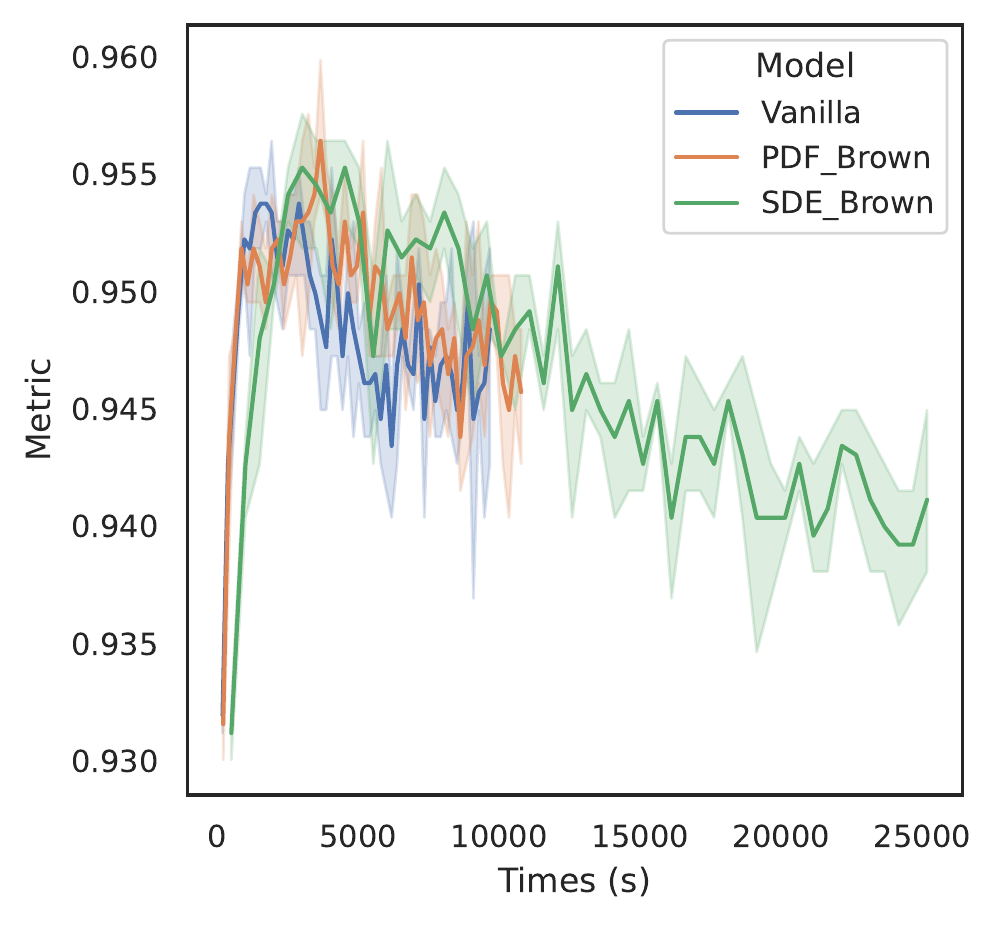}
        \caption{SST-2}
    \end{subfigure}
    \hfill
    \begin{subfigure}[b]{0.24\textwidth}
        \centering
        \includegraphics[width=\textwidth]{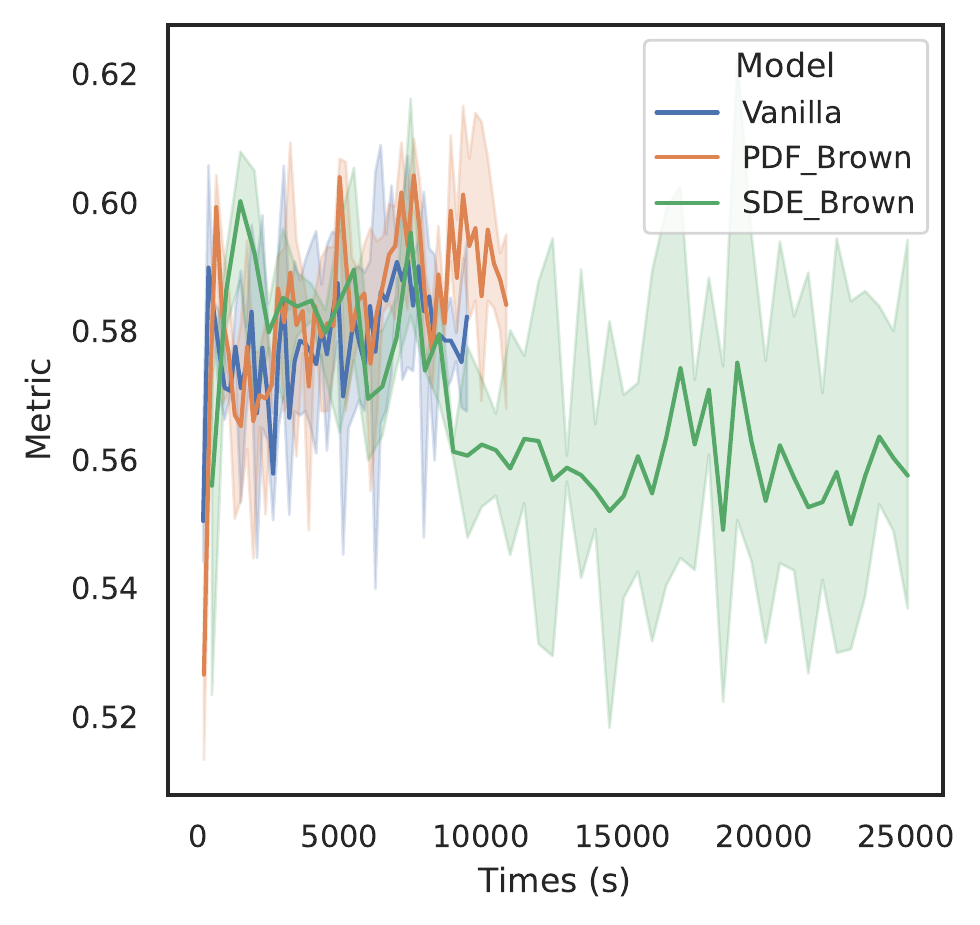}
        \caption{CoLA}
    \end{subfigure}
    \begin{subfigure}[b]{0.24\textwidth}
        \centering
        \includegraphics[width=\textwidth]{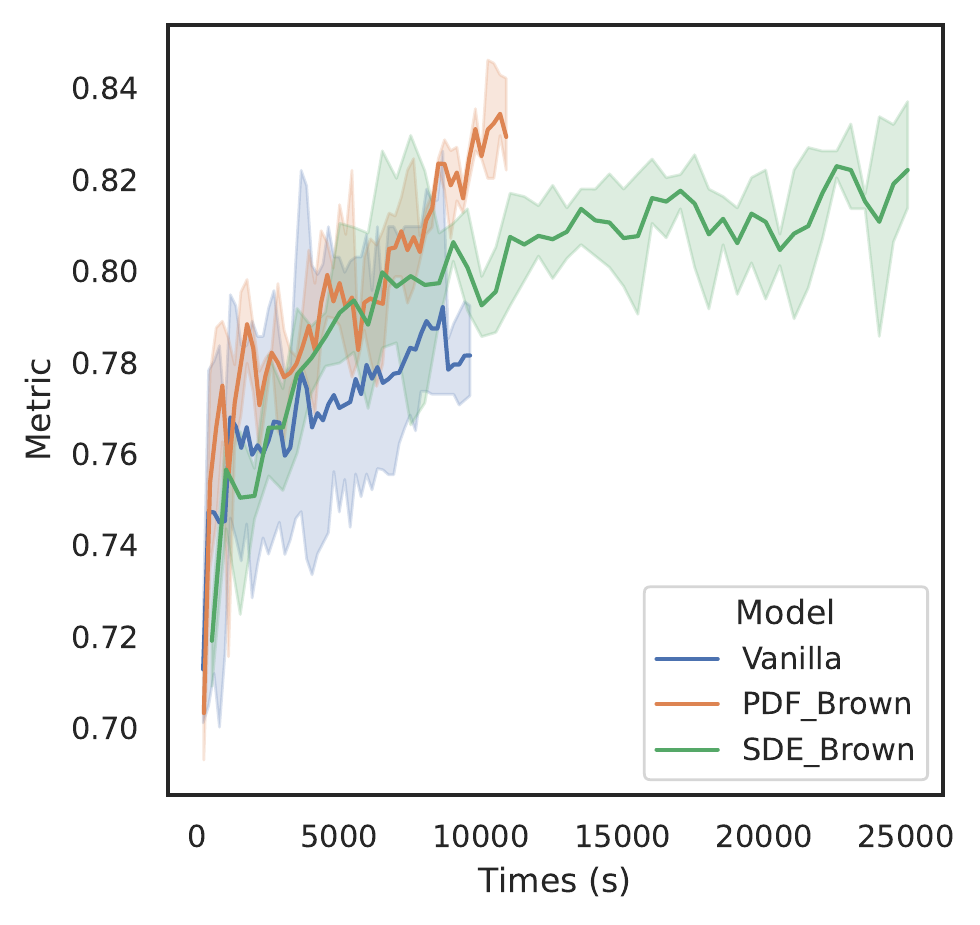}
        \caption{MRPC}
    \end{subfigure}
    \begin{subfigure}[b]{0.24\textwidth}
        \centering
        \includegraphics[width=\textwidth]{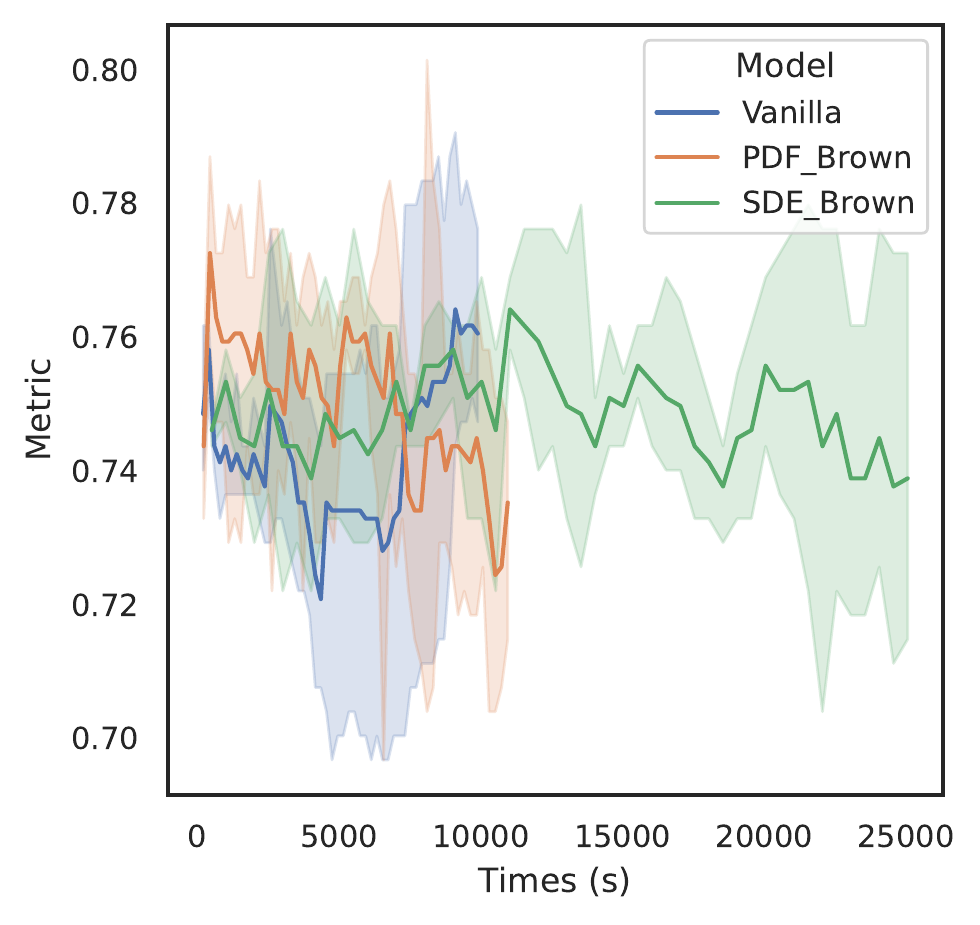}
        \caption{RTE}
    \end{subfigure}
    \hfill
     \caption{Time-Metric curve for regularizers on \emph{BitFit}.}
    \label{fig:speed_bias}
\end{figure*}
\begin{figure*}[t!]
    \centering
    \begin{subfigure}[b]{0.24\textwidth}
        \centering
        \includegraphics[width=\textwidth]{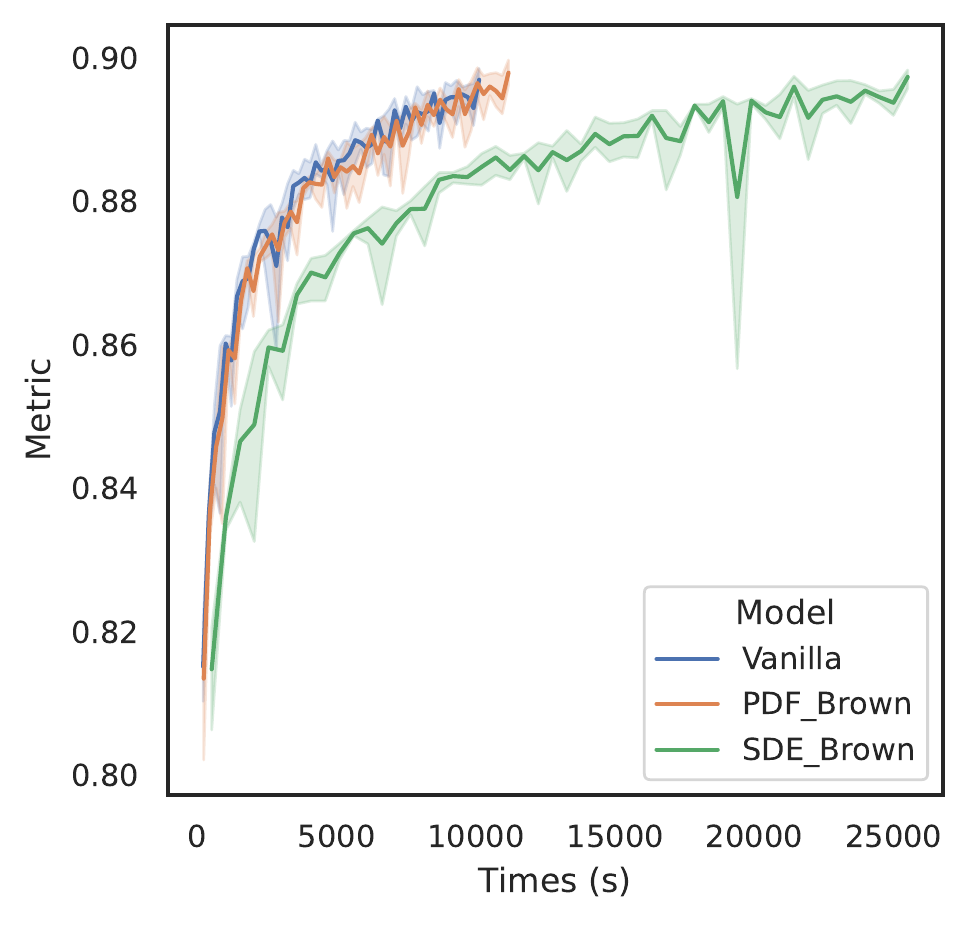}
        \caption{QQP}
    \end{subfigure}
    \hfill
    \begin{subfigure}[b]{0.24\textwidth}
        \centering
        \includegraphics[width=\textwidth]{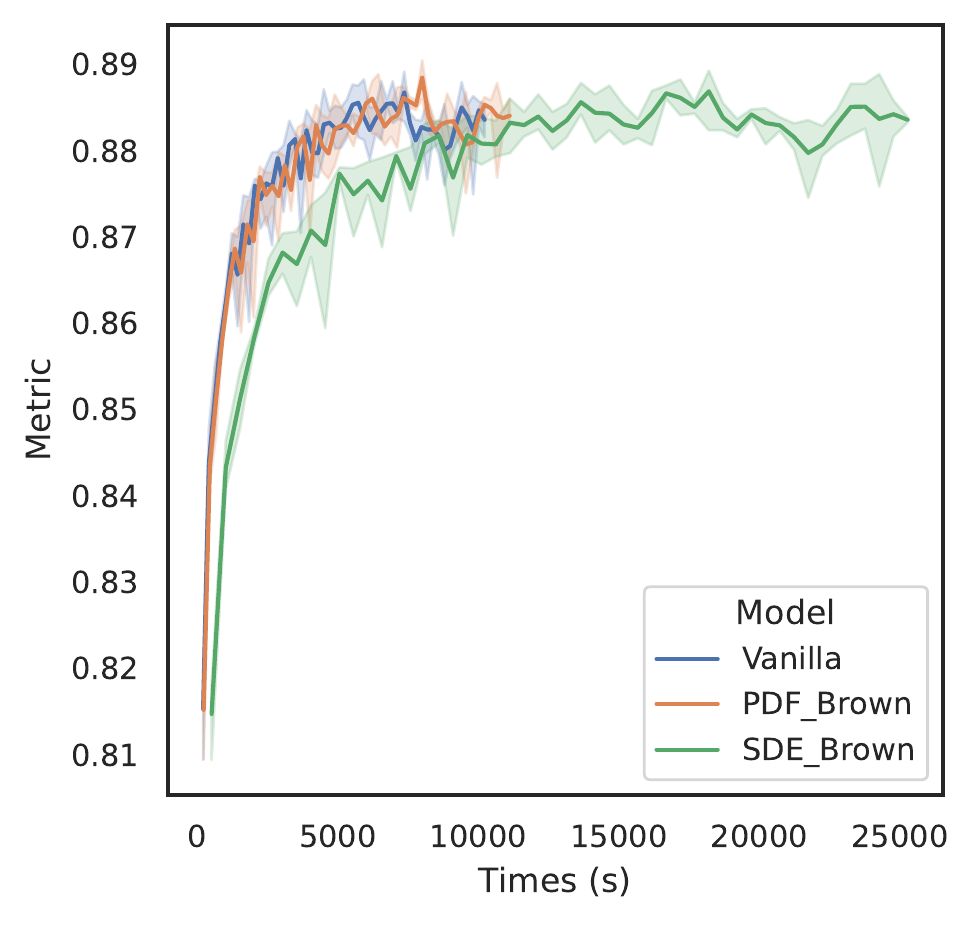}
        \caption{MNLI}
    \end{subfigure}
    \hfill
    \begin{subfigure}[b]{0.24\textwidth}
        \centering
        \includegraphics[width=\textwidth]{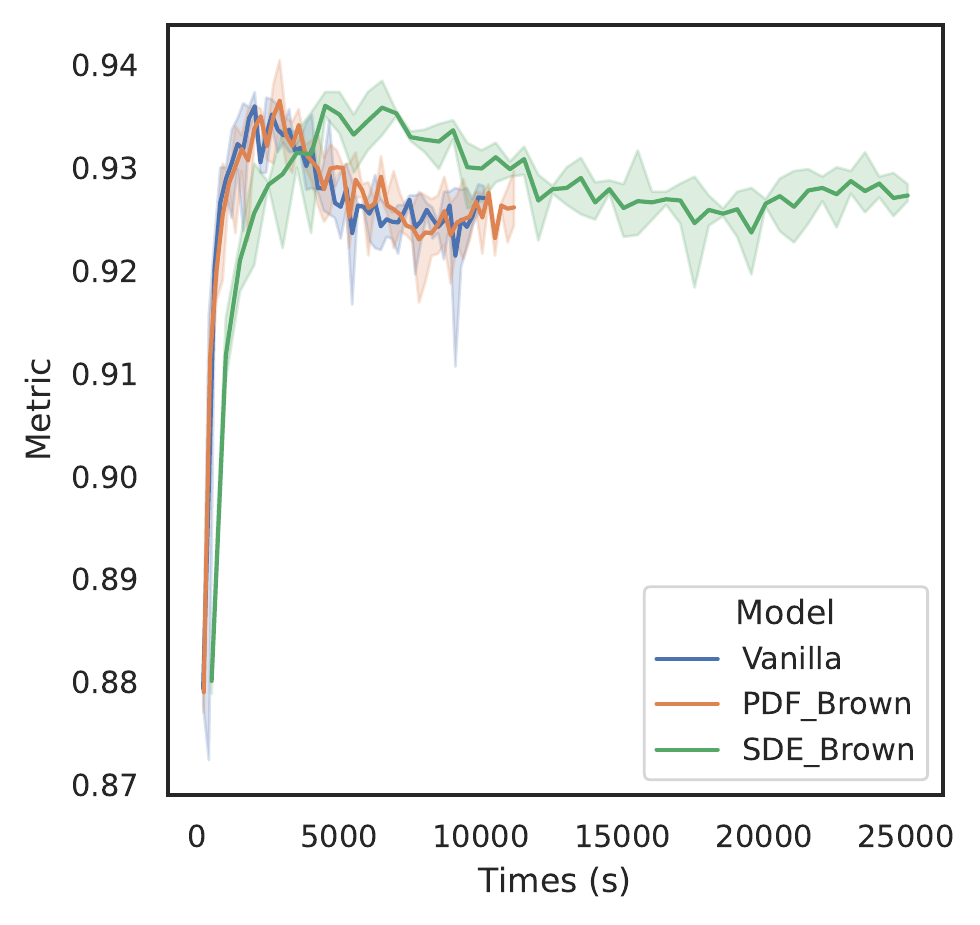}
        \caption{QNLI}
    \end{subfigure}
    \hfill
    \begin{subfigure}[b]{0.24\textwidth}
        \centering
        \includegraphics[width=\textwidth]{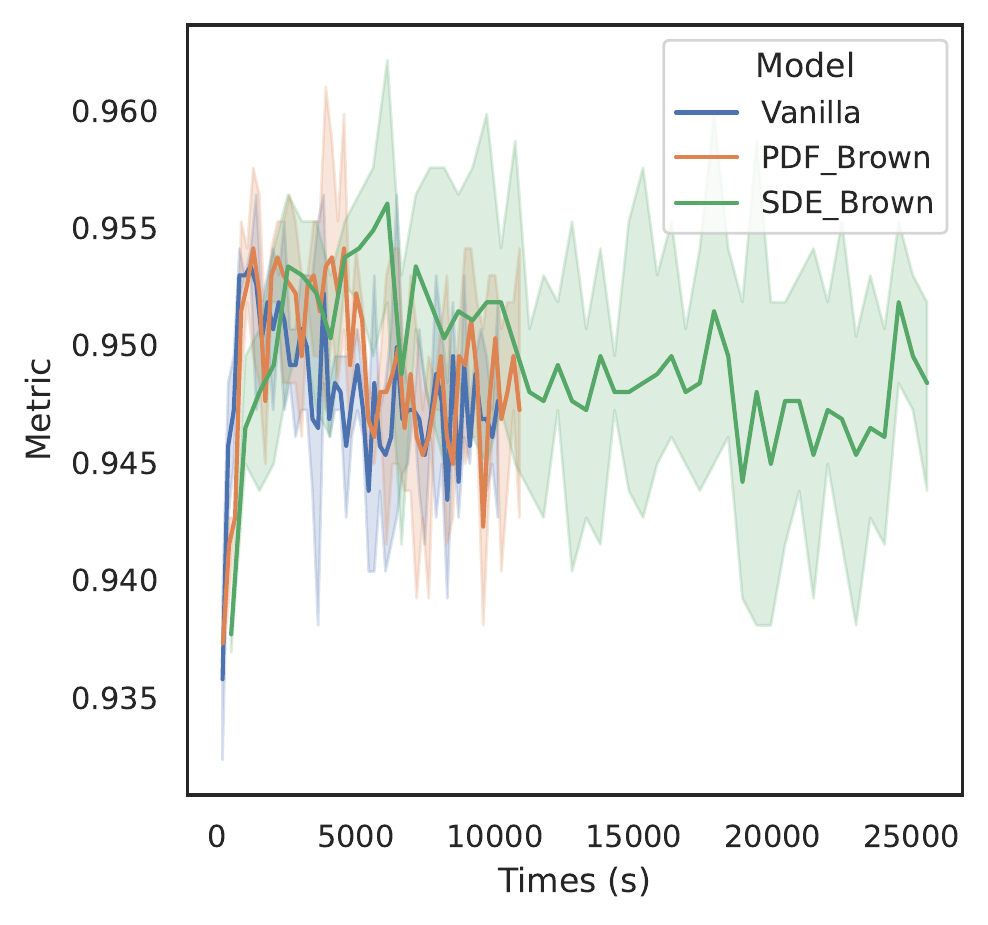}
        \caption{SST-2}
    \end{subfigure}
    \hfill
    \begin{subfigure}[b]{0.24\textwidth}
        \centering
        \includegraphics[width=\textwidth]{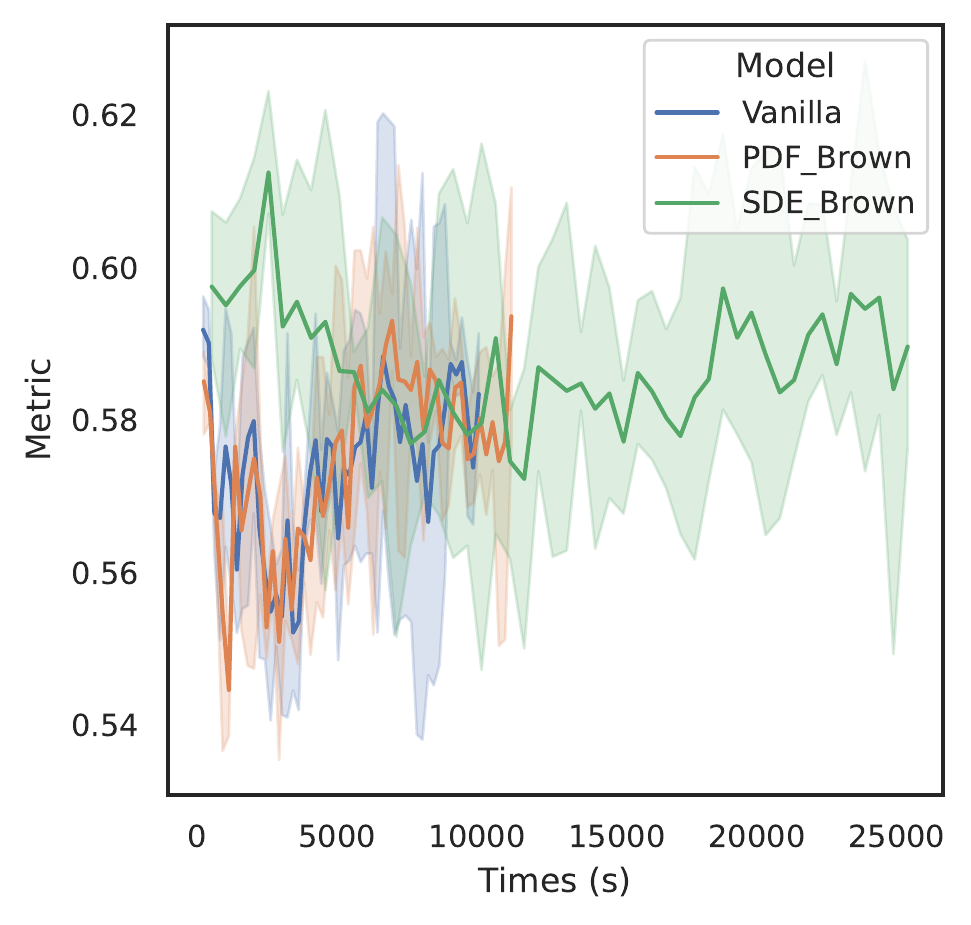}
        \caption{CoLA}
    \end{subfigure}
    \begin{subfigure}[b]{0.24\textwidth}
        \centering
        \includegraphics[width=\textwidth]{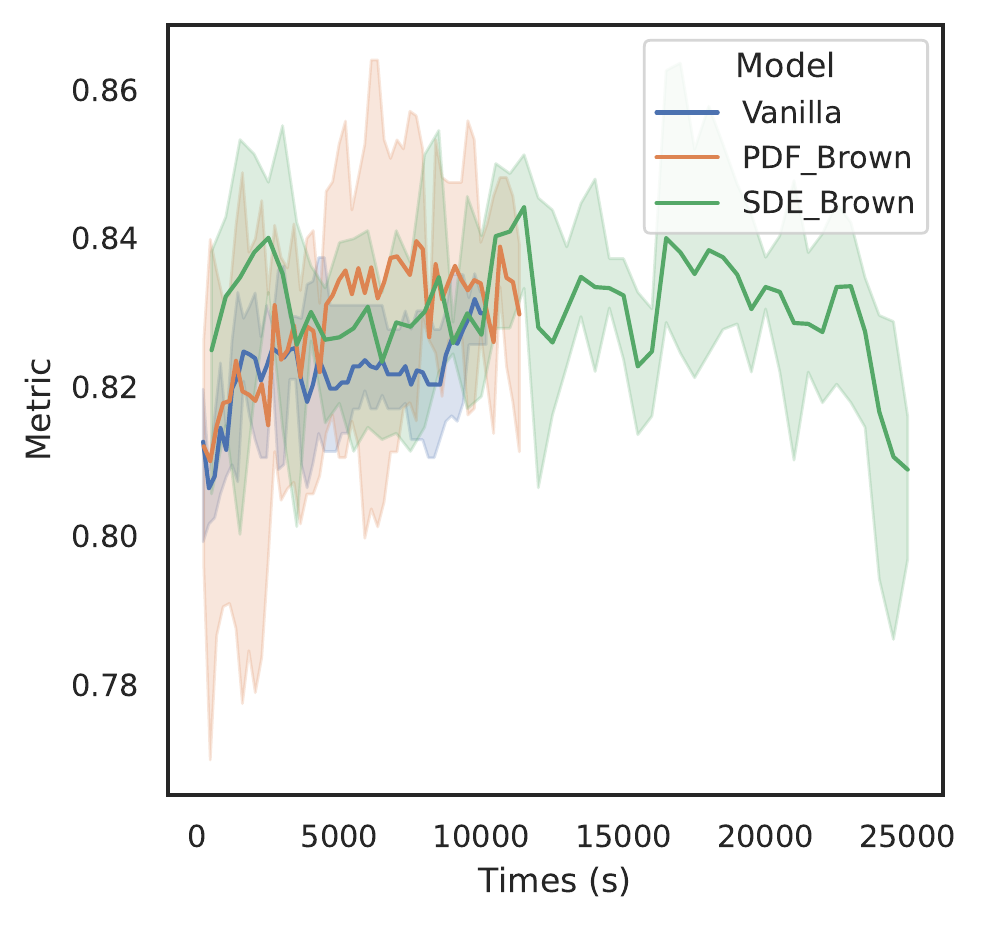}
        \caption{MRPC}
    \end{subfigure}
    \begin{subfigure}[b]{0.24\textwidth}
        \centering
        \includegraphics[width=\textwidth]{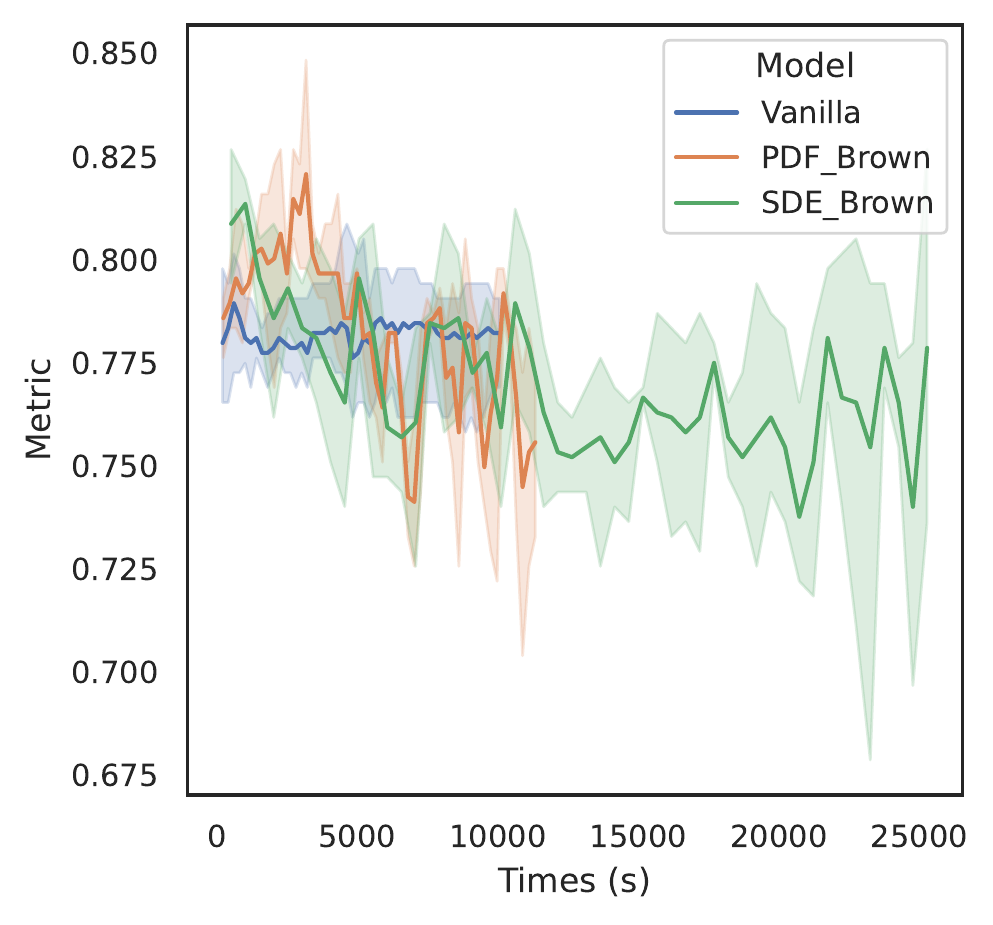}
        \caption{RTE}
    \end{subfigure}
    \hfill
     \caption{Time-Metric curve for regularizers on \emph{Adapter}.}
    \label{fig:speed_adapter}
\end{figure*}



\end{document}